\documentclass[acmlarge]{acmart}
\AtBeginDocument{%
  \providecommand\BibTeX{{%
    \normalfont B\kern-0.5em{\scshape i\kern-0.25em b}\kern-0.8em\TeX}}}
\setcopyright{acmcopyright}
\acmJournal{IMWUT}
\acmYear{2021} \acmVolume{5} \acmNumber{3} \acmArticle{120} \acmMonth{9} \acmPrice{15.00}\acmDOI{10.1145/3478074}
\usepackage{xcolor}
\usepackage{amsmath}
\usepackage{amsfonts}
\usepackage{graphicx}
\usepackage{textcomp}
\usepackage{microtype}
\usepackage{booktabs} 
\usepackage{xspace}
\usepackage{subfig}
\usepackage{mathtools}
\usepackage{algorithm, algorithmic}
\usepackage{hyperref}
\usepackage{multirow}

\newcommand{\latinphrase}[1]{\textit{#1}}
\newcommand{\etal}{\latinphrase{et~al.}\xspace}
\newcommand{\ie}{\latinphrase{i.e.}\xspace}
\newcommand{\eg}{\latinphrase{e.g.}\xspace}

\newcommand{\dana}{\text{DANA}\xspace}
\newcommand{\eval}{\latinphrase{validate}\xspace}
\newcommand{\etest}{\latinphrase{test}\xspace}
 
\begin{document}

\title{{\dana}: Dimension-Adaptive~Neural~Architecture for Multivariate Sensor Data}

\author{Mohammad Malekzadeh}
\authornote{Corresponding Author}
\email{m.malekzadeh@imperial.ac.uk}
\affiliation{%
  \institution{Imperial College London}
  \country{UK}
}
\author{Richard Clegg}
\email{r.clegg@qmul.ac.uk}
\affiliation{%
  \institution{Queen Mary University of London}
  \country{UK}
}
\author{Andrea Cavallaro}
\email{a.cavallaro@qmul.ac.uk}
\affiliation{%
  \institution{Queen Mary University of London}
  \country{UK}
}
\author{Hamed Haddadi}
\email{h.haddadi@imperial.ac.uk}
\affiliation{%
  \institution{Imperial College London}
  \country{UK}
}

\renewcommand{\shortauthors}{Malekzadeh et al.}

\begin{abstract}
Motion sensors embedded in wearable and mobile devices allow for dynamic selection of {\em sensor streams} and {\em sampling rates}, enabling several applications, such as power management and data-sharing control. While deep neural networks~(DNNs) achieve competitive accuracy in sensor data classification, DNN architectures generally process incoming data from a fixed set of sensors with a fixed sampling rate, and changes in the dimensions of their inputs cause considerable accuracy loss, unnecessary computations, or failure in operation. To address these limitations, we introduce a {\em dimension-adaptive pooling}~(DAP) layer that makes DNNs flexible and more robust to changes in sensor availability and in sampling rate. DAP operates on convolutional filter maps of variable dimensions and produces an input of fixed dimensions suitable for feedforward and recurrent layers. Further, we propose a {\em dimension-adaptive training}~(DAT) procedure for enabling DNNs that use DAP to better generalize over the set of feasible data dimensions at inference time. DAT comprises the random selection of dimensions during the forward passes and optimization with accumulated gradients of several backward passes. Combining DAP and DAT, we show how to transform existing non-adaptive DNNs into a {\em Dimension-Adaptive Neural Architecture}~(\dana), while keeping the same number of parameters. Compared to existing approaches, our solution provides better average classification accuracy over the range of possible data dimensions at inference time and does not require up-sampling or imputation, thus reducing unnecessary computations. Experimental results on seven datasets (four benchmark real-world datasets for human activity recognition and three synthetic datasets) show that \dana prevents significant losses in classification accuracy of the state-of-the-art DNNs and, compared to  baselines, it better captures correlated patterns in sensor data under dynamic sensor availability and varying sampling rates.   
\end{abstract}


\maketitle

\section{Introduction}\label{sec:intro}
Health and wellness applications~\cite{nam2016sleep, wang2016crosscheck, mohr2017personal} exploit motion sensors embedded in mobile and wearable devices to infer body movements and temporal changes in the physiological state of the wearer~\cite{chen2013unobtrusive, bulling2014tutorial, hansel2018put, reinertsen2018review}. For example, static acceleration (the magnitude and direction of the earth's gravitational force) helps to recognize the wearer's posture, whereas dynamic acceleration (changes in the velocity of the wearer) can be mapped to the wearer's  activity~\cite{shepard2008derivation}. The time series generated by these motion sensors are processed over temporal windows and classified by deep neural networks~(DNNs)~\cite{lecun2015deep, schmidhuber2015deep}, which mostly process sensor data with pre-defined, fixed dimensions~\cite{yang2015deep, ronao2016human, katevas2017practical, ignatov2018real, ordonez2016deep, zhao2018deep, yao2017deepsense, wang2019deep, jeyakumar2019sensehar, malekzadeh2020privacy}. Existing DNNs cannot reliably handle dynamic situations (\eg when the sampling rate changes or some sensors are dropped), which are important for energy preservation~\cite{chu2011balancing, liang2014energy}, privacy protection~\cite{raij2011privacy, malekzadeh2018mobile} and fault tolerance~\cite{choubey2015power, mathur2019unsupervised}. Moreover, DNNs are often trained on datasets collected from specific devices, but might be used for inference in a wider set of devices, with different combinations of sensors and sampling rates~\cite{koping2018general}.
  
Previous works, which are mostly based on non-DNN-based approaches, have addressed either sensor selection~\cite{zappi2008activity, gordon2012energy, zhu2013apt, ghasemzadeh2014power, saeedi2015activity, yang2020instance} or adaptive sampling rate~\cite{qi2013adasense, khan2016optimising, cheng2018learning, walton2018evaluation, hounslow2019assessing}, and dedicated a separate classifier for each feasible setting of available sensors~\cite{hsu2015two, choi2019embracenet, richoz2020transportation}. The trade-off between classification accuracy and power consumption varies with changes in the sampling rate, the sensor streams used by the classifier, the type of the current activity, and the physical characteristics of the wearer~\cite{yan2012energy}. Power-aware sensor selection may use a meta-classifier~\cite{zappi2008activity}, or the prediction of future activities from the current one to deselect sensors that are not useful for future activities~\cite{gordon2012energy}. Alternatively, a graph model representing the correlation among sensors with a greedy approximation can be used for sensor selection~\cite{ghasemzadeh2014power}, or a subset of sensors can be dynamically selected by minimizing an objective function that takes classification accuracy and the number of sensors as inputs~\cite{yang2020instance}. 

Defining the minimum sampling rate that captures discerning frequency components in different human activities, for example, to minimize power consumption~\cite{chu2011balancing}, is challenging~\cite{khan2016optimising} as the minimum required sampling rate varies across users,  activities, and sensor positions. Khan~\etal~\cite{khan2016optimising} show that the minimum sampling rate across activity recognition datasets varies between \mbox{22 and 63 Hz}; with a 99\% Kolmogorov-Smirnov similarity test~\cite{corder2014nonparametric}. It is therefore important to avoid aliasing to maintain the discernibility of activities characterized by higher frequencies~\cite{walton2018evaluation, hounslow2019assessing}. Yan~\etal~\cite{yan2012energy} propose to use a classifier with the highest sampling rate and then, after recognizing the current activity, switch to a lower sampling rate with another classifier and only monitoring whether the current activity changes or not. In order to lower power consumption, AdaSense~\cite{qi2013adasense} periodically detects changes in the current activity, by sampling data at higher frequencies, to check whether a lower sampling rate is suitable. To determine the best trade-off between power consumption and classification accuracy, Cheng~\etal~\cite{cheng2018learning} find an optimal classification model as well as appropriate sampling rates using a continuous state Markov decision process that is only appropriate for training simple classifiers, such as softmax regression, and not applicable in training DNNs. Current DNN architectures for processing multivariate sensor data~\cite{ordonez2016deep, ignatov2018real, ronao2016human} cannot handle changes in the dimensions of their input without adding a data preprocessing stage to their pipeline. Thus, there is a lack of DNN architectures that can be trained for reliably and accurately handling variable data domains at inference time without any preprocessing stages.

In this paper, we introduce Dimension-Adaptive Neural Architecture~(\dana), a unified, flexible and, more robust solution to variable sampling rates and sensor selection. At inference time, \dana works on any combination of sensors that were present at training time. Specifically, we introduce a dimension-adaptive pooling~(DAP) layer that captures temporal correlations between consecutive samples and dynamically adapts to all feasible data dimensions. To enable DNNs, that use DAP, to generalize at inference time over the set of feasible dimensions, we propose a dimension-adaptive training~(DAT) procedure, which incorporates dimension randomization and optimization with accumulated gradients. In each forward pass, DAT re-samples a batch of time widows to a new rate and may also randomly remove streams from some sensors. Then, gradients from multiple batches are accumulated before updating the parameters. Combining DAP and DAT, we show how to transform existing DNNs into an adaptive architecture, with the same number of parameters and classification accuracy, and improving the inference time. In addition to allowing adaptive sampling rate and sensor selection with a unified solution, \dana enables the reduction of the computations to be performed according to the dimensions of the sampled data, which is desirable for power-constrained devices. We compare \dana with other existing baselines that use non-adaptive DNNs in terms of {\em the average classification accuracy} across a range of possible scenarios at inference time, where the more accurate a solution is on average, the more robust that solution is considered.
Experimental results on seven datasets (four benchmark real-world datasets for human activity recognition and three synthetic datasets) show that \dana can maintain classification accuracy in dynamic situations where existing DNNs drop their accuracy, and better generalizes to unseen environments. Code and data to reproduce results are publicly available at~\url{https://github.com/mmalekzadeh/dana}.

The paper is organized as follows. Section~\ref{sec:related_work} covers the background and related work. In Section~\ref{sec:dap}, we elaborate on how a DAP layer works, and in Section~\ref{sec:dat} we show how DNNs that use DAP can be efficiently trained using DAT. In Section~\ref{sec:eval} we evaluate \dana and compare it to non-adaptive DNNs and other baselines on real-world datasets. In Section~\ref{sec:synth}, we provide experiments on three synthetic datasets where we control the correlation between two sensors and analyze the performance of \dana, compared to other baselines. Finally, in Section~\ref{sec:discc}, we discuss some limitations and open directions in our work, for future studies. 

As for the notation, we will use lower-case {\it italic}, \eg $x$, for single-valued variables; upper-case {\it italic}, \eg $X$, for single-valued constants; standard {\bf bold} font, \eg $\mathbf{X}$, for vector, matrices, and tensors; and standard blackboard bold, \eg $\mathbb{X}$, for unordered sets. We use $\times$ to separate the size of each dimension while $\cdot$ denotes product. 

\section{Background and Related Work}~\label{sec:related_work}

The architecture of a DNN is mainly defined by the type and number of  layers, and how these layers are connected to each other~\cite{goodfellow2016deep}. The most common DNN architecture for sensor data classification consists of a convolutional neural network~(CNN)~\cite{lecun1995convolutional} followed by a feedforward neural network~(FNN) or a recurrent neural network~(RNN)~\cite{yang2015deep, ordonez2016deep, ronao2016human, yao2017deepsense,  ignatov2018real,  zhao2018deep, wang2019deep, jeyakumar2019sensehar, malekzadeh2020privacy}. The main difference between FNNs and RNNs is that a FNN has no feedback connections.  A feedback connection feeds the output of a neuron in a neural net, to the neuron itself. Every neuron in an RNN includes a feedback connection which helps  to capture temporal correlations among consecutive samples. 

A CNN is inherently adaptive to input data of variable dimensions~\cite{long2015fully} and is mainly used for feature extraction for the downstream task~\cite{goodfellow2016deep}. {In computer vision, a prominent and motivational property of CNNs is that these layers can learn to be {\em invariant} to some changes (such as {\em translation}, {\em rotation}, or {\em scaling}) of objects in an image at inference time. For instance, it is well known that the convolution operation commutes with respect to the translation operation, meaning that when we convolve data $\mathbf{X}$ with a filter $\mathbf{F}$, the output of is the same if (i) we convolve data and filter, and then translate the convolved output, or if (ii) we first translate data and, then convolve the translated data with the filter; \ie  $\tau(\mathbf{X}*\mathbf{F}) = \tau(\mathbf{X})*\mathbf{F}$, where $\tau$ and $*$ denote the convolution and translation operations, respectively. Such a property of convolutional layers, alongside other architectural (\eg max-pooling operation) and training (\eg data augmentation) techniques, allow desired invariances in image processing to be satisfied~\cite{lecun1995convolutional, xu2014scale, long2015fully, he2015spatial, kayhan2020translation}. These capabilities motivate our design of the DAP layer in Section~\ref{sec:dap} as well as the proposed DAT Section~\ref{sec:dat}, which are aimed to force the ultimate model to be invariant to changes in sensor data characteristics, such as the sampling rate of sensor streams (similar to translation and scaling) and the {\em order} (or permutation) of available sensors (similar to  rotation). In this paper, we put such changes in the sampling rate, order, and/or availability of sensor streams under the single umbrella of changes in the {\em dimensions} of sensor streams.} 

Features extracted by CNN to an FNN or an RNN improve the classification accuracy and provide better generalization compared to raw sensor data~\cite{bengio2013representation}. However, FNNs only work on fixed-dimension input data, and RNNs only accept a fixed number of streams, thus making the combination of CNNs with FNNs/RNNs non-adaptive to the changes in input dimensions. To address this limitation, a preprocessing stage can be added to the inference pipeline that  up/down-samples data to a fixed  rate or imputes dummy data to compensate for missing samples~\cite{ae2018missing}, but these solutions reduce classification accuracy and can even raise other challenges~\cite{quiring2020adversarial}. For example, when downsampling data, some discerning patterns are either removed or changed such that the resampled data will be miss-classified, which can be exploited by adversaries that take advantage of their knowledge about the used downsampling method to generate adversarial examples~\cite{quiring2020adversarial}. Our work eliminates the need for up/down-sampling the DNN's input at inference time, and thus less susceptible to such attacks.   

Alternatively, to turn deep architectures based on CNNs and FNNs/RNNs adaptive to  changes in input-data dimensions, a preferable solution is to include an adaptive layer between the CNN and the FNN/RNN. A global pooling layer~\cite{lin2013network} takes the maximum or average over each filter map of the CNN's output to make a one-dimensional vector whose length is equal to the number of the convolutional filter maps, and independent of the dimensions of these filter maps. The main drawback of global pooling is ignoring the inherent spatial structure of the data, and we show that it causes accuracy loss. 

To mitigate the shortcomings of global pooling in visual object recognition, spatial pyramid pooling~(SPP)~\cite{he2015spatial} runs pooling on a pyramid that is created by hierarchically dividing a feature map into equally sized segments. SPP is supported by a weight-sharing mechanism for training CNN-FNN architectures on different image resolutions, and is used for transfer learning: the CNN part trained on the source dataset can be used with other FNNs on the target dataset~\cite{abdu2019versatl}.  
SPP has promising results in image processing~\cite{zhao2017pyramid}, but is not applicable to CNN-RNN architectures where recurrent layers, such as LSTMs~\cite{hochreiter1997long}, need two-dimensional inputs that preserve the temporal correlation among consecutive  samples and across different sensor streams. Therefore, global or pyramid layers are not directly applicable to RNNs, which are often preferred for time-series classification~\cite{karim2017lstm}. Moreover, given DNNs for multivariate sensor data, there is no way for existing pooling layers to make them adaptive to temporal changes in input layer dimensions without changes that increase or decrease the number of trainable parameters and consequently may affect the model's accuracy. 

A DNN that can process  variable input dimensions should also be trained to produce accurate outcomes when the dimensions of its input can change. {\em Weight Averaging}~\cite{he2015spatial} was proposed to build a single DNN by averaging the weights across multiple DNNs trained in parallel across different settings, which is also used in federated learning where the goal is to collaboratively train a shared model across different users~\cite{bonawitz2019towards}. Similarly, in meta-learning, the goal is to train a model across a large number of different tasks~\cite{finn2017model}. {\em Reptile}~\cite{nichol2018first} is a meta-learning algorithm that uses the average parameters of multiple DNNs, each trained on a different task, as input to the optimizer for updating the DNN's parameters, instead of directly using the gradients of the loss function.

Unlike existing adaptive pooling layers~\cite{lin2013network, he2015spatial}, our proposed DAP layer considers the temporal correlations in data as well as the absence of some of the sensors. This is particularly important as RNNs are more efficient in processing temporal data than FNNs~\cite{ordonez2016deep, yao2017deepsense,zhao2018deep}. Moreover, the proposed DAT resolves the need for weight sharing via training multiple DNNs, providing a faster and more accurate training algorithm, compared to the existing ones~\cite{he2015spatial, bonawitz2019towards, nichol2018first}. Finally, unlike the methods proposing multimodal fusion~\cite{choi2019embracenet, richoz2020transportation}, our solution resolves the need for training and deploying multiple DNNs which is particularly important for applications running mobile and wearable devices with limited power and processing resources.

\section{Dimension-Adaptive Pooling (DAP)}\label{sec:dap}

\begin{figure*}[t!]
    \centering
    \includegraphics[width=\textwidth]{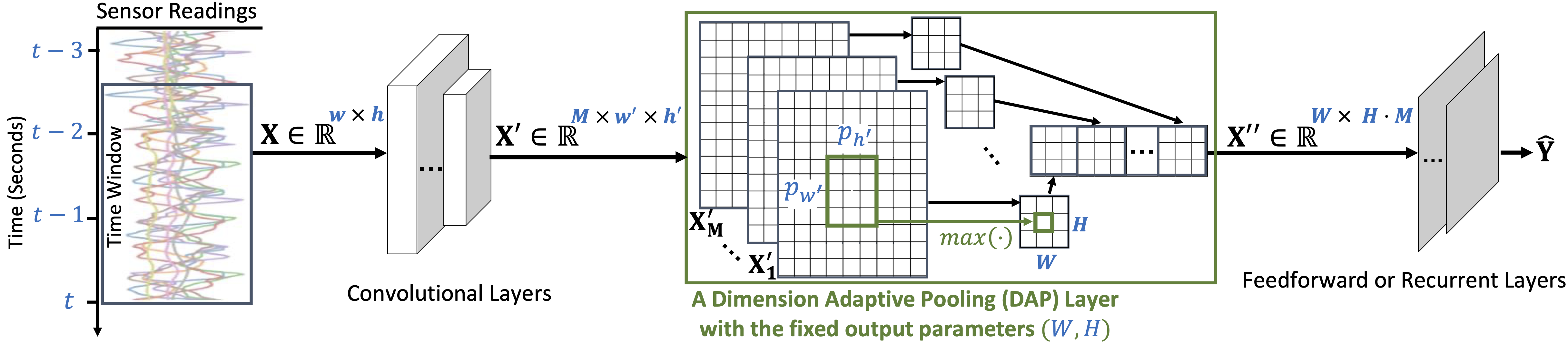}
    \caption{A DAP layer enables a DNN performing classification on input data of variable dimensions. At time $t$, a window~$\mathbf{X}$ of dimensions $ w \times h$ is received by the DNN. The number of {\em streams}, $h$, depends on the number of available sensors, while the number of {\em samples}, $w$, per each stream depends on the current sampling rate~(in Hz) and the length of time window~(in seconds). For example, a $2.5$-seconds time window of 50~Hz data generated by three 3-dimensional motion sensors (\eg accelerometer, gyroscope, and magnetometer of a smartwatch) has dimensions of $w=125$ and $h=3\cdot 3 =9$. Convolutional layers process $\mathbf{X}$, then DAP is applied to the $M$ outputs of the last convolutional layer $\mathbf{X'}$. DAP uses an adaptive kernel of size $p_{w^{'}} \times p_{h^{'}}$ that is calculated based on DAP's chosen parameters, $(W,H)$, and the dimensions of CNN's outputs, $w'\times h'$. Data of fixed dimensions, $\mathbf{X^{\prime\prime}}$, is provided for the following feedforward or recurrent layers that output the classification's result $\hat{\mathbf{Y}}$.} 
    \label{fig:dana_arch}
\end{figure*}

We propose the DAP layer to handle situations where one or more sensors may dynamically be deselected at inference time. The flexibility of DAP aims not only to make DNNs adaptive to changes in the dimension of data, but also to allow efficiently training the DNN such that it provides reliable performance across several combinations of data dimensions.

Considering a multivariate time-series~(see Fig.~\ref{fig:dana_arch}), let $\mathbf{X}$ be a time window of dimensions $w\times h$ where $w$ is the number of samples, which depends on the sampling rate and the length of the time window, and $h$ is the number of sensor streams. Particularly in our case, motion sensors have three spatial axes $(x, y, z)$, thus $h=3\cdot s$, where $s$ is the number of available sensors. Thus, if a sensor is not available, or deselected, then $\mathbf{X}$ will have 3 fewer streams. In real-world situations, when a sensor is available it produces measurement in all three axes, and it is not usual that, for example, an accelerometer only produces measurement on only one axis or two of its three axes. There might be situations where the produced data for an axis is noisy or wrong,\eg due to hardware issues, but we still receive a sensor stream in that axis. We emphasize that we do not investigate distinguishing between faulty or normal sensor streams in this paper, and we only consider the availability or not availability of sensor streams. 

A DNN can cope with inputs of variable dimensions by using a convolutional layer as the input layer. A two-dimensional convolutional layer slides $M$~fixed-sized {\em filters} across the input data, and computes the dot products between each filter's entries and the data at the current position of the filter, resulting in $M$ two-dimensional filter maps ($\mathbf{X}'$s in Fig.~\ref{fig:dana_arch}). In a stack of convolutional layers, the dimensions of the CNN's output, $w'\times h'$, mainly depend on the dimensions of the input data\footnote{Design parameters such as the number of convolutional filters,  $M$, size of the filters, the chosen padding mode, and the stride length of the filters are fixed at inference time.}, $ w \times h$. 

The outputs of a CNN's last layer are in a three-dimensional shape\footnote{Note that in our notation, $\times$ separates the size of each dimension while $\cdot$ denotes product.}, $w' \times h' \times M$, which is not suitable as direct input to a FNN or RNN. The usual practice is to reshaped such three-dimensional data into one-dimensional data of size $w' \cdot h' \cdot M$ for FNNs (by stacking all $h'\cdot M$ vectors of length $w'$ next to each other) or two-dimensional data $w' \times h'\cdot M$ for RNNs (by stacking all $M$ matrices of length $w' \cdot h'$ next to each other). For specific details, see ReshapeLayer in TensorFlow library~\cite{abadi2016tensorflow}. However, these typical DNNs can cope only with input data of fixed dimensions.  One cannot simply use existing layers to make DNNs, which are proposed for processing sensor time-series, adaptive to the sampling rate and sensor selection without imposing any architectural changes. For instance, the single-dimensional data produced by SPP~\cite{he2015spatial} is not appropriate for RNNs where the input must be provided in two dimensions: consecutive samples and parallel streams. DAP addresses the aforementioned limitations without enforcing assumptions on the DNN architecture to be used. DAP builds upon the global and pyramid pooling ideas and aims to map the outputs of dimensions $w'\times h' \times M$ into data whose dimensions are consistent with the next FNN/RNN layer. The size of the pooling filters in DAP is not limited to be square, hence it generalizes existing adaptive layers~\cite{lin2013network, he2015spatial}.

\begin{algorithm}[t!]
  \caption{\textbf{-- Dimension Adaptive Pooling (DAT).} A layer to be located between the last convolutional layer and the first feedforward/recurrent layer in a DNN. DAT is designed to make DNN flexible to changes in sampling rate and sensor availability. A flexibility that is further utilized in Algorithm~\ref{alg:dat} (DAT) for training and making the DNN robust to such changes at inference time, in terms of the average classification accuracy.}
   \label{alg:dap_layer}
   \begin{algorithmic}[1] 
   \STATE {\bfseries Input:} $\mathbf{X'}$: filter maps received from a CNN, ($W$, $H$): the fixed output parameters. 
   \STATE {\bfseries Output:} $\mathbf{X''}$: input to the next layer.
   \STATE $M, w', h' = dimensions\_of(\mathbf{X'})$
   \STATE $\mathbf{X''}$ = \{\} 
     \FOR {$m = 1$ {\bfseries to} $M$}
        \STATE $\mathbf{V} = \mathbf{X}'[m]$
           \STATE $\mathbf{Z} = copy\_of(\mathbf{V})$
           \STATE $a =  {max}\big(\lceil(H-h')/3\rceil,0\big)$
           \FOR {$i = 1$ {\bfseries to} $a$}
           \STATE $\mathbf{Z} = {concatenate\_vertically}(\mathbf{Z},\mathbf{V})$ 
           \ENDFOR
           \STATE $\mathbf{Z} = \mathbf{Z}[0:w',0:max(h',H)]$ 
           \STATE $p_{w^{'}} = {w'/ W}$  
           \STATE $p_{h^{'}} = {h'/ H}$
           \STATE $\mathbf{Q} = \{\}$
           \FOR {$i=1$ {\bfseries to} $W$} 
           \FOR {$j=1$ {\bfseries to} $H$} 
           \STATE $r_1 = \lfloor i\cdot p_{w^{'}} \rceil$  
           \STATE $r_2 = \lfloor (i+1)\cdot p_{w^{'}} \rceil$  
           \IF {a=0}
           \STATE $c_1 = \lfloor j\cdot p_{h^{'}}\rceil$  
           \STATE $c_2 = \lfloor (j+1)\cdot p_{h^{'}}\rceil$
           \ELSE
           \STATE $c_1 = \lfloor j\cdot \lfloor(a+1)\cdot p_{h^{'}}\rfloor\rceil$  
           \STATE $c_2 = \lfloor(j+1)\cdot \lfloor(a+1)\cdot p_{h^{'}}\rfloor\rceil$
           \ENDIF 
           \STATE $\mathbf{Q} = append\Big(\mathbf{Q}, max\big(\mathbf{Z}[r_1:r_2,c_1:c_2]\big)\Big)$ 
           \ENDFOR 
           \ENDFOR
        \STATE $\mathbf{X''} = append(\mathbf{X''}, \mathbf{Q})$ 
     \ENDFOR
\end{algorithmic}
\end{algorithm}

Algorithms~\ref{alg:dap_layer} shows the functionality of the DAP layer for DNNs processing motion sensor data\footnote{$\lfloor\rfloor$ denotes floor, $\lceil\rceil$ denotes ceiling, and $\lfloor\rceil$ denotes  rounding to the nearest integer.}. Let $(W,H)$ be the pre-specified hyper-parameters for pooling all the $M$ feature maps into an output $\mathbf{X}''$ of single dimension of size $W\cdot H\cdot M$ (if the next layer is FNN) or two dimensions $W \times H\cdot M$ (if the next layer is RNN). DAP first calculates the pooling parameters $(p_{w^{'}}=\lfloor\frac{w'}{W}\rfloor, p_{h^{'}}=\lfloor\frac{h'}{H}\rfloor)$ for the received inputs. It then, for every segment of size $(p_{w^{'}}, p_{h^{'}})$ on the received input, chooses the maximum value to create a the output which aims to be of fixed-dimensions $(W,H)$. For larger data dimensions, the pooling parameters adaptively cover a larger segment of the data and for  smaller data dimensions the pooling parameters will shrink appropriately. Hence, DAP always produces an output of fixed dimensions. 

As an example, consider a CNN-RNN and $s=3$, $w'=128$, $h'=9$, $W=16$, $H=3$, $M=32$. The maximum of every 8 consecutive samples among all 3 axes of each sensor will be chosen for the output, thus producing $\mathbf{X}''$ with dimensions $16 \times 3\cdot32$. If we deselect two sensors and choose a sampling rate half of the original, this means $w'=64$ and $h'=3$, then DAP adaptively keeps the output fixed, by choosing the maximum of every $8/2=4$ consecutive samples of each axis of the only available sensor.

The input of DAP has dimensions $w' \times h'$, which can be variable at training and inference time; whereas the output  of DAP, $\mathbf{X''}$, is fixed and includes $W\cdot H \cdot M$ values. The first inner loop in Algorithm~\ref{alg:dap_layer} (lines 9-11) handles situations when one or more sensors are unavailable. Depending on the value of $H$, we may need to replicate some of the existing streams to satisfy fixed-sized outputs. For a DAP layer with $W=16$, $H=9$, and 3 sensors, the possible situations are as follows.

(1) All the sensors are available~($h'=9$), thus $a=0$ (in line 8 of Algorithm~\ref{alg:dap_layer}) and the algorithm skips the loop~(Lines 9-11).

(2) One sensor is unavailable~($h'=6$), thus $a=1$ and the loop fills the gap by {\em vertically concatenating} a copy of the data from available sensors to the second dimension of the input data. Therefore $\mathbf{Z}$ will have  $h'=12$ streams and Line 12 truncates $\mathbf{Z}$ on its second dimension to ensure that its second dimension satisfies $H=9$.

(3) Two sensors are unavailable~($h'=3$), thus $a=2$ and the algorithm fills the gap by {\em vertically concatenating} a copy of the data from the only available sensor two times. So, we will again satisfy $H=9$.
 
\noindent Note that Line 12 guarantees that, in case of concatenating data to $\mathbf{Z}$, the number of streams in data never exceeds $H$, and it has a neutral effect when $h'=H$. The next two loops, in Lines 16-29, perform the pooling operation over the consecutive segments of $\mathbf{Z}$. 

Following our example, in each iteration we consider a segment of data including $w'/ W $ samples and $h'/ H$ streams. For example, if $w'=96$ then the maximum of every $6$ consecutive samples of each sensor stream (\eg the accelerometer's x-axis) is calculated and appended to $\mathbf{Q}$~(Line 27). If we only change $H$ to $3$, instead of $9$, $a$ will be zero, thus no concatenation happens. On the other hand, every segment of data includes $6\cdot3=18$, $6\cdot2=12$, or $6\cdot1=6$ samples if all three, two, or only one sensor(s) are (is) available, respectively.  Therefore, DAP adaptively changes the size of pooling segments to ensure that the output dimensions are fixed. Also note that, in Lines 16-29, the first loop runs on the first dimension $W$ and the second loop on the second dimension $H$: this order  preserves the temporal correlation between consecutive samples, which is necessary for DNNs that use recurrent layers.

\section{Dimension Adaptive Training (DAT)}\label{sec:dat}

\begin{algorithm}[t!]
   \caption{\textbf{-- Dimension Adaptive Training (DAT).} To train A DNN that is made flexible to changes in sampling rate and sensor availability by DAP layer in Algorithm~\ref{alg:dap_layer}. During training, DAT randomly encounters the DNN with a range of feasible data dimensions at inference time and imitatively updates the DNN with the accumulated gradients of multiple feasible data dimensions.}  
   \label{alg:dat}
\begin{algorithmic}[1]
   \STATE {\bfseries Input:} $\mathbb{D}$: training datasets, $\Theta$: trainable parameters, $E$:~number of epochs, $\mathbb{U}$: a subset of all feasible dimensions, $B$:  number of batches used in each optimization round, $K$: size of each batch.
   \STATE {\bfseries Output:} $\Theta$: optimized parameters.
   \FOR {$e=1$ {\bfseries to} $E$}
   \WHILE {not feeding the DNN all data in $\mathbb{D}$}
   \STATE $\mathcal{G} = 0$
   \FOR {$b=$ 1 {\bfseries to} $B$}
   \STATE $\mathbb{X} = {generate\_random\_batch}(\mathbb{D},K)$ 
   \STATE $\mathbb{X} =  {dimension\_randomization}(\mathbb{X}, R)$ 
   \STATE $\hat{\mathbb{Y}} = {forward\_pass}(\Theta, \mathbb{X})$
   \STATE $\mathcal{G} = \mathcal{G} + Gradients\big(Loss(\mathbb{Y}, \hat{\mathbb{Y}})\big)$
   \ENDFOR
   \STATE $\Theta = {optimizer}(\Theta, \mathcal{G})$
   \ENDWHILE
    \ENDFOR
\end{algorithmic}
\end{algorithm}
 
\begin{figure}[t] 
    \centering 
    \includegraphics[width=.7\columnwidth]{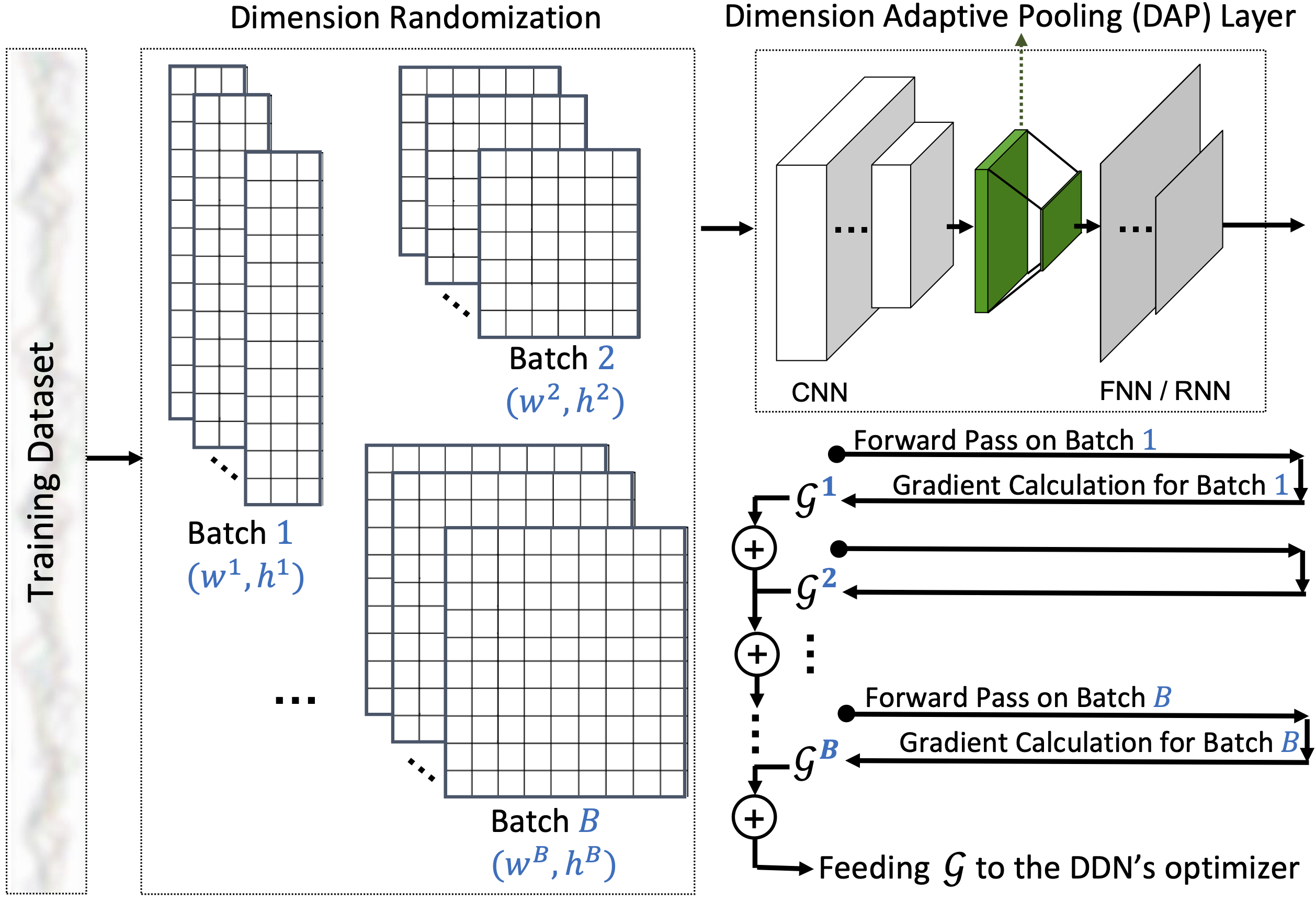}
    \caption{Overview of each iteration in DAT (Algorithm~\ref{alg:dat}). $B$ batches of time windows are randomly generated from a training dataset. All time windows in each batch are transformed into the same randomly chosen dimensions $w^b \times h^b$, thus samples within each batch have the same dimension while samples among batches have different dimensions. Every batch is iteratively fed into DNN and the corresponding gradients with respect to the current loss value are accumulated into $\mathcal{G}$. After computing the gradients of all $B$ batches, the parameters of the DNN are updated based on $\mathcal{G} = \sum^{B}_{i=1} \mathcal{G}^{i}$.}
    \label{fig:dat}
\end{figure}

Although a DNN using DAP can accept input of any dimensions, it should be properly trained for being adaptive to changes, otherwise, its classification accuracy will be low when the input's dimensions change.  {In particular, we cannot anticipate which sensors are available at inference time, and consequently, an adaptive model should be ready for such changes not only in the sampling rate but also in the type and order of receiving sensor streams. To this end, we leverage the capability of CNNs (discussed in Section~\ref{sec:related_work}) in being invariant to changes in the scale and location of observed patterns in the data. To cover this important requirement, we need a training procedure for adaptive DNNs that encounters the model with such potential changes in the type of sensor streams, that could happen at inference time, to make the model less sensitive to the changes in the scale, type, and order of incoming sensor streams.}

The {\em multi-size training} procedure~\cite{he2015spatial} for images trains in parallel two DNNs with different dimensions but with shared weights. This multi-size training  is problematic with sensor data because of the large  variety of possible training situations. For example, 10 sampling rates (e.g.~5, 10, 15, $\ldots$, 50 in Hz) and the 7 possible combinations of 3 sensors lead to 70 DNNs to train. To overcome this challenge and provide a reasonable, converging training strategy, we propose dimension adaptive training~(DAT). In DAT, we train a single DNN, and therefore there is neither a need for weight sharing nor for choosing a specific set of data dimensions. DAT comprises of two main ideas: {\em dimension randomization} and {\em optimization with accumulated gradients} (Fig.~\ref{fig:dat}).  {The DAT process works by training the DNN on input data of several randomly selected dimensions and sensor orders. In this way, DAT forces the model to leverage the underlying correlations between data streams coming from different motion sensors and learn dimension- and order-invariant features. Specifically, DAT randomly encounters convolutional filters with sensor patterns that appear in different locations and with different scales. Then, by accumulating and averaging the gradients computed on $B$ such random observations, DAT guides these convolutional filters toward achieving the capability of being invariant to potential changes at inference time.} For efficiency, the data is processed in batches of input data that has the same dimension. The details of DAT are shown in Algorithm~\ref{alg:dat}.

Each round of optimization includes two steps. First, a random batch, $\mathbb{X}$, of $K$ time windows of the highest available sampling rate is generated from the dataset. For batch number, $b$, the available sensors are  chosen randomly and a random sampling rate is chosen with data downsampled using bilinear interpolation. All the time windows in batch $b$ have the same dimensions\footnote{Note that, the key point of efficiently training DNNs on GPUs is in eliminating loops by matrix multiplications that force all samples in each batch, $b$, to have the same dimensions in a forward pass.} $w_b \times h_b\in \mathbb{U}$. Considering an example where the possible sampling rates range from 6~Hz to 50~Hz, instead of considering all the 45 possible cases, we only choose a subset $\mathbb{U}$, for instance including $8$ sampling rate \{6, 12, 18, 25, 31, 37, 43, 50\}~(in Hz). In the epoch $e$, $dimension\_randomization()$ in Line 8 randomly and uniformly chooses, for instance $B=4$ of this $8$ sampling rates without replacement. Second, a forward pass is performed on batch $b$ giving a vector of predictions, $\hat{\mathbb{Y}}$. we calculate the average loss value (\eg categorical cross-entropy) of the predictions compared with the true labels, $\mathbb{Y}$, and its gradients, corresponding to the $\Theta$, is accumulated into $\mathcal{G}$. These two stages are repeated $B$ times, then using the accumulated gradients, $\mathcal{G}$, the parameters the DNN, $\Theta$, are updated at once.

As DNNs tend to forget previously learned information upon learning from new data, updating $\Theta$ immediately after computing losses for each batch of data causes catastrophic forgetting~\cite{french1999catastrophic}, in which the DNN may repeatedly forget how to perform classification on the previously trained dimensions. We show that the combination of dimension randomization and gradient accumulation not only helps DAT to prevent catastrophic forgetting but also to converge better.  

It is worth noting that the gradient accumulation in DAT is similar but a different concept than the typical method of keeping track of gradient {\em momentum} in stochastic gradient descent~\cite{qian1999momentum}. In a momentum-based optimization, a scaled version of the gradients used in the previous round is added to the gradients of the current round. Thus, at each round, the DNN's parameters are updated based on the current gradients and the history of the past gradients. On the other hand, DAT is a procedure for computing the required gradients in a single round, hence we can use DAT and a momentum-based optimization, such as Adam~\cite{kingma2014adam}, together.

\begin{table}[t!]
\caption{Details of the four real-world datasets used in our evaluations. For all datasets, the original sampling rate is 50~Hz.}\label{tab:har_datasets}
\centering
\resizebox{\columnwidth}{!}{
    \begin{tabular}{lcccc}
         &\multicolumn{4}{c}{{Characteristics}} \\\cline{2-5}
         Dataset  & Sensors & Type of Activities & Number of Users & Device Position \\\cline{1-5}  
        \multirow{2}{*}{UCI-HAR\cite{anguita2013public}}
        & 
        \multirow{4}{*}{\shortstack{Accelerometer \\ Gyroscope}} &
        \multirow{2}{*}{\shortstack{\{Sit, Stand, Walk, Lie, \\ Stairs-Down, Stairs-Up\}}} &
        \multirow{2}{*}{30} &  
        \multirow{2}{*}{Chest Mounted} \\
    &&&&\\\cline{3-5} 
    \multirow{1}{*}{MobiAct\cite{vavoulas2016mobiact}}
    &
    & 
    \multirow{2}{*}{\shortstack{\{Sit, Stand, Walk, Jog, \\ Stairs-Down, Stairs-Up\}}}
    & 
    \multirow{1}{*}{61}
    &
    \multirow{1}{*}{Trousers Front Pocket}\\
    \cline{4-5}
    \multirow{1}{*}{MotionSense\cite{malekzadeh2018protecting}}
    &
    &
    &
    \multirow{1}{*}{24}
    &
    \multirow{1}{*}{Trousers Front Right Pocket} \\
    \cline{2-5}
    \multirow{3}{*}{UTwente\cite{shoaib2016complex}}
    &
    \multirow{3}{*}{\shortstack{Accelerometer \\ Gyroscope\\ Magnetometer}}
    &
    \multirow{3}{*}{\shortstack{\{Sit, Stand, Walk, Jog, Bike, 
    \\Smoke, Drink, Eat, Talk, Type,
    \\Write, Stairs-Down, Stairs-Up\}}}
    &
    \multirow{3}{*}{10}
    & 
    \multirow{3}{*}{Right Wrist}\\
    &&&&\\
    &&&&\\\hline
    \end{tabular}
}
\end{table}
\section{Evaluation}
\label{sec:eval}
 
\subsection{Evaluation Setup}\label{sec:eval_setup}
Since existing DNN architectures require fixed-dimension input data, the only DNN-based alternative to our proposed \dana is to add a data preprocessing stage to the inference pipeline that up/down-samples data to the fixed  sampling rate and imputes synthetic data to compensate for unavailable sensor streams. As discussed in Section~\ref{sec:intro}~and~\ref{sec:related_work}, our solution offers competitive advantages over this alternative, such as eliminating the need for preprocessing as well as adapting the required computations at inference time while having the same size as existing DNNs. In the following, we focus on {\em classification accuracy} as the main evaluation criterion for our comparisons. In particular, we measure the classification accuracy of \dana and the other alternative (which we refer to it as the {\em original} model) across a range of possible scenarios. In each scenario, the more a DNN maintains its average accuracy, despite changes in the dimension at inference time, the more robust the DNN is considered.

{We evaluate \dana on four public datasets of human activity recognition: {\em UCI-HAR}~\cite{anguita2013public}, {\em UTwente}~\cite{shoaib2016complex}, {\em MobiAct}~\cite{vavoulas2016mobiact}, {\em MotionSense}~\cite{malekzadeh2018protecting}. We show how to transform three state-of-the-art CNN-FNN/RNN architectures for sensor-based human activity recognition~\cite{ronao2016human,ignatov2018real,ordonez2016deep} into a \dana, and discuss how DAP works with and without using DAT. We also evaluate the advantages of \dana at training time, compared with three other training procedures: {\em standard}, {\em weight averaging}~\cite{he2015spatial}, and {\em Reptile}~\cite{nichol2018first}, and at inference time compared with two alternatives baselines: {\em imputation} and {\em resampling} with and without {\em data augmentation}. Finally, we perform an experiment across two datasets to show the generalization of \dana.}

{\bf UCI-HAR}~\cite{anguita2013public} is a widely used dataset~\cite{ronao2016human,ordonez2016deep,ignatov2018real, zhao2018deep} of 30 users performing 6 activities. Accelerometer and gyroscope data were collected by a smartphone worn on the waist. Data from 21 users are used for  training and that of the other 9 users for  testing purposes~\cite{anguita2013public}.

{\bf UTwente}~\cite{shoaib2016complex} includes data of 10 users performing 13 activities. We use accelerometer, gyroscope, and magnetometer data collected from the device on the (right) wrist. We  divide the dataset into 80\% training and 20\% validation.

{\bf MobiAct}~\cite{vavoulas2016mobiact} and {\bf MotionSense}~\cite{malekzadeh2018protecting} include accelerometer and gyroscope data from 61 and 24 users, respectively, with a smartphone in the pocket of the trousers. We use the data from the 6 activities in common among the two datasets. In the MobiAct dataset, for each activity, we keep 2/3 of trials for training and 1/3 of them for validation~\cite{malekzadeh2018protecting}. We use the entire MotionSense dataset for test purposes and do not use it in the training. For all experiments, we use a time window $T=2.56$ seconds (i.e.~a maximum of 128 samples per window)~\cite{ronao2016human, ignatov2018real, ordonez2016deep}.
Table~\ref{tab:har_datasets} shows the details of the datasets, including the activity classes.

\subsection{Transforming a DNN into \dana}
%
\begin{figure}[t!] 
    \centering
    \includegraphics[width=5cm]{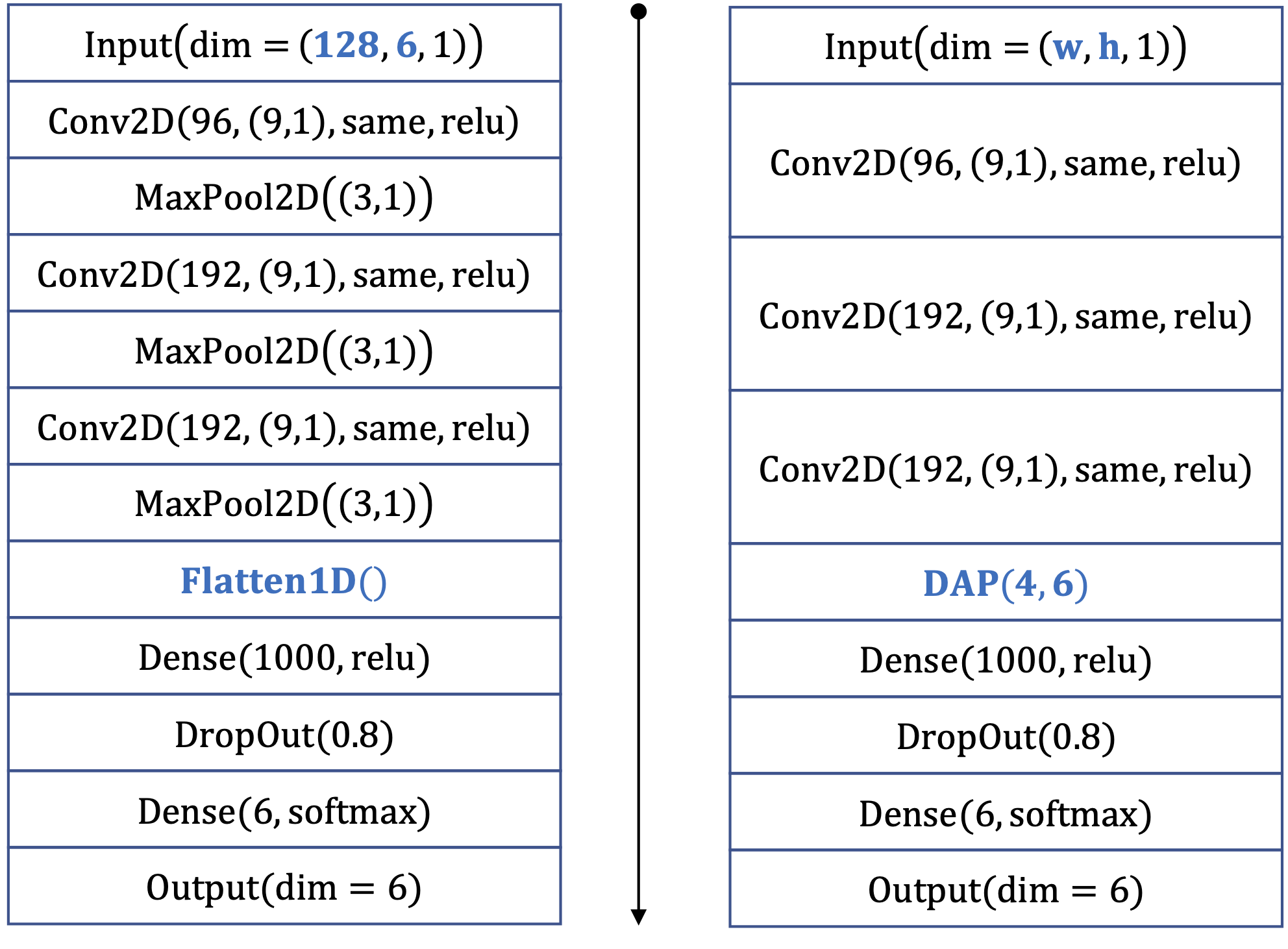}
    \hfill
    \includegraphics[width=5cm]{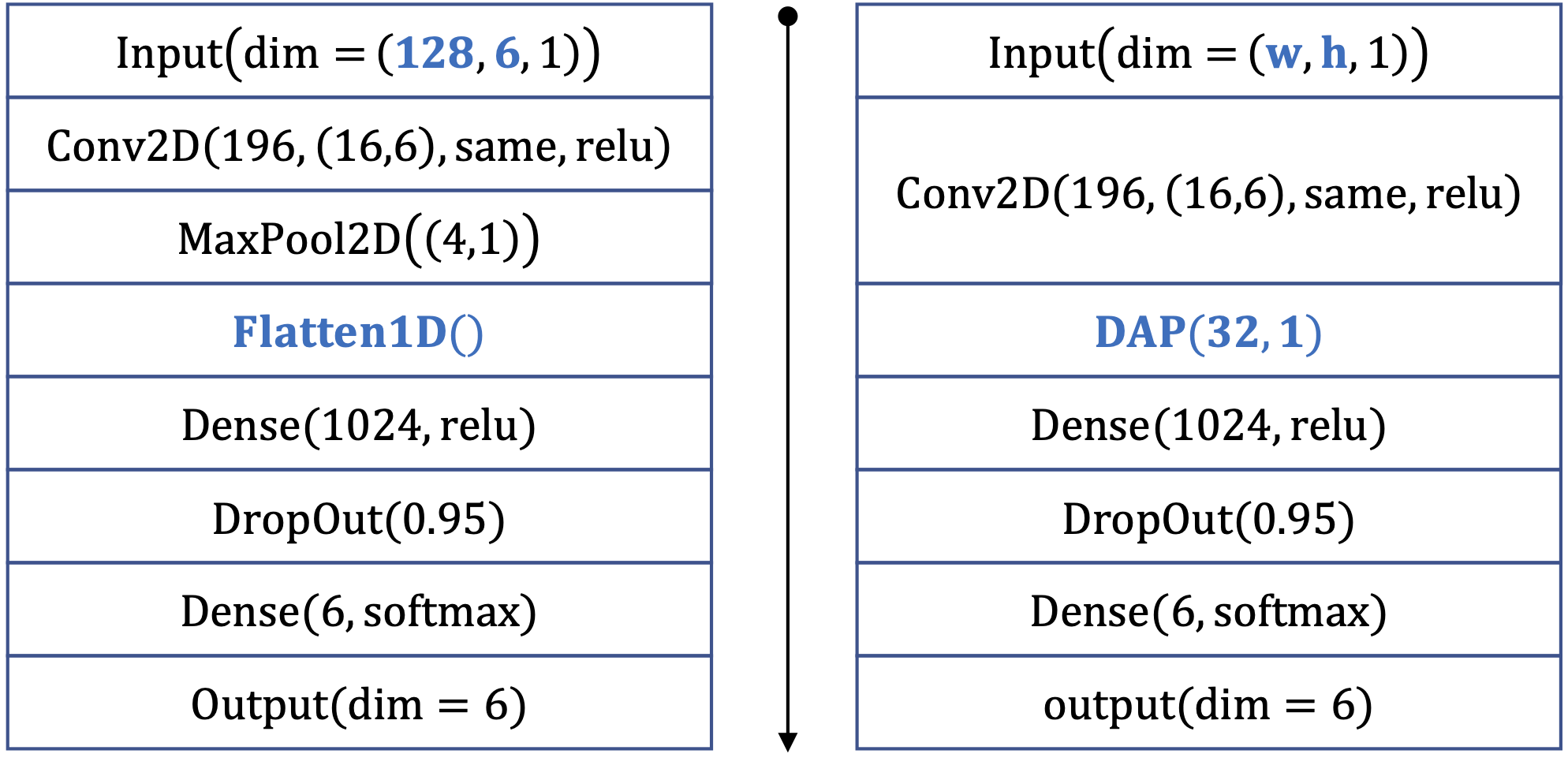}
    \hfill 
    \includegraphics[width=5cm]{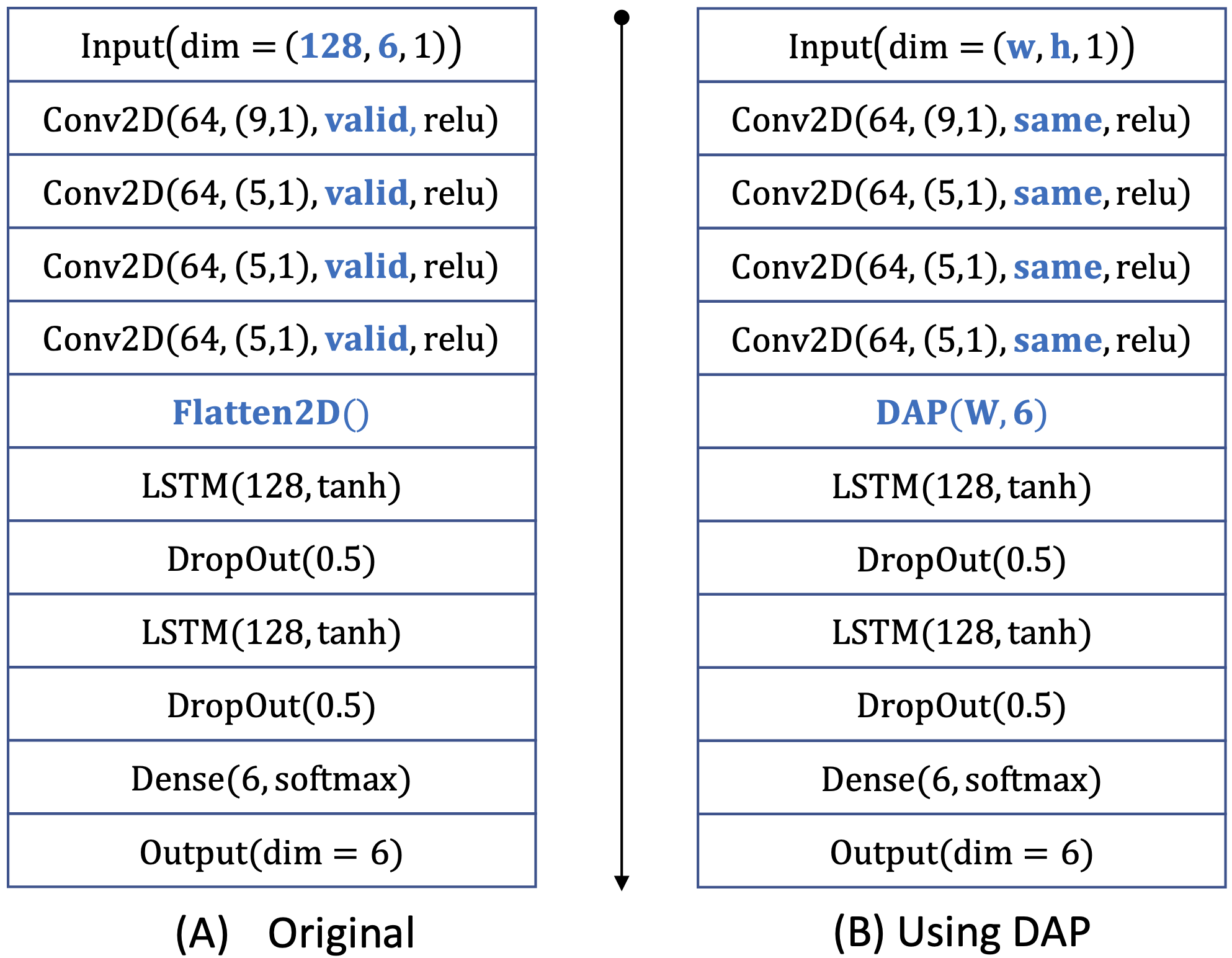} 
    \par\medskip{(A) \qquad\qquad\qquad\qquad\qquad\qquad\qquad (B) \qquad\qquad\qquad\qquad\qquad\qquad\qquad (C) }\par\medskip
    \caption{The original DNNs  (left plots) proposed in (A)~\cite{ronao2016human}, (B)~\cite{ignatov2018real}, and  (C)~\cite{ordonez2016deep}, versus their \dana version (right plots). The differences between each pair of DNNs are shown in blue bold font. We use TensorFlow's naming conventions~\cite{abadi2016tensorflow}.}  
    \label{sup_mat_fig:dnn_1_ronao}
\end{figure}

Fig.~\ref{sup_mat_fig:dnn_1_ronao} compares the original architectures proposed in the related work~\cite{ronao2016human,ignatov2018real,ordonez2016deep} and our modified version using DAP. A 2D convolutional layer is shown by $\mathrm{Conv2D}(n,(k_1,k_2), padding, activation)$ where $n$ is the number of neurons, $(k_1,k_2)$ is the size of the 2D kernel used by each neuron; if $padding$ is $same$, the output and the input of the layer have the same dimensions, otherwise if it is $valid$ then the output has $k_1-1$ and $k_2-1$ values fewer than the number of inputs in the first and second dimensions, respectively; $activation$ shows the nonlinear activation function  applied to the output of the layer; $\mathrm{Dense}(n,activation)$ shows a fully-connected feedforward layer;  $\mathrm{LSTM}(n, activation)$ shows a recurrent LSTM~\cite{hochreiter1997long} layer;  $\mathrm{MaxPool2D}\big((p_1,p_2)\big)$ shows a 2D maximum pooling layer that outputs the maximum value of each window including $p_1 \cdot p_2$ data points; $\mathrm{Dropout}(q)$ applies Dropout~\cite{srivastava2014dropout} with probability $q$  that randomly sets the output of a neuron to 0 during each forward pass in training time; ${\mathrm{FlattenXD}()}$ reshapes its inputs into 1D or 2D outputs.

To transform a non-adaptive DNN into a \dana, we only use a single DAP layer, instead of ``maximum pooling`` layers, with appropriate pooling parameters $(W,H)$, and instead of ``valid'' padding, we use ``same'' padding. Thus, we can keep the total number of trainable layers and parameters exactly the same as the original DNN.

\subsection{Original DNNs versus Their DANA Version}

\begin{table}[t]
\caption{Classification accuracy of three benchmark DNNs (see Figure~\ref{sup_mat_fig:dnn_1_ronao}) on UCI-HAR dataset, the original non-adaptive models proposed in \cite{ronao2016human}, \cite{ignatov2018real}, and \cite{ordonez2016deep}, versus the corresponding adaptive model with a DAP layer. Notice that we set $(W,H)$ such that the transformed architecture has the same model size as the original DNN. The results, where the sampling rate is fixed to 50Hz and both sensors are available, show that making these DNN flexible, by our proposed DAP layer, does not lead to a loss in the DNN's accuracy.}
\label{tab:uci_har}
\begin{tabular}{lrlccc}
 & & & & \multicolumn{2}{c}{Accuracy~(\%)} \\\cline{5-6}
 Architecture & Size & {DNN [$(W,H)$]} & Setting & {Mean$\pm$STD} & Maximum \\\toprule
 
\multirow{4}{*}{CNN-FNN} & \multirow{4}{*}{5,114,014} & \multirow{2}{*}{\cite{ronao2016human}} & \eval & {93.85$\pm$.49} & 94.51 \\
&&& \etest &91.92$\pm$.97& 93.14\\\cline{3-6}
 & & \multirow{2}{*}{with DAP(4,6)} & \eval & {93.73}$\pm$.84 & 95.39 \\
&&& \etest &91.88$\pm$1.1 & 93.04\\\hline 

\multirow{4}{*}{CNN-FNN} & \multirow{4}{*}{6,448,714} & \multirow{2}{*}{\cite{ignatov2018real}} & \eval & 93.80$\pm$.59 & 94.78 \\
&&& \etest &92.75$\pm$.73& 94.02\\\cline{3-6}
&& \multirow{2}{*}{with DAP(32,1)} & \eval & {93.89}$\pm$.41 & 94.57 \\
&&& \etest &92.57$\pm$.60& 93.65\\\hline 
 
\multirow{10}{*}{CNN-RNN} &
\multirow{10}{*}{457,030} &
\multirow{2}{*}{\cite{ordonez2016deep}} & \eval & 94.13$\pm$.53 & 95.25 \\
&&& \etest & 91.66$\pm$1.3& 93.45\\\cline{3-6}
&& \multirow{2}{*}{with DAP(4,6)} & \eval & 94.69$\pm$.31 & 95.32 \\
&&& \etest & 94.07$\pm$.39& 94.77\\\cline{3-6}
&& \multirow{2}{*}{with DAP(8,6)} & \eval & 94.82$\pm$.34 & 95.32 \\
&&& \etest & 93.47$\pm$.81& 94.33\\\cline{3-6}
&& \multirow{2}{*}{with DAP(16,6)} & \eval & 94.69$\pm$.48 & 95.18 \\
&&& \etest &93.19$\pm$.59& 94.02\\\cline{3-6}
&& \multirow{2}{*}{with DAP(32,6)} & \eval & 94.74$\pm$.33& 95.32\\
&&& \etest & 93.23$\pm$.92& 94.70\\
\bottomrule
\end{tabular}
\end{table}

To ensure the comparison is fair among all DNNs, we use the same number of epochs~(1,000), early stopping patience~(100 epochs), and batch size~(128). We also use the same optimizer as reported in the corresponding works: for the models with FNN, Adam~\cite{kingma2014adam}, and for the models with RNN, RMSProp~\cite{tieleman2012lecture}. We run each model 10 times and report the mean and standard deviation for classification accuracy. 

Table~\ref{tab:uci_har} compares the classification accuracy for each DNN architecture with that architecture using DAP instead of their standard maximum pooling. Although we set $(W,H)$ such that the transformed architecture has the same model size as the original DNN; it is possible to choose any other values getting a larger or smaller size DNN. Here, two settings are considered for evaluations. (i) {\em Validate}: where the whole training dataset is used for training, and the test dataset is used for validation. Thus, each DNN is trained on the training dataset and the validation dataset is used to check the accuracy of the model after each epoch. This is the setting that is used by other works re-implemented~\cite{ronao2016human,ignatov2018real}. (ii) {\em Test}: where 10\% of the training dataset is randomly chosen as the validation set and the rest (90\%) is used as the training set. Here, the test dataset is used to evaluate the best trained DNN on the validation set. 

Table~\ref{tab:uci_har} shows that for all three architectures the classification accuracy of the adaptive version is either slightly improved or almost the same as the original DNN, in both {\em validate} and {\em test} settings. This suggests that the DAP layer does not lead to a loss in accuracy while providing the desirable adaptivity, that is further utilized by DAT. The CNN-RNN model has 10 times fewer parameters than the FNN but has a better classification accuracy. The CNN-RNN with DAP significantly improves accuracy in the {\em test} setting, thus suggesting that DAP helps the model to better generalize to the test data.  RNNs are also flexible on their first dimension, which is the number of samples per stream. Thus, unlike FNNs, we can use DAP layers having different pooling parameters for the first dimension while keeping the rest of the architecture the same. Changing the pooling parameters leads to slightly different accuracies, which can be used as part of a fine-tuning pipeline for DNN models but would not be possible with FNNs.

\begin{table}[t!]
\caption{Classification accuracy~(\%) of three benchmark DNNs on the UTwente dataset (that does not use in any of the original papers~\cite{ronao2016human}, \cite{ignatov2018real}, and \cite{ordonez2016deep}). Results show that the DNN using RNN layers~\cite{ordonez2016deep} offers better generalization than the other two DNNs using FNNs. Similar to Table~\ref{tab:uci_har}, we see that the adaptive version of~\cite{ordonez2016deep}, using our DAP layer, achieves a similar, and even slightly better, classification accuracy.
\label{tab:UTwente_FNN_RNN}}
\begin{tabular}{lrlcc}
 & & &  \multicolumn{2}{c}{Accuracy} \\\cline{4-5}
DNN & Size & {Ref.} &  {Mean$\pm$STD} & Maximum \\\toprule
\multirow{2}{*}{CNN-FNN} & 7,425,021 & \cite{ronao2016human} &  59.20$\pm$9.6& 76.99 \\\cline{2-5} 
& 12,878,417 & \cite{ignatov2018real} &  78.85$\pm$6.2& 88.01 \\\hline
\multirow{3}{*}{CNN-RNN} & \multirow{3}{*}{556,237} & \cite{ordonez2016deep} & 93.68$\pm$.36 & 94.55 \\\cline{3-5}
& & {with DAP(8,9)} & 94.35$\pm$.37 & 95.16 \\
& & {with DAP(16,9)} & 94.64$\pm$.40 & 95.40 \\
\bottomrule
\end{tabular}
\end{table}

To see the performance of the DNNs and \dana on another dataset, we evaluate existing DNNs (without fine-tuning) on UTwente, which is a different dataset from those datasets used in the original papers of the implemented DNNs.  Table~\ref{tab:UTwente_FNN_RNN} shows that the CNN-RNN architecture~\cite{ordonez2016deep} has better generalization than the other two CNN-FNN architectures. The model using DAP maintains a comparable accuracy. 

We see that transforming a DNN into \dana does not change the trainable parameters of the DNN, whereas using other global pooling or SPP layers would require a change in the size of the FNN/RNN layers and consequently the number of trainable parameters of the original DNN. As an experiment, to measure the effect of using the global average pooling layer~\cite{lin2013network}, we run a similar experiment on CNN-RNN~\cite{ordonez2016deep} with UCI-HAR. The model size is reduced from 457K to 293K, but also the classification accuracy~(\%) reduced from $94.13\pm.53$ to $78.6\pm.81$. 

It should be noted that this experiment only aims to show that using a DAT layer does not degrade the accuracy, and not to show that using a DAT layer we can achieve better accuracy in a fixed-dimensions scenario. As we show in other experiments, the main advantage of \dana is keeping high accuracy in variable-dimensions scenarios. As the {\em validate} setting is what is considered in other works~\cite{ronao2016human,ignatov2018real}, where they report the {\em best accuracy} that corresponding DNN can achieve on the dataset, for the rest of the experiment we use the {\em validate} setting for fair comparisons. Note that as we do not tune any hyper-parameters and do not change the size of trainable parameters in all DNNs, similar relative results are achieved in the {\em test} setting. But, unlike the {\em validate} setting which has only one instance (\ie a training set and a validation set), the {\em test} setting could be biased because there are many possible instances depending on the randomly chosen 10\% validation set.

\subsection{Changes in Sampling Rate and Sensor Availability}

\begin{figure}[t!]
    \begin{minipage}[t]{.49\columnwidth}
    \includegraphics[width=\columnwidth]{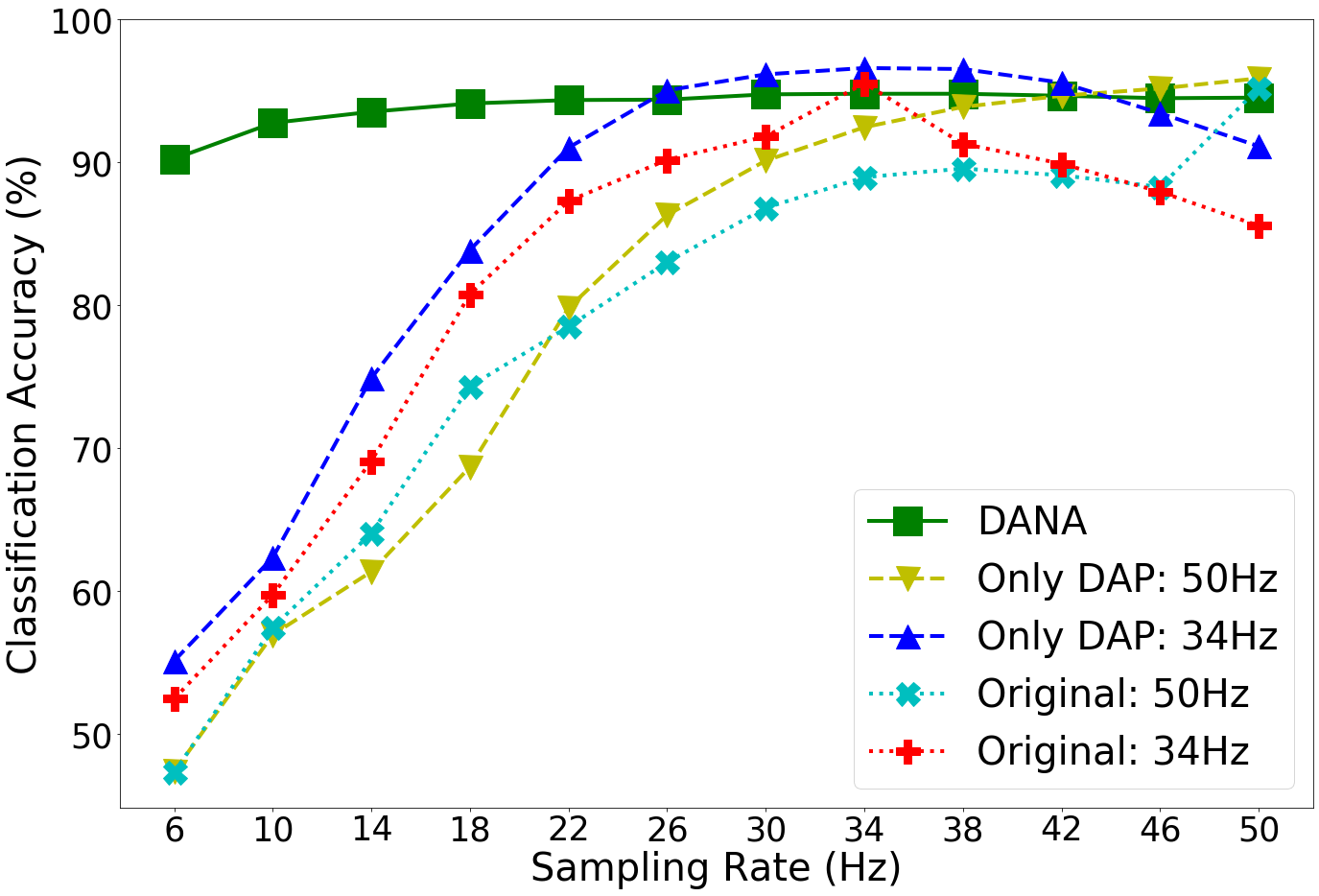}
    \caption{Classification accuracy of CNN-RNN~\cite{ordonez2016deep} on UCI-HAR dataset for varying sampling rates (but fixed available sensors). \dana keeps the accuracy above 90\% in all situations; due to DAT training. Only sing a DAP layer, without training with DAT, achieves a similar performance to the original DNN. We evaluated the original DNN, on all fixed-sampling rates from 6Hz to 50Hz, and found 34Hz proposing the maximum test accuracy.\label{fig:uci_lstm_vsr}}
    \end{minipage}
    \hfill 
    \begin{minipage}[t]{.47\columnwidth}  
    \includegraphics[width=\columnwidth]{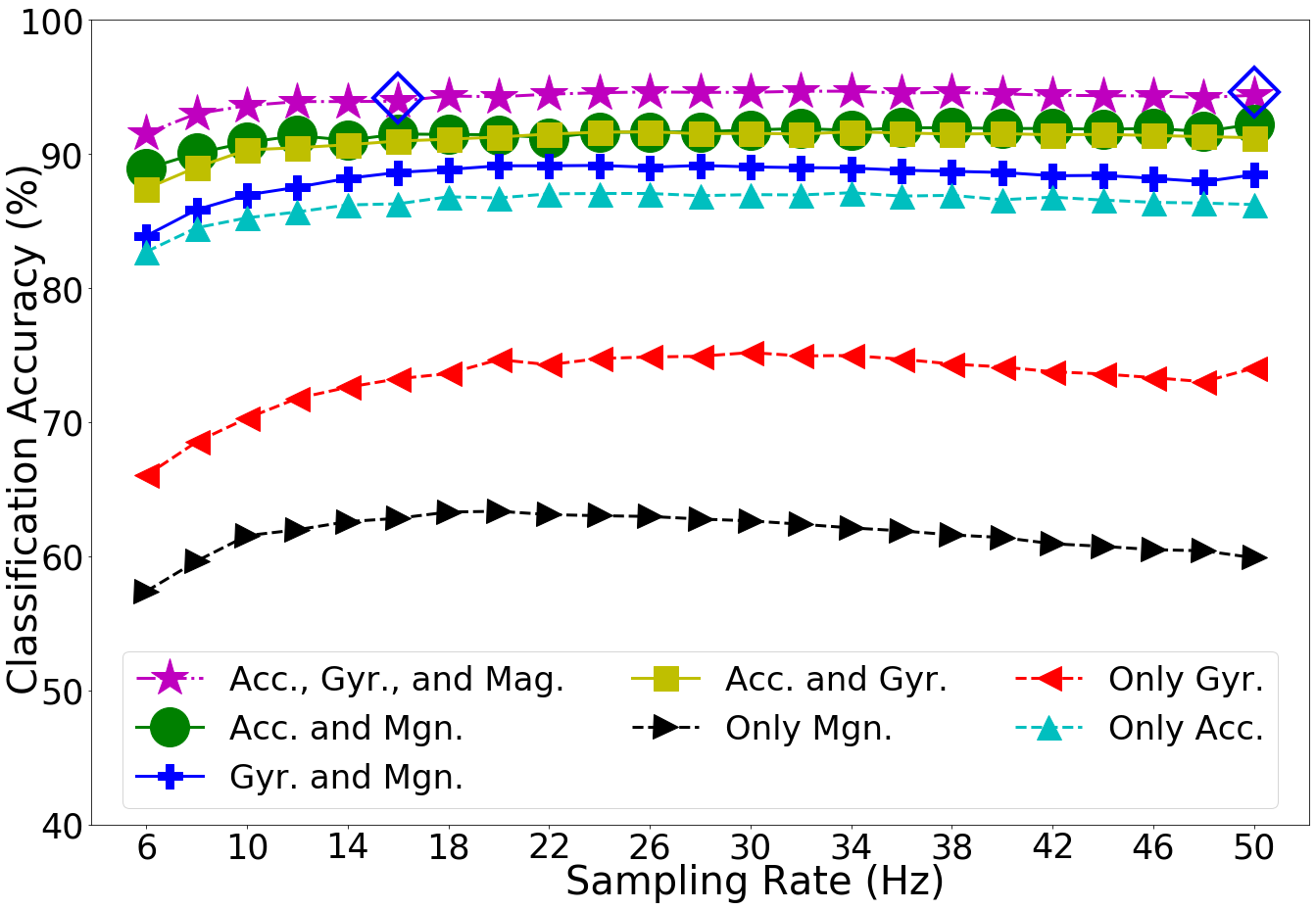}
    \caption{Classification accuracy of \dana with DAT for both dimensions} on UTwente dataset with the \dana version of the CNN+RNN proposed by~\cite{ordonez2016deep}. The two points, shown by \textcolor{blue}{\textbf{$\Diamond$}} at 16~Hz and 50~Hz, show the accuracy of the original DNN when  trained on these fixed sampling rates.\label{fig:utw}
    \end{minipage}
\end{figure}

Fig.~\ref{fig:uci_lstm_vsr} compares the original and \dana version of the DNN proposed by~\cite{ordonez2016deep}, as CNN-RNN performs better. First, using the original DNN, the model can only be trained and validated on data of fixed dimensions. To see how the original model performs if we change the sampling rate, we keep the weights and parameters of the original model and use them on a version using DAP (lines with a cyan cross and red plus). Second, we have a DNN version that only uses DAP during the training but is only trained on a fixed sampling rate (lines with yellow and blue triangles). Finally, we have the \dana version (the line with green square) which not only uses DAP, but it also uses DAT with $B=4$ and $\mathbb{U}$=\{6, 12, 18, 25, 31, 37, 43, 50 \}~(in Hz). 

Although \dana version slightly reduces the accuracy on the specific sampling rates the other DNN versions were trained on~(34Hz and 50Hz), it considerably outperforms the other DNN versions across the whole range of sampling rates, where the accuracies of other DNNs drops quickly when moving away from their pre-defined settings. Because the best accuracy of \dana occurs at 34~Hz, for this reason, we trained other DNNs on this specific sampling rate, thus suggesting that DAT is consistent in its performance while being adaptive. 

\begin{figure}[t!]
    \centering
    \includegraphics[width=\columnwidth]{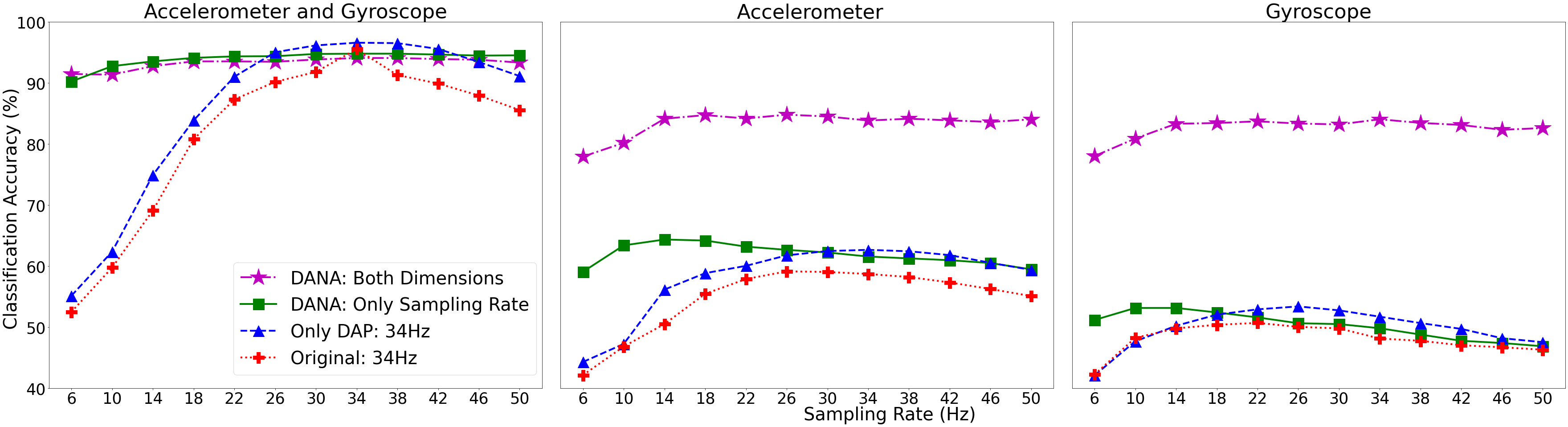}
    \caption{Classification accuracy of CNN-RNN~\cite{ordonez2016deep} on UCI-HAR dataset with a variable sampling rate and sensor availability. When both sensors are available (left), versus when only the accelerometer (middle) or gyroscope (right) is available. Overall, with DAT for both dimensions (\textcolor{purple}{$\star$}) performs much better than DANA with DAT for only sampling rate (\textcolor{teal}{$\blacksquare$}).} 
    \label{fig:uci_lstm_vsvsr} 
\end{figure}

To see the effect of sensor selection, Fig.~\ref{fig:uci_lstm_vsvsr} shows the replication of the previous experiment (in Fig.~\ref{fig:uci_lstm_vsr}) in different sensor-selection scenarios. Here, \dana is trained to account for the possibility of deselecting (or missing) sensors, and the other lines are the same as in Fig.~\ref{fig:uci_lstm_vsr} where the training only considers sampling rate selection and does not account for sensor selection. During training, 50\% of the time, \dana is trained on both sensors, and 50\% of the time with one of accelerometer or gyroscope (randomly chosen with equal probability). The values of $B$ and $\mathbb{U}$ are the same as Fig.~\ref{fig:uci_lstm_vsr}.

Fig.~\ref{fig:uci_lstm_vsvsr}~(left) shows that the adaptivity to the sensor selection is not associated with a large penalty when all sensors are present.   Fig.~\ref{fig:uci_lstm_vsvsr}~(middle) shows that when the gyroscope is deselected, \dana maintains its accuracy around 85\% while the accuracy of other DNNs falls rapidly to around 60\%. Similarly, Fig.~\ref{fig:uci_lstm_vsvsr}~(right) shows that when the accelerometer has deselected the accuracy for \dana remains around 85\% while other DNNs fall to 50\% or less. It is interesting that while for other DNNs deselecting accelerometer data causes more accuracy loss than deselecting gyroscope, the type of the deselected sensor has a reduced effect on \dana.


In Fig.~\ref{fig:utw} we make the CNN-RNN model adaptive to both sampling rate and sensor availability using DAP and DAT on UTwente, and we achieve similar accuracy to the original model. Note that, while original DNNs are not reliable when the data dimensions change, \dana provides at least 55\% accuracy across 7$\cdot$45=315 feasible data dimensions.

\begin{figure}[t]
    \begin{minipage}[t]{.52\columnwidth}
    \includegraphics[width=\columnwidth]{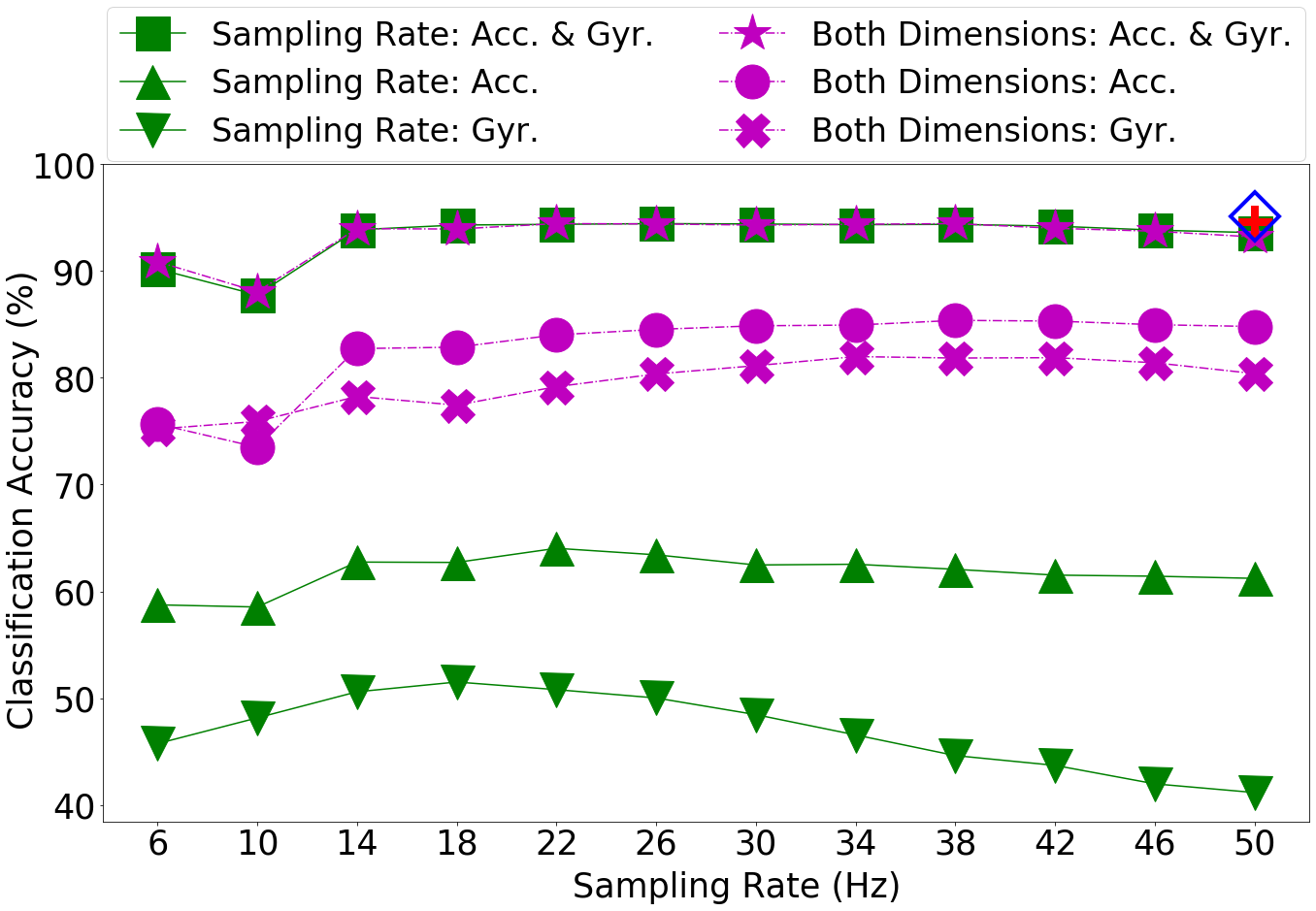}
    \caption{Classification accuracy of \dana on UCI-HAR  using a CNN-FNN~\cite{ronao2016human}. The  data points \textcolor{red}{\textbf{+}} 
    and \textcolor{blue}{\textbf{$\Diamond$}} at 50~Hz refer to Table~\ref{tab:uci_har} accuracies.  \label{fig:ronao}}
    \end{minipage}
    \hfill
     \begin{minipage}[t]{.4\columnwidth} 
    \includegraphics[width=\columnwidth]{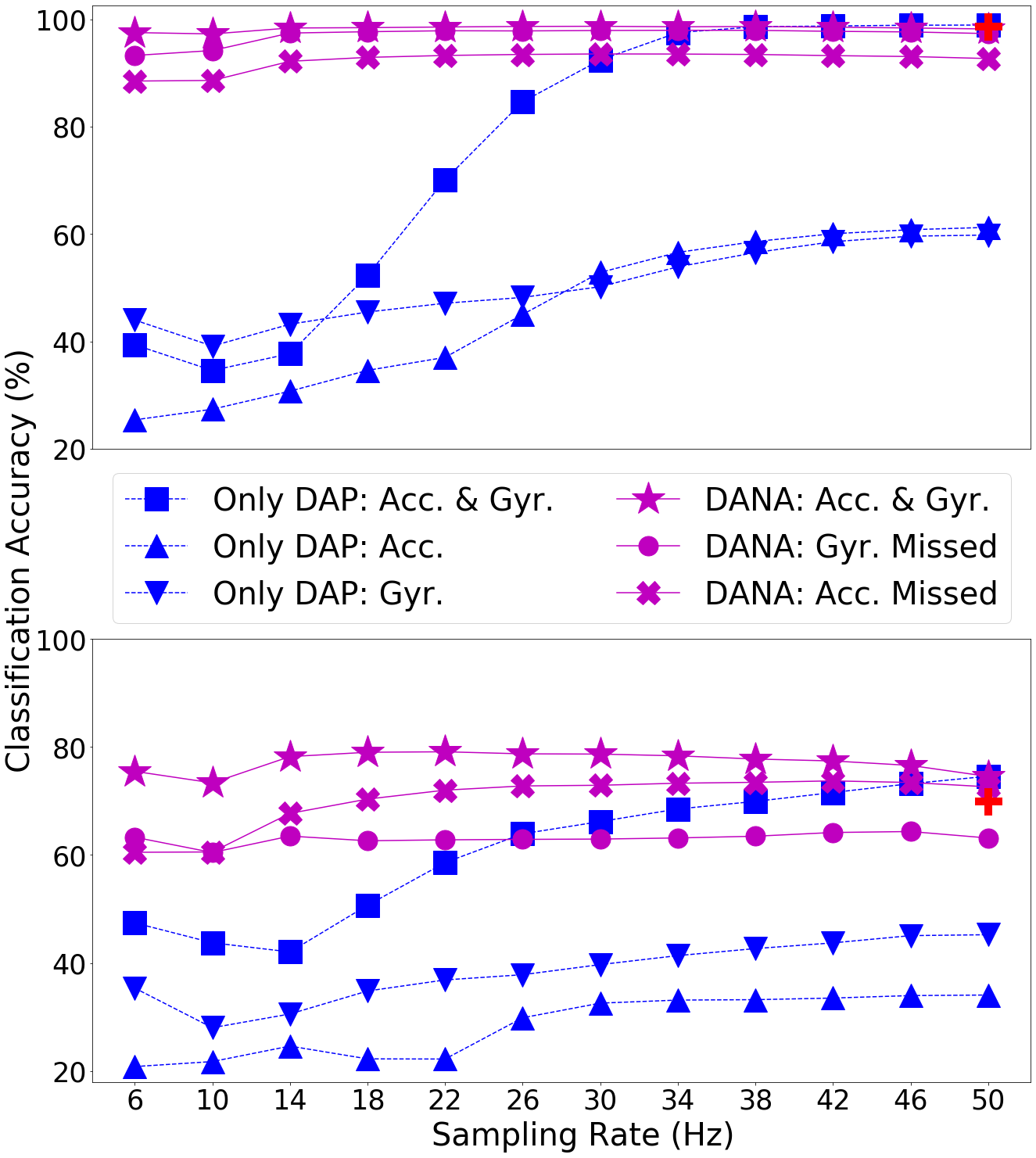}
    \caption{Classification accuracy of a CNN+RNN~\cite{ordonez2016deep} trained on MobiAct~\cite{vavoulas2016mobiact}  and tested on MotionSense~\cite{malekzadeh2018protecting}. The single data point shown by \textcolor{red}{\textbf{+}} at 50~Hz refers to the accuracy achieved with the original DNN in Table~\ref{tab:mobi_sense}.\label{fig:ma_ma} } 
     \end{minipage}
\end{figure}

Fig.~\ref{fig:ronao} compares the accuracy for different versions of a CNN-FNN proposed by~\cite{ronao2016human}. The  accuracy of \dana, trained on both variable sampling rates and variable sensors, remains high for a variety of sampling rates. When one of two sensors is deselected, the accuracy loss is much smaller than the one obtained from training only on variable sampling rates, with fixed sensors. The accuracy loss for this generalization is small when compared with the single points that show the situations where the DNN is only trained on a fixed sampling rate with all sensors present without and with DAP, respectively.

\subsection{A Cross-Dataset Experiment} 

\begin{table}[t]
\caption{Accuracy (\%) of CNN-RNN architecture proposed in~\cite{ordonez2016deep} trained on the MobiAct training dataset and validated on the MobiAct and MotionSense test datasets.}
\label{tab:mobi_sense}
\begin{center}
\begin{tabular}{lccc}
\multirow{2}{*}{DNN} &  \multirow{2}{*}{MobiAct} & \multicolumn{2}{c}{MotionSense} \\\cline{3-4} 
&& Normalized & Pseudo-Normalized\\\toprule
Original & 98.64 & 69.87 & 44.16 \\
Only~DAP & {98.91} & {74.65} & {49.65} \\ 
DANA & 98.18 & 74.60 & 47.97
\\\bottomrule
\end{tabular} 
\end{center}
\end{table} 

We train three versions of the CNN-RNN proposed in~\cite{ordonez2016deep} on MobiAct and test them on MotionSense.  We follow two settings: {\em normalized}, where we normalize the MotionSense data to  zero mean and unit standard deviation, and {\em pseudo-normalized}, where we use the MobiAct statistics to normalize the MotionSense data to mean zero and unit standard deviation. The latter is standard practice as when streaming data, the mean and standard deviation are not known in advance. ll three models have the same number of parameters and were trained under the same training setting.

Table~\ref{tab:mobi_sense} shows that \dana generalizes better on the test dataset, in terms of accuracy, by about 5 percentage points. This result confirms that for the corresponding datasets, using DAT to make a DNN adaptive helps to  achieve a more accurate model in an unseen environment\footnote{MobiAct was collected with a Samsung Galaxy S3 in the right or left pocket of the trousers. MotionSense was collected with an iPhone 6s in the front, right pocket of their trousers.}.  {Although the \dana version has a slightly lower accuracy than the one using only DAP, \dana is the only reliable model when the dimensions of the input data change (see Fig.~\ref{fig:ma_ma}). Thus, while \dana shows a bit smaller accuracy than the original DNN on MobiAct, this will considerably pay off when \dana is used in dynamic settings which we have shown in the results of the previous experiments. }

\begin{figure}[t!]
    \centering
    \includegraphics[width=.85\columnwidth]{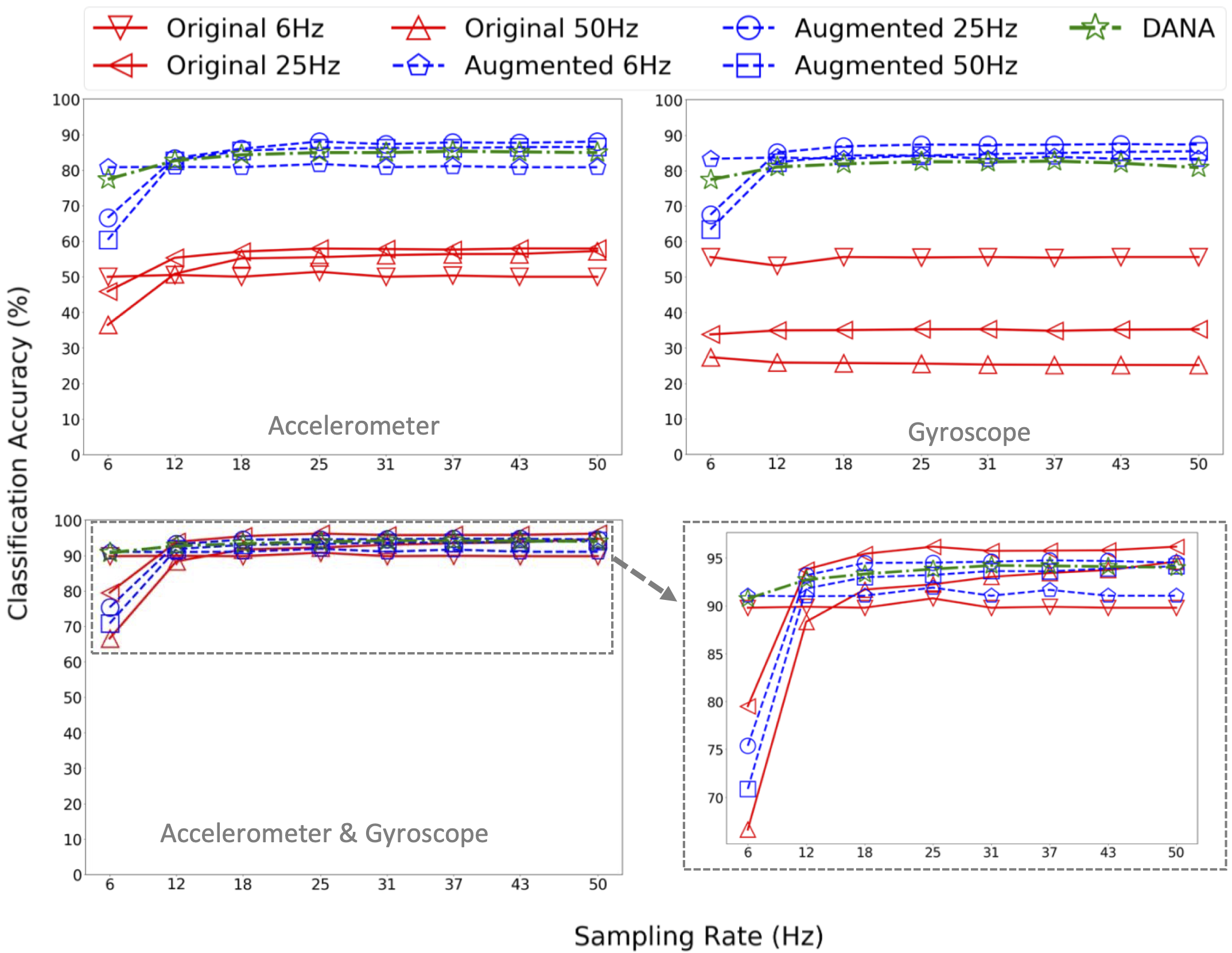}
    \caption{Comparing \dana with two baselines approaches using CNN-RNN~\cite{ordonez2016deep}, where  the non-adaptive DNN is trained on the UCI-HAR dataset as it is ({\em Original}), or via an extended version of UCI-HAR dataset by making a copy of every sample for all possible combinations of sensors availability ({\em Augmented}). Each non-adaptive DNN is trained on three fixed sampling rates (6, 25, and 50Hz). For situations at inference time where incoming data has a different sampling rate, we use down/up-sampling to the fixed sampling rate used in training time.}
    \label{fig:baselines_uci}
\end{figure}

\begin{figure}[t!]
    \centering
    \includegraphics[width=.79\columnwidth]{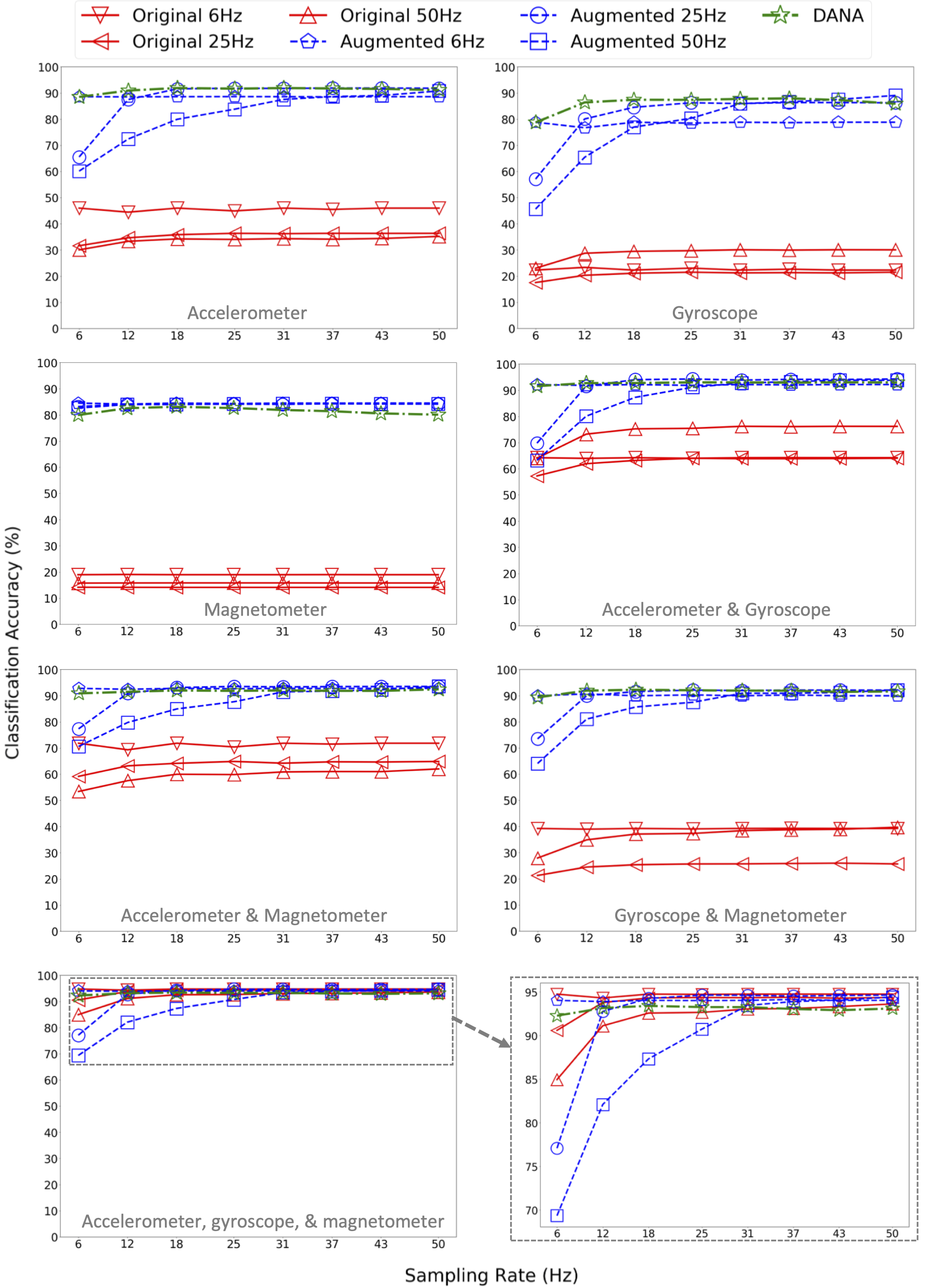}
    \caption{Comparing \dana with two baselines using CNN-RNN~\cite{ordonez2016deep}. Everything is similar to the experiment in Fig.~\ref{fig:baselines_uci}, except the dataset that is UTwente that contains three sensors.}
    \label{fig:baselines_ut}
\end{figure}

\subsection{Comparison with Baselines}~\label{subsec:baselines}

\begin{figure}[t!]
    \centering
    \includegraphics[width=.7\columnwidth]{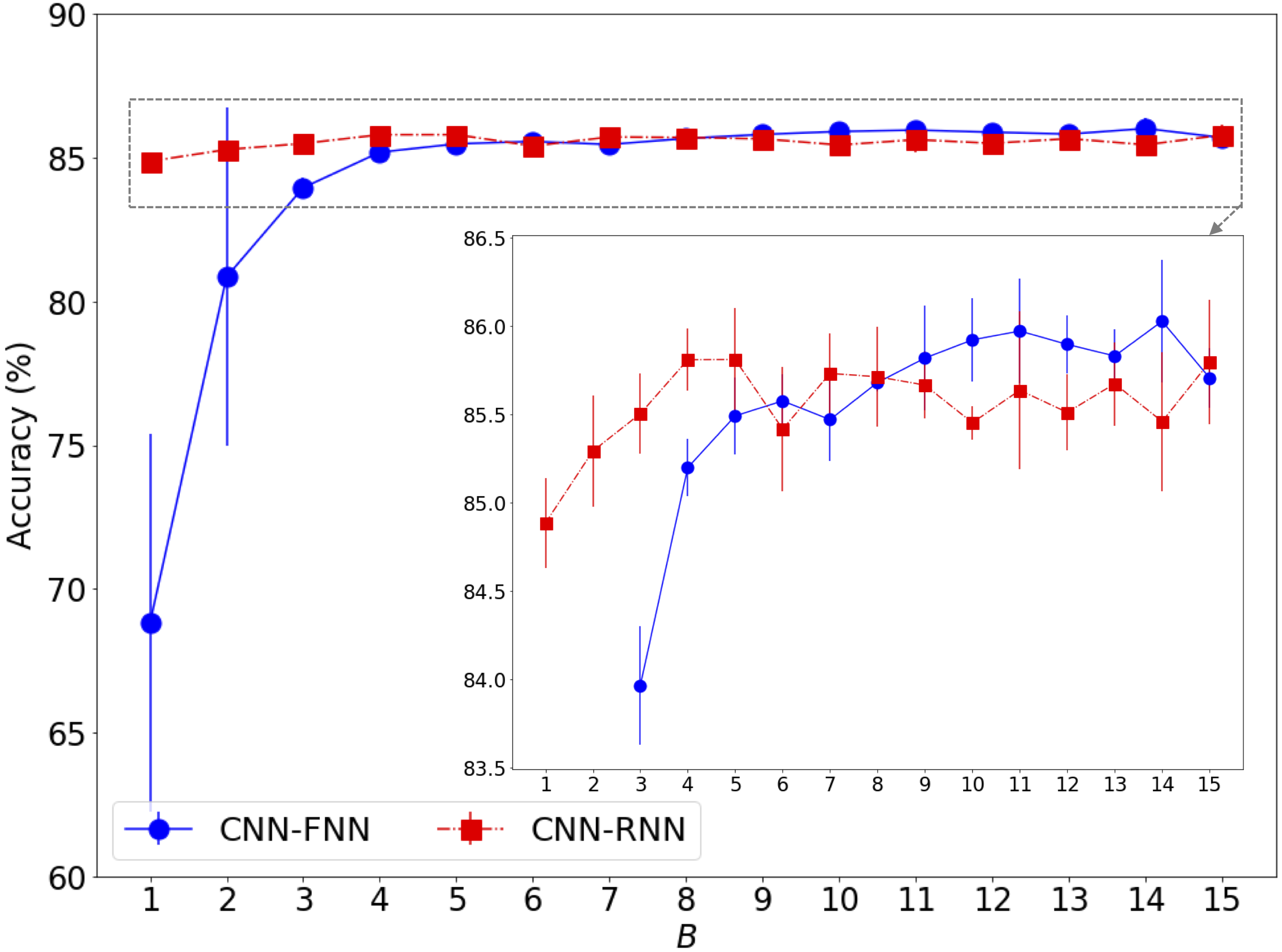}
    \caption{Effect of the size of resampling batches $B$ on UCI-HAR with CNN-RNN~\cite{ordonez2016deep}. Each point is the average and each segment shows the standard deviation for 5 runs.}
    \label{fig:different_B}
\end{figure}
 
As we discussed earlier on in Section~\ref{sec:eval_setup}, an alternative baseline to \dana is to take the original model, and every time the sampling rate is changed, we fix the problem by re-sampling data to the original sampling rate, and if some sensors are missing or deselected, we impute synthetic or dummy data, \eg inserting zeros, using the {\em mean values} for each stream computed on the training dataset, or just {\em copying} the data of available sensor(s). Moreover, when choosing the copying strategy, one can also use an {\em augmented} training dataset by making a copy of the data for every possible combination of sensors availability. For example, with 3 sensors, we augment the training dataset with 6 other time windows per each sample time window in the training dataset. For example, assume that we only have accelerometer data available at inference time, among three possible sensors. And, thus we add sample time windows to the training dataset that includes accelerometer data copied two more times to cover the other two unavailable sensors. Notice that, the training time for the {\em augmented} case is $2^s$ times the training time for the \dana (where $s$ is the number of sensors).  

Fig.~\ref{fig:baselines_uci} and \ref{fig:baselines_ut}  compares \dana with these baselines. We see that \dana can train a single model that outperforms the original model across all 56 settings, and also outperforms the augmented case for lower sampling rates (while  remaining competitive for larger sampling rates). Note that the augmented baseline needs a longer training time and also does not reduce the computations when sampling rate changes or some sensors are unavailable or deselected. These results show that \dana can capture the correlation, or information redundancy, between different sensor streams. For example, when the original models miss a sensor or two, their accuracies considerably fall, because they are not able to substitute the missed information using the available ones. But \dana keeps the accuracy at the same level as the augmented one, while it does not need to endure the difficulties of the augmented one at the training and inference time.

\subsection{Comparison with Other Training Approaches}

We explore the impact of the value of the training hyper-parameter $B$ on the accuracy. Fig.~\ref{fig:different_B} shows, for both CNN-FNN and CNN-RNN, the classification accuracy is comparable for RNN for $B \geq 4$ (whereas FNN cannot be trained with these values), and in general it suggests setting a $B \geq 5$.

Table~\ref{tab:training_methods} shows that DAT either outperforms or achieves comparable accuracy to other methods in several settings, and is always faster in training. 
The  training approaches we compare with are {\em Standard}, the typical training procedure that includes a forward pass on a batch of data to calculate loss value, following with a backward pass to update model parameters based on the  gradients; {\em WeightAvg}~\cite{he2015spatial, bonawitz2019towards}, when each of the $B$ copies of the model is trained on a dedicated batch of data using the {\em Standard} approach, and then the average parameter values of updated $B$ copies will be used to update the central model; and {\em Reptile}~\cite{nichol2018first}, which performs the same procedure as {\em WeightAvg}, except the last step when the average parameters of $B$ copies, each trained on a different batch of  data, is fed to the chosen DNN optimizer, instead of the typical gradients of the loss function, to update the central model. The value of $B$ is fixed to 5 for all experiments in Table~\ref{tab:training_methods}. The {\em time} shows the average training time per each iteration over the whole training dataset~(\ie one epoch). Note that, DAT is not sensitive to the chosen optimizer, or dataset, or DNN architecture. Also, Reptile works with Adam (that is mentioned in the original paper~\cite{nichol2018first}), but it cannot be useful with the RMSProp optimizer.

\begin{table}[]
\caption{Comparison of different training methods for turning a DNN adaptive to changes in the input data dimensions. Accuracy denotes the classification accuracy and times (average time per each training epoch) are in the unit of seconds (s).}
\label{tab:training_methods}
\resizebox{\textwidth}{!}{%
\begin{tabular}{r|ll|ll|ll|ll} 
Dataset & \multicolumn{6}{c}{UCI-HAR~\cite{anguita2013public}} & \multicolumn{2}{c}{UTwente~\cite{shoaib2016complex}}\\
DNN Model & \multicolumn{2}{c}{CNN+FNN~\cite{ronao2016human}} & \multicolumn{4}{c|}{CNN+RNN~\cite{ordonez2016deep}} & \multicolumn{2}{c}{CNN+RNN~\cite{ordonez2016deep}} \\
Optimizer  & \multicolumn{2}{c|}{Adam} & \multicolumn{2}{c}{RMSProp} &\multicolumn{2}{c|}{Adam} & \multicolumn{2}{c}{Adam}  \\\toprule
 {\bf Method} &{\bf Accuracy (\%)} & {\bf Time} (s) &{\bf Accuracy (\%)} & {\bf Time} (s) & {\bf Accuracy (\%)} & {\bf Time} (s) & {\bf Accuracy (\%)} & {\bf Time} (s)\\
 Standard & 69.94$\pm$6.08 & 4.24$\pm$.29& 86.21$\pm$.11 &  6.03$\pm$.17  & 86.98$\pm$.21& 5.55$\pm$.15 & 87.02$\pm$.51 & 14.24$\pm$.44\\
 WeightAvg & 85.90$\pm$.25 & 5.63$\pm$.13 & 86.61$\pm$.19  & 6.54$\pm$.10 & 87.02$\pm$.21&6.07$\pm$.14 & 88.14$\pm$.99 & 15.03$\pm$.45\\ 
 Reptile  & 85.73$\pm$.25 & 5.50$\pm$.11 & 15.91$\pm$2.25 & 6.23$\pm$.10 & 86.87$\pm$.26 & 5.96$\pm$.19 & 87.87$\pm$.74 & 15.01$\pm$.43\\
 DAT (Ours) & 85.74$\pm$.20 & 3.68$\pm$.08 & 87.12$\pm$.13  & 5.58$\pm$.15 & 87.50$\pm$.11& 5.43$\pm$.15 & 88.91$\pm$.59 & 13.30$\pm$.57 \\\bottomrule     
\end{tabular}%
}
\end{table}

\section{Controlled Experiments}\label{sec:synth}
Here, we design a controlled experiment to better understand the capabilities of \dana, compared to other available solutions. We generate a synthetic five-class dataset that simulates two motion sensors in a way that we can adjust the correlation between patterns observed in two sensors. Thus, we provide evaluations in three different settings of correlation: low, moderate, and high correlation.

\subsection{Synthetic Dataset}\label{sec:synth_data}
When looking at real-world data of motion sensors, we observe that there are not only meaningful correlations between axes $x$,$y$, and $z$ of a sensor, but, more importantly, there are also considerable correlations between the same axes of different sensors. For example, in 
UTwente\cite{shoaib2016complex} dataset the average absolute of {\em Pearson correlation coefficient}\footnote{A measure of linear correlation between two sets of data with the absolute value between 0 and 1; see \url{https://en.wikipedia.org/wiki/Pearson\_correlation\_coefficient}.} between accelerometer and gyroscope, accelerometer and  magnetometer, and gyroscope and magnetometer are 0.25$\pm$.20, 0.34$\pm$.29, and 27$\pm$.21 respectively. Similarly, for UCI-HAR\cite{anguita2013public} we observe 0.21$\pm$.17 as the average absolute of Pearson correlation coefficient between accelerometer and gyroscope. It is important to note that there definitely are other types of non-linear correlations between the streams of two sensors. While such non-linear dependencies are not easy to measure, non-linear machine learning models, such as DNNs, can capture and utilize them.

Motivated by these facts, we generate a five-class dataset of six-variate time series (simulating two motion sensors), while: (1) we  control the correlation between two time series, and (2) we assign a class (\ie label) to each time series such that samples within the same class share some similarities while having some differences with samples of other classes. In the following, we briefly explain the ideas we applied for this purpose. The code and complete instructions, including all the details and parameters used for generating these synthetic datasets, are published at~\url{https://github.com/mmalekzadeh/dana}.
 
First, for axis $x$ of two sensors $S1$ and $S2$ (the same process is followed for other axes $y$ and $z$.), and for class $c\in\{0,1,2,3,4,\}$, we sample $S1_x$ and $S2_x$ from correlated Gaussian distributions with mean 0 and variance $\sigma^2_c \in \{0.6, 0.7, 0.8, 0.9, 1\}$ in three settings of correlations $\rho=\{low:0.01 ,\  moderate:0.58,\  high:0.89\}$, such that:
$S1_x \sim \mathcal{N}(0, \sigma^2_c) \quad \text{and} \quad S2_x \sim \mathcal{N}(\rho x^{2}_{i}, (1-\rho^2)\sigma^2_c).$
Considering width $w=50$, we repeat sampling for $i\in\{1,\dots,50\}$, that with our choices of value for $\rho$, gives us two time series with exact mutual information $I({S}1_x; {S}2_x) = -\frac{w}{2}\log(1-\rho^2) =\{low:0.01,\  moderate:0.20,\  high:0.80\}$. This is a standard technique for generating controllable correlated vectors~\cite{belghazi2018mutual, poole2019variational}. However, this is not enough, because in the current $S1_x$ and ${S}2_x$ we only have point-wise correlations between two sensors and no temporal correlation between sample points in each sensor. Thus, as the next step, we add some temporal patterns observed in real-world time series~\cite{shumway2000time}. In particular, we add three {\em periodic},  one {\em trend}, one {\em smoothing} (via moving average), and one {\em white noise} components to each time-series $S1_x$ and ${S}2_x$ generated in the previous step. We set the characteristics of these additional components such that they have meaningful contributions to the base Gaussian signals but do not entirely dominate them. For the low correlations setting, we add these components with different characteristics to each sensor, but for the high correlations setting, we add these components with similar characteristics to each sensor. For the moderate correlations setting, we add a mixture of components with similar and different characteristics to each sensor. Also, to make our classification task non-trivial, the characteristics of each component are chosen differently across different classes.

\begin{figure}[t]
    \begin{minipage}[t]{.48\columnwidth}
    \centering
    \includegraphics[width=\columnwidth]{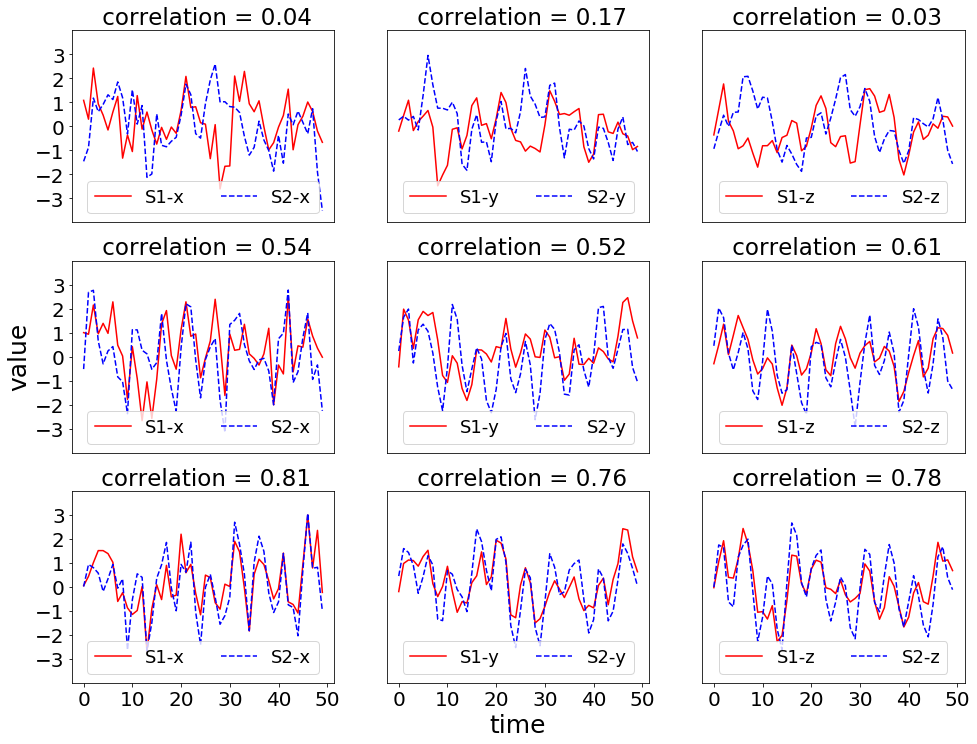}
    \end{minipage}
    \hfill
    \begin{minipage}[t]{.48\columnwidth}
    \centering
    \includegraphics[width=\columnwidth]{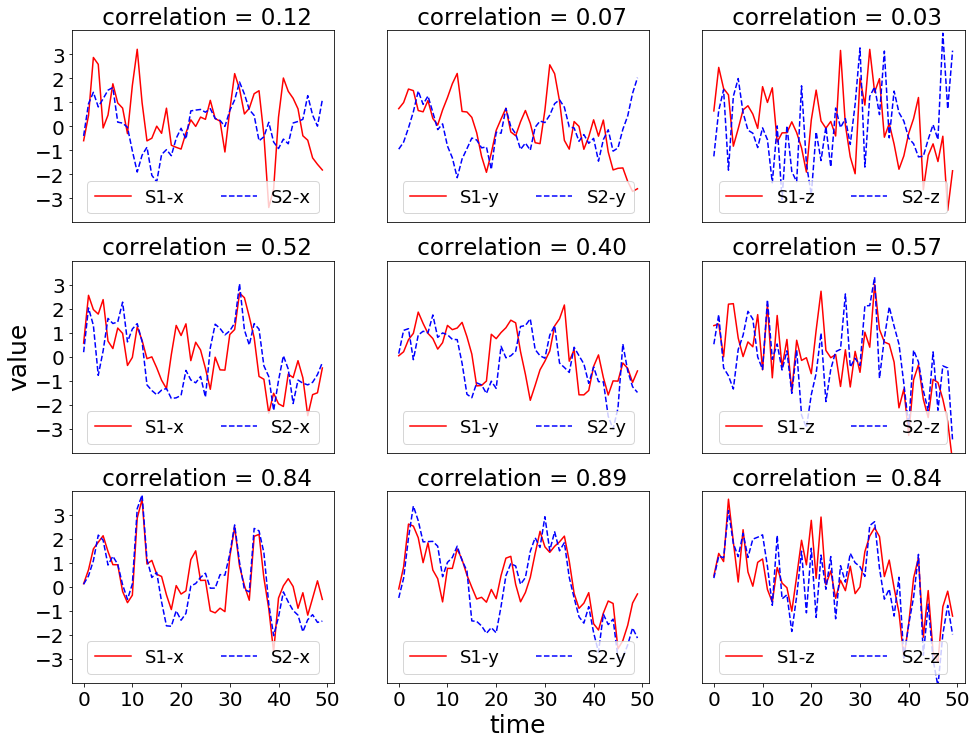}
    \end{minipage}
    \caption{Sample examples of the synthetic five-class dataset used in our controlled experiments. The nine plots on the left are samples of class 0, and those at the right are samples of class 4.  We consider two sensors $S1$ and $S2$ in three settings with average Pearson correlation (computed on the corresponding training dataset for each setting): $0.15\pm.10$ (as the low correlation setting, plots in the top row), $0.54\pm.11$ (as the moderate correlation setting, plots in the middle row), and $0.82\pm.05$ (as the high correlation setting, plots in the bottom row). The Pearson correlation coefficient for each sample is shown at the top of each plot.}\label{fig:synth_data_samples} 
\end{figure}

To build intuition, we plot some sample examples of the generated data in Fig.~\ref{fig:synth_data_samples}. For each setting (of low, moderate, and high correlation), we generate 4,000 time windows of dimensions $(w=50, h=6)$ as the training set; 900 time windows per class. Similarly, we generate 1,000 time windows as the test set; 100 time windows per class. We cannot compute the mutual information for the ultimate time series, because they include other non-Gaussian components, but we can compute Pearson correlation coefficient. For each setting, the average absolute of Pearson correlation coefficient is: (1) $0.15\pm.10$ for the low correlation setting, (2) $0.54\pm.11$ for the moderate correlation setting, and (3) $0.82\pm.05$ for the high correlation setting; which they are consistent with our initial correlations considered for different settings. In the training and test set of each class, the characteristics of all described components are the same, except the noise component (to keep training and test samples of each class similar, while having a slight difference between them).

\begin{figure}[t]
    \begin{minipage}[t]{\columnwidth}
    \centering
    \includegraphics[width=.573\columnwidth]{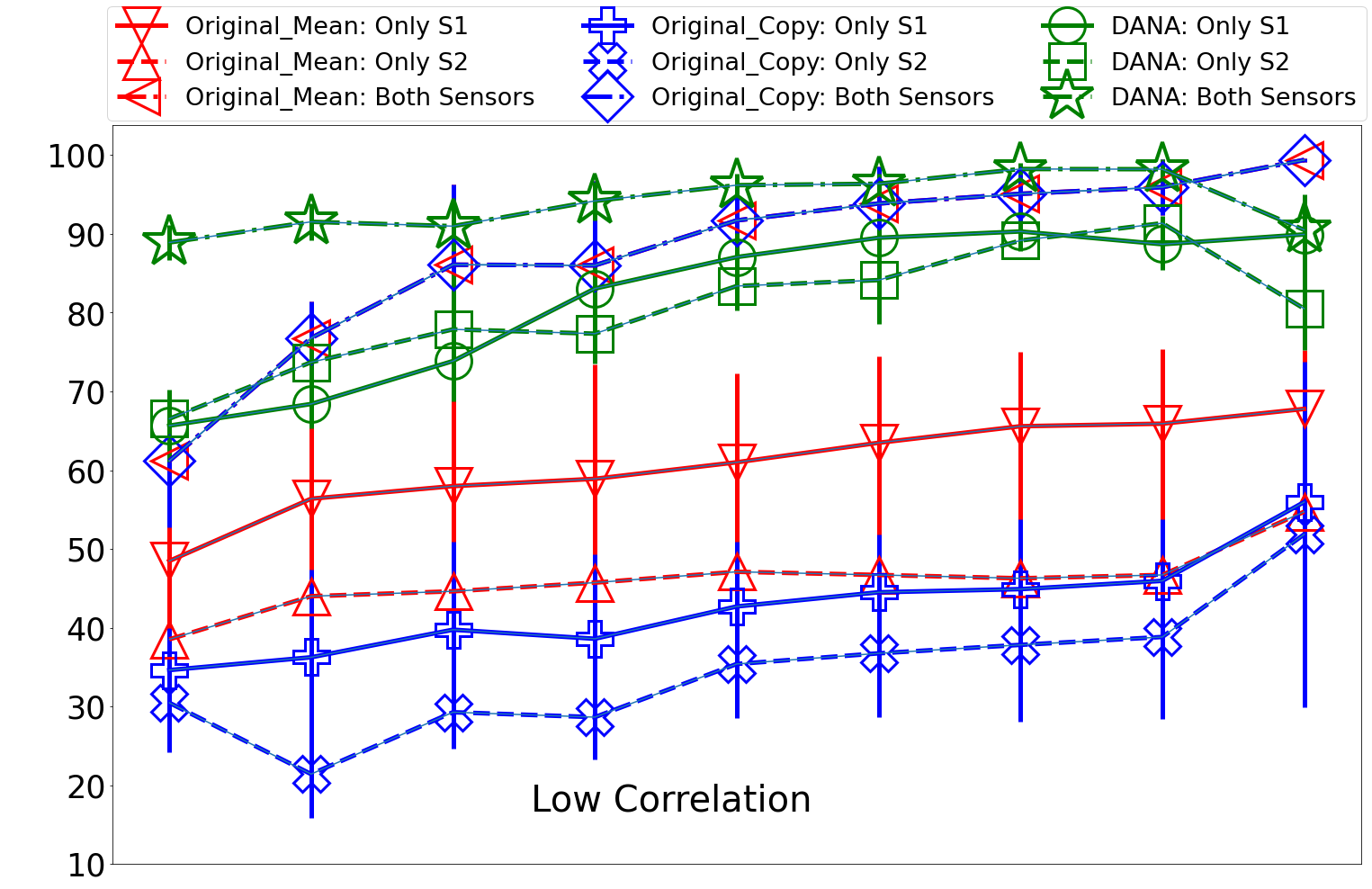}
    \end{minipage}
    \begin{minipage}[t]{\columnwidth}
    \centering
    \includegraphics[width=.573\columnwidth]{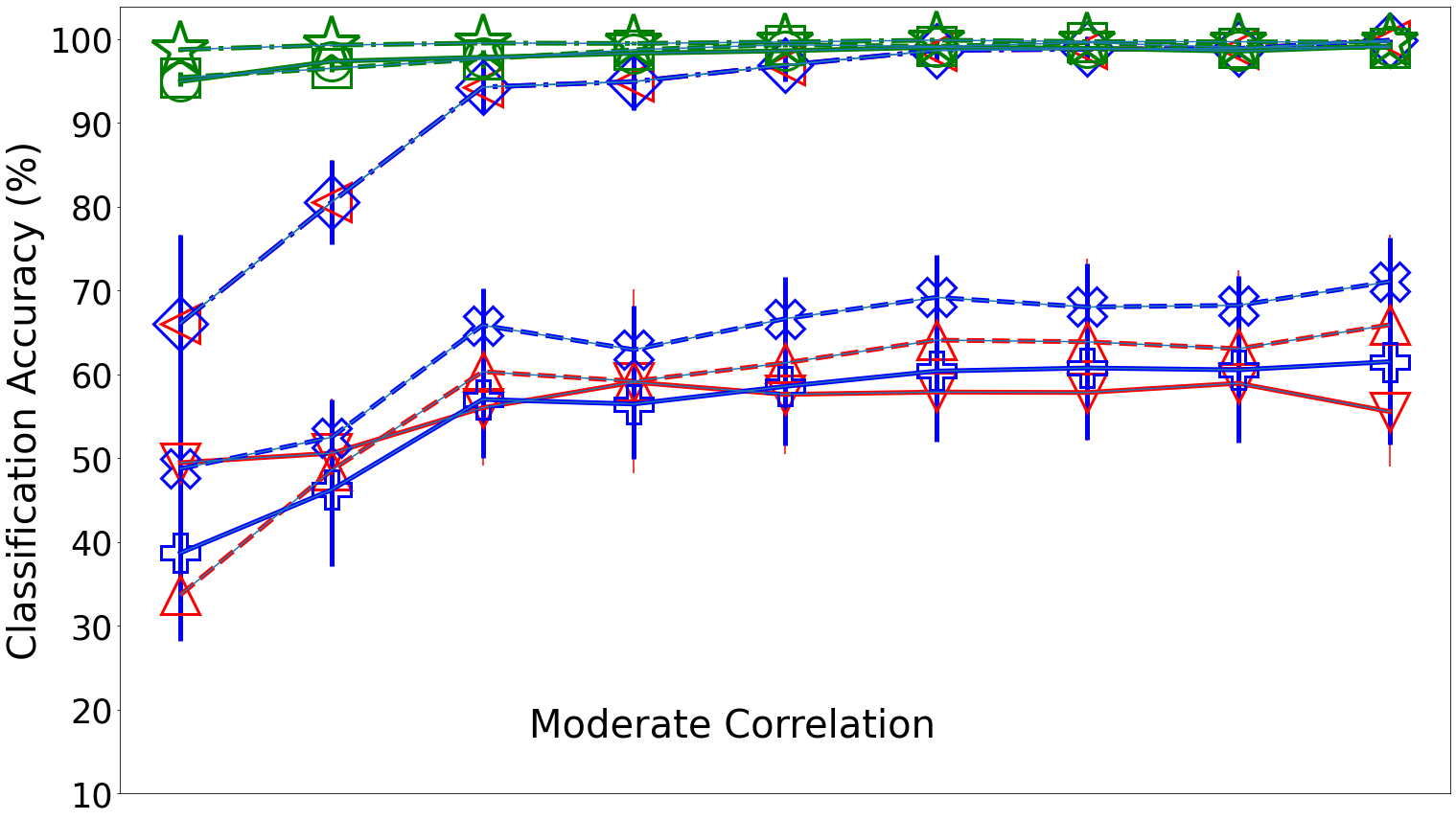}
    \end{minipage}
    \begin{minipage}[t]{\columnwidth}
    \centering
    \includegraphics[width=.573\columnwidth]{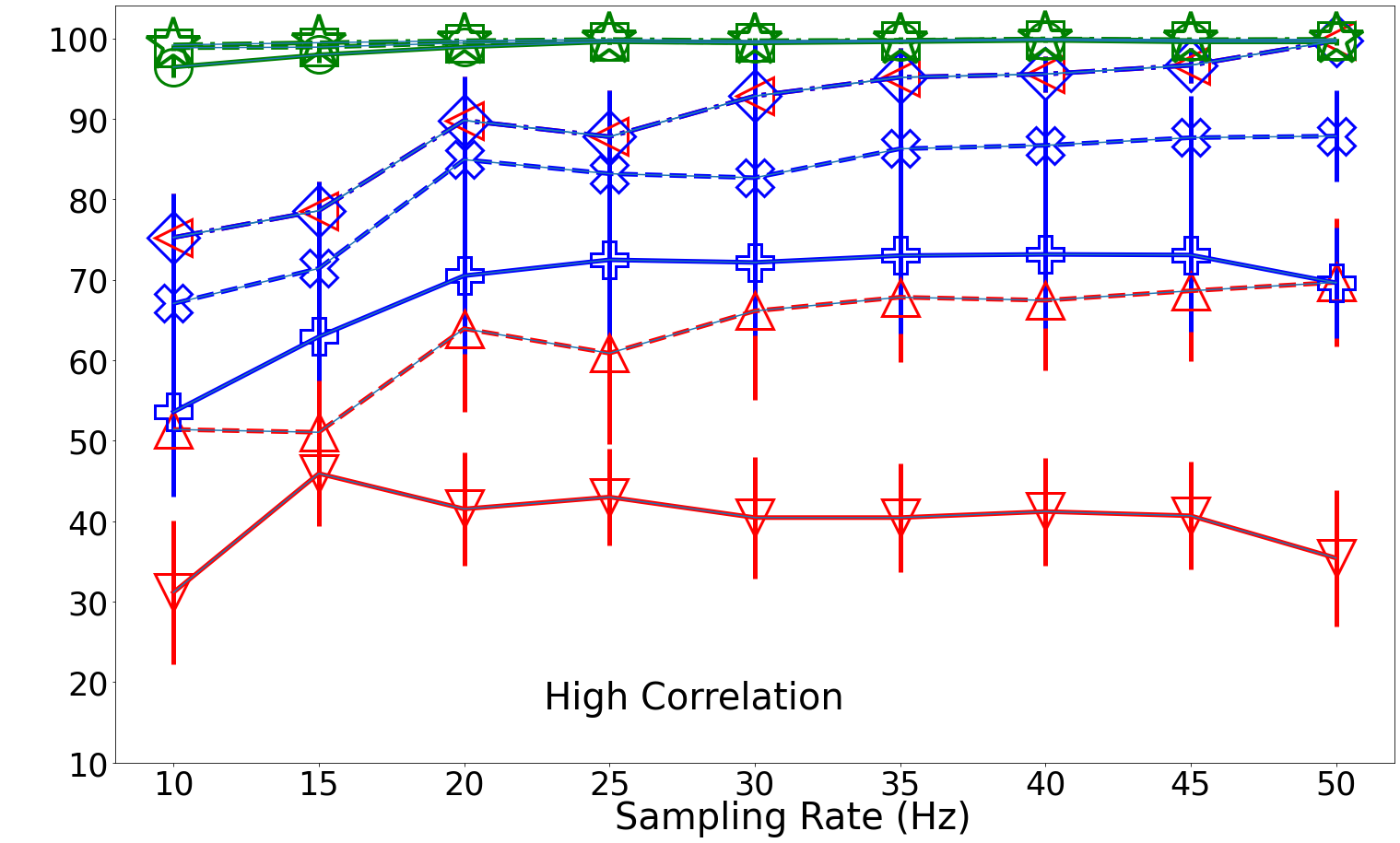}
    \end{minipage}
    \caption{Comparing \dana with original model on the synthetic dataset in three different settings: (top) {\em low} correlation, (middle) {\em moderate}, and (bottom) {\em high} correlation. Original\_Mean means imputing mean values for the sensor that is not available. Original\_Copy means duplicating the data of the available sensor for the sensor that is not available.}\label{fig:synth_data_results}
\end{figure} 

\subsection{Experimental Results}

We consider a very simple version of the architecture proposed in~\cite{ordonez2016deep} (see Figure~\ref{sup_mat_fig:dnn_1_ronao}) including 2 convolutional layers followed by 1 LSTM layer, where each layer has only 5 neurons. In total, this neural network has less than 2K trainable parameters. For each setting of {\em low}, {\em moderate}, and {\em high} correlation between two sensors, we compare: (1) the original model with standard training using two baselines, and (2) the \dana version of the model including DAP layer with DAT training. As our two baseline approaches, for the original model to deal with the missed (or deselected) sensor at inference time, we consider: (1) {\em original\_mean} where we use the mean values obtained from the training dataset for filling instead of the missed sensor, and (2) {\em original\_copy} where we duplicate the data of available sensor for the missed sensor. Moreover, to deal with changes in the sampling rate, in both baseline approaches, we up-sample the received data to 50Hz; as
the original (non-adaptive) model is trained for this sampling rate.  Fig.~\ref{fig:synth_data_results} shows the average classification accuracy for each model in different settings and three scenarios when the model receives data from both sensors $S1$ and $S2$, or only from one of them. In Fig.~\ref{fig:synth_data_perclass} we show the average performance (among all three scenarios of receiving data from both sensors $S1$ and $S2$, or only from one of them) presented separately for each class. Our findings are as follows:

\begin{figure}[t]
    \begin{minipage}[t]{.32\columnwidth}
    \centering
    \includegraphics[width=\columnwidth]{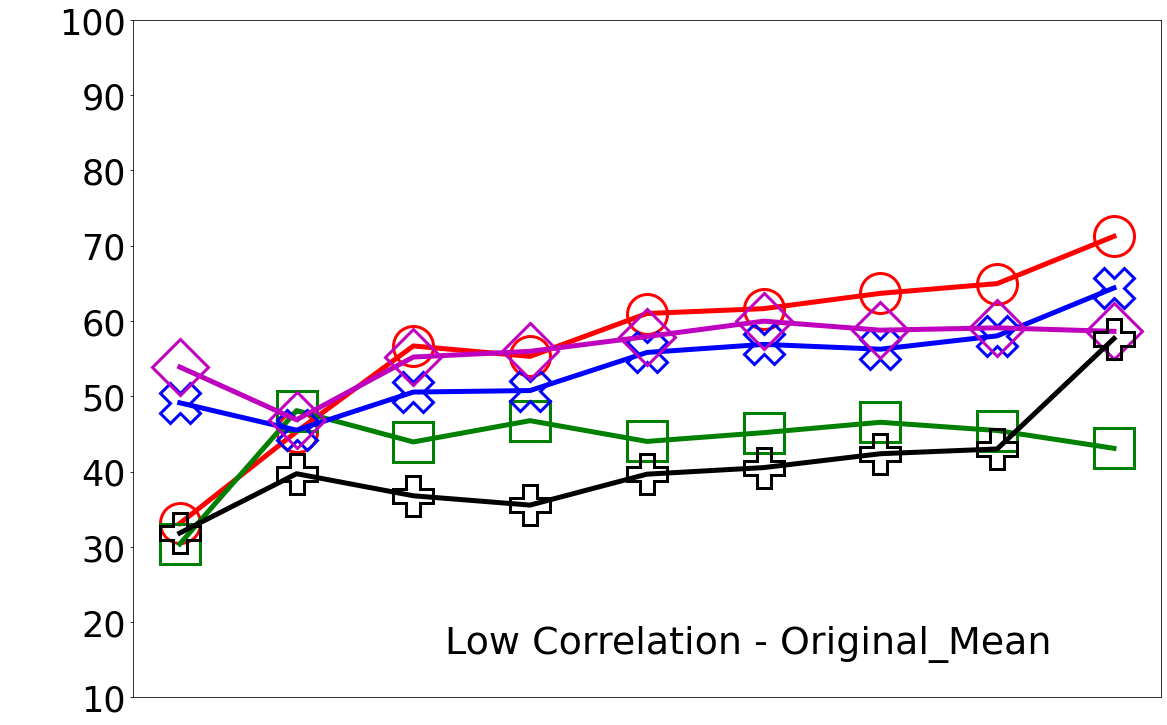}
    \end{minipage}
    \begin{minipage}[t]{.307\columnwidth}
    \includegraphics[width=\columnwidth]{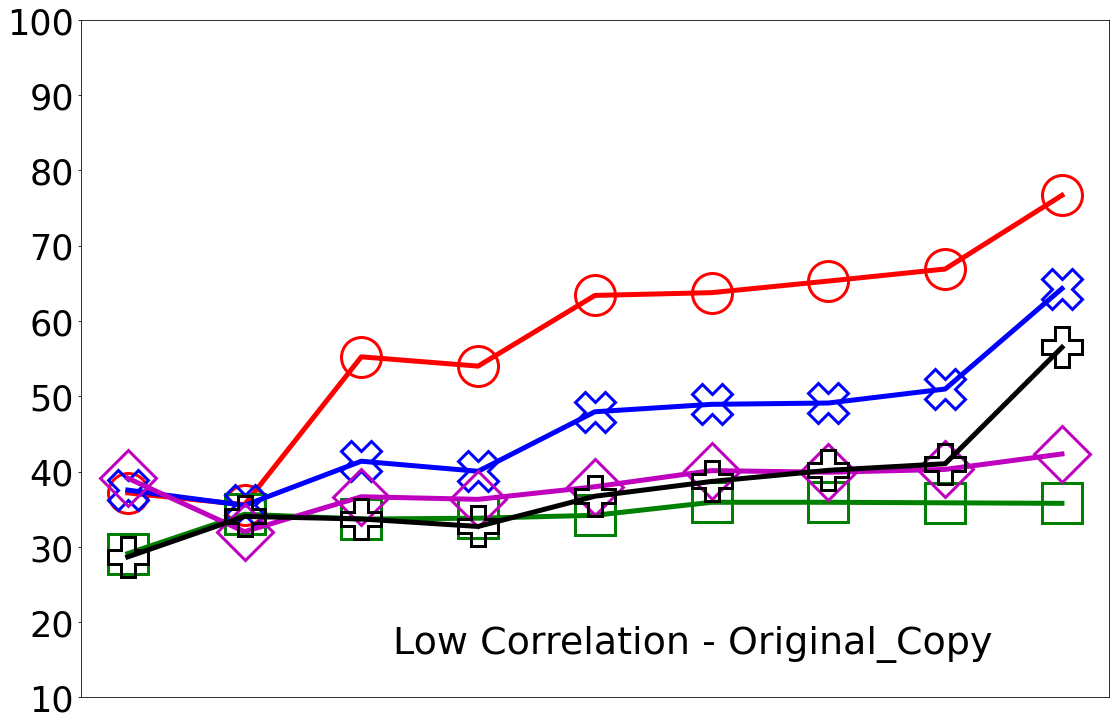}
    \end{minipage} 
    \begin{minipage}[t]{.303\columnwidth}
    \centering
    \includegraphics[width=\columnwidth]{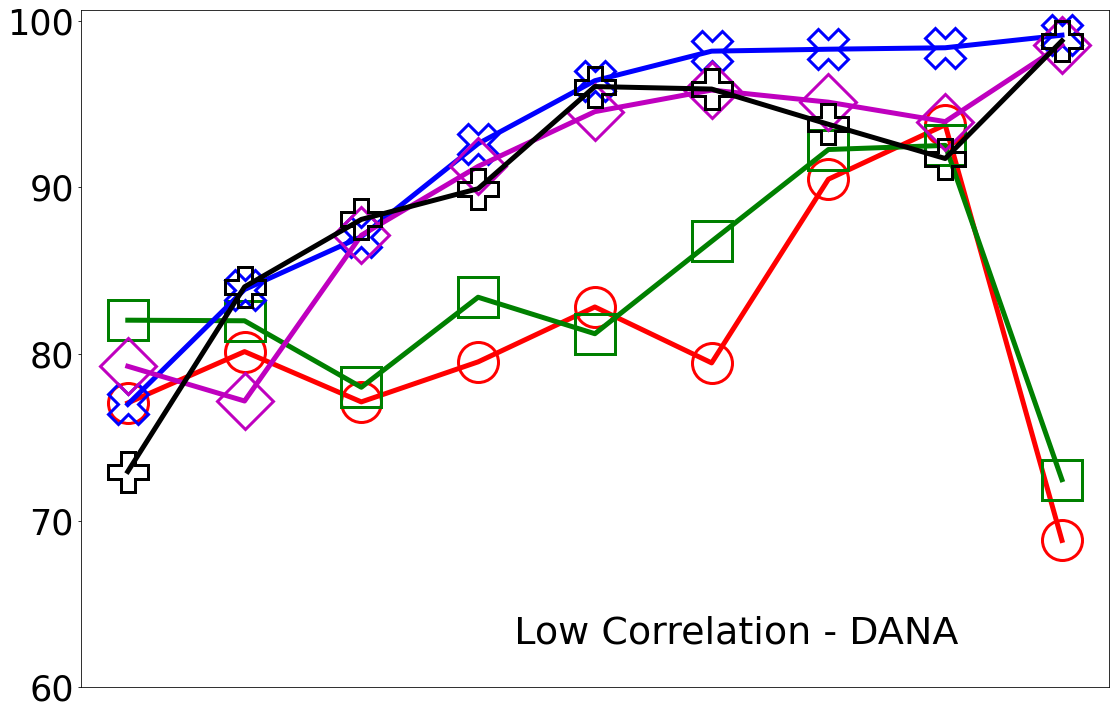} 
    \end{minipage} 
    \begin{minipage}[t]{.32\columnwidth}
    \centering
    \includegraphics[width=\columnwidth]{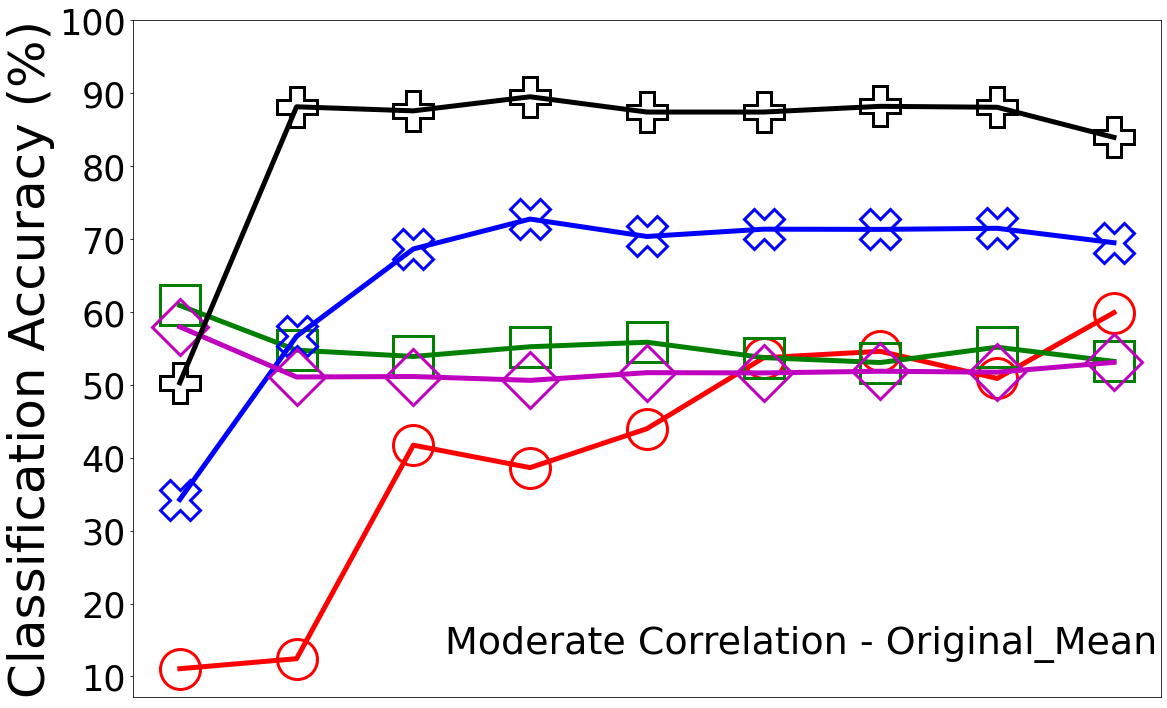}
    \end{minipage}
    \begin{minipage}[t]{.307\columnwidth}
    \includegraphics[width=\columnwidth]{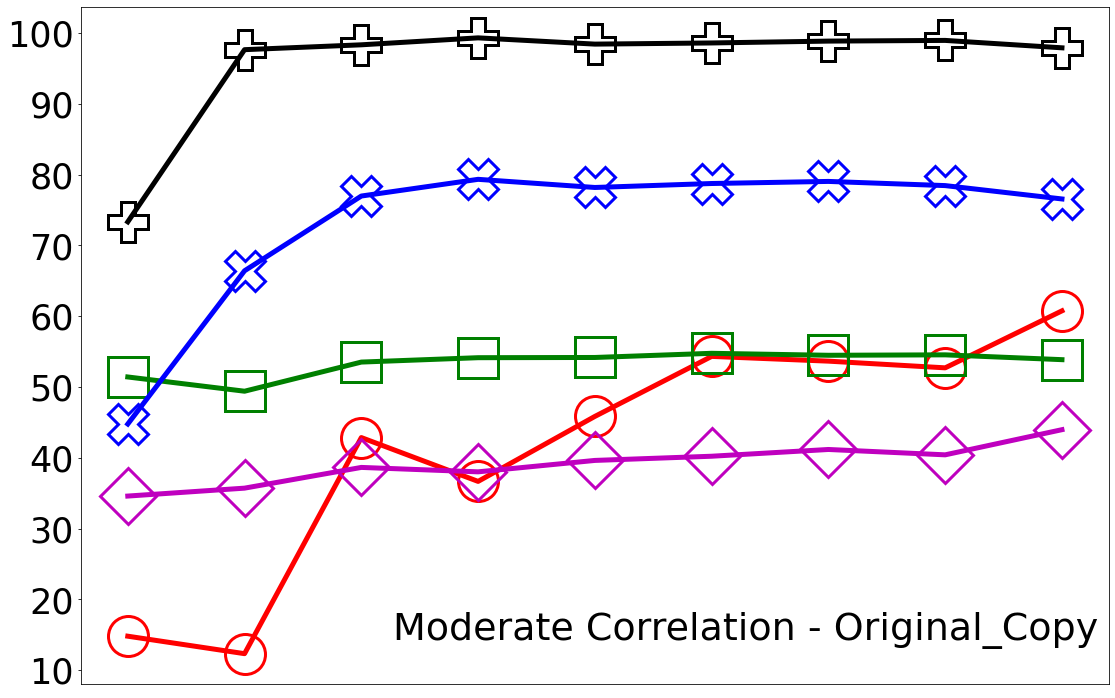}
    \end{minipage} 
    \begin{minipage}[t]{.303\columnwidth}
    \centering
    \includegraphics[width=\columnwidth]{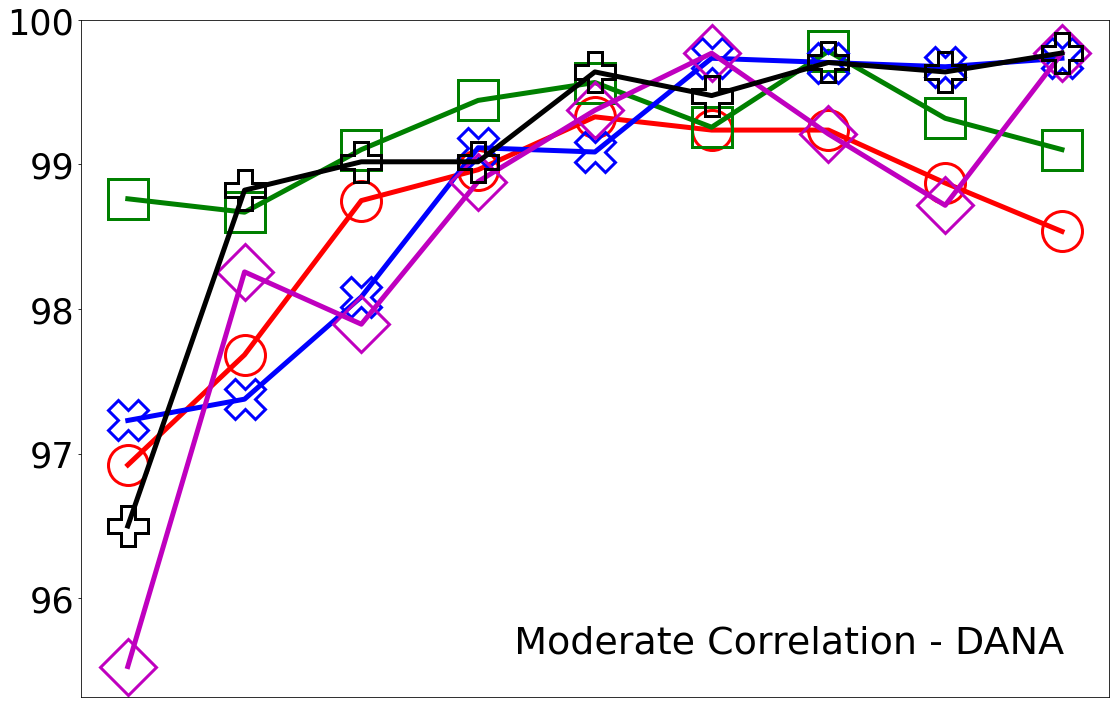} 
    \end{minipage} 
    \begin{minipage}[t]{.32\columnwidth}
    \centering
    \includegraphics[width=\columnwidth]{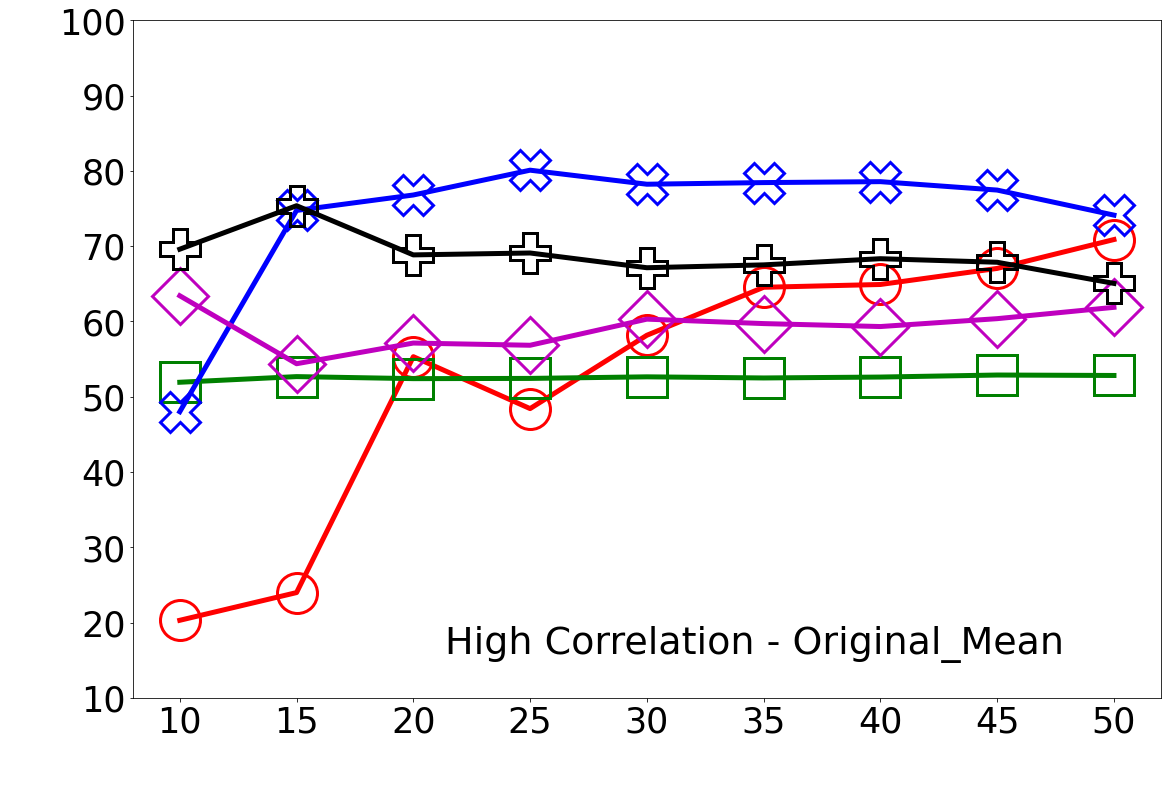}
    \end{minipage}
    \begin{minipage}[t]{.307\columnwidth}
    \includegraphics[width=\columnwidth]{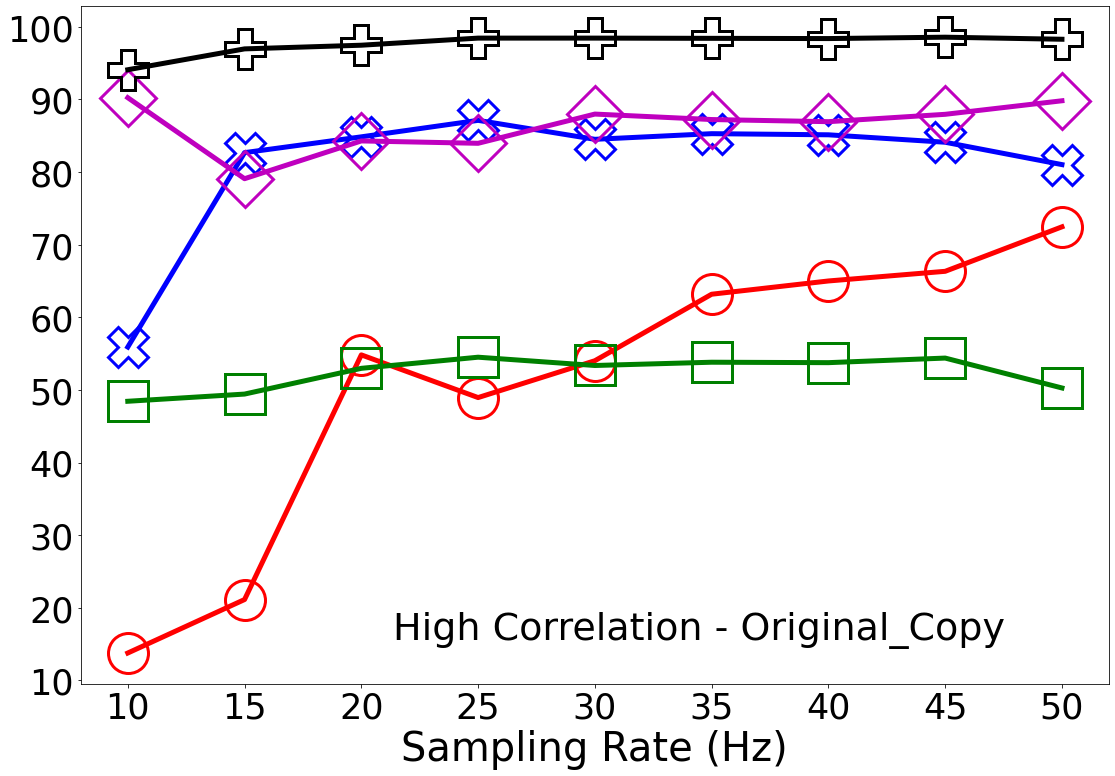}
    \end{minipage} 
    \begin{minipage}[t]{.303\columnwidth}
    \centering
    \includegraphics[width=\columnwidth]{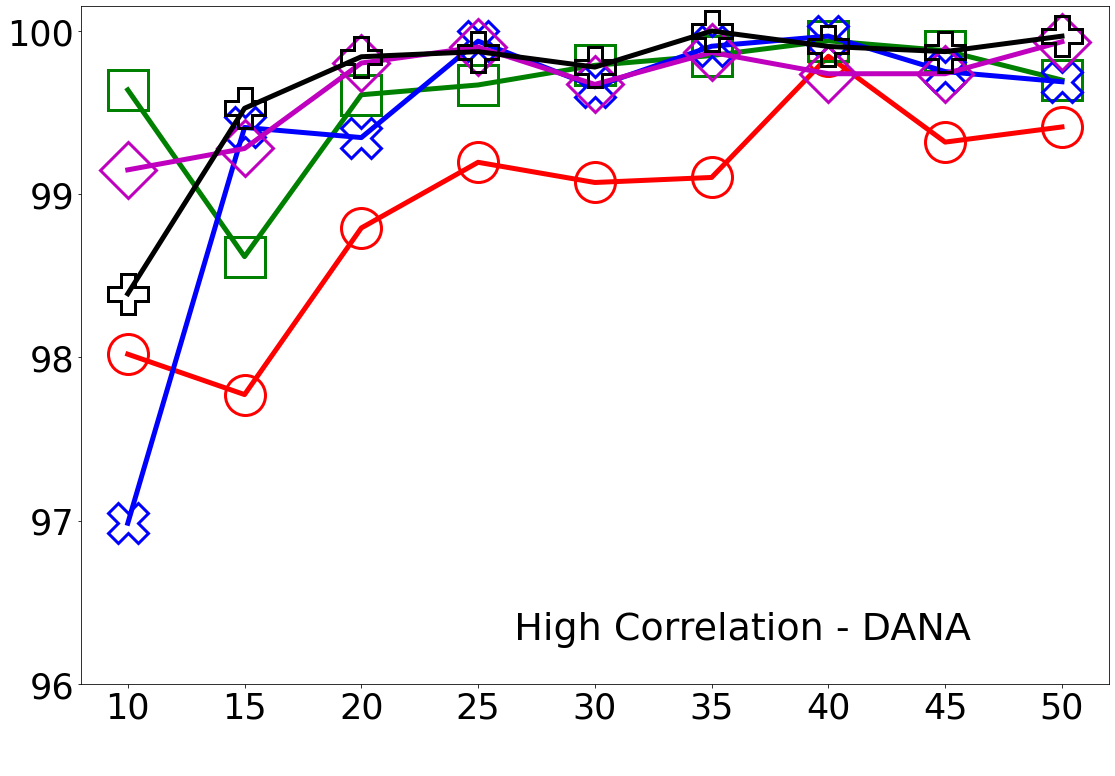} 
    \end{minipage} 
    \begin{minipage}[t]{\columnwidth}
    \centering
    \includegraphics[width=.6\columnwidth]{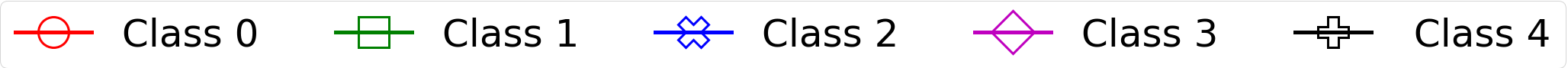}  
    \end{minipage} 
    \caption{The average classification accuracy (per each class in the dataset) of three scenarios of receiving data from both sensors $S1$ and $S2$, or only from one of them.\label{fig:synth_data_perclass}}  
\end{figure}

First, Fig.~\ref{fig:synth_data_results} shows that when both sensors are available and the sampling rate is 50Hz, the original model achieves a near 100\% accuracy in all three settings; which is expected as our synthetic data is not too complex. However, while in this situation, \dana achieves a performance similar to the original model in the moderate and high correlation setting, for the low correlation setting we observe a fall in the accuracy. Looking at Fig.~\ref{fig:synth_data_perclass}, where we plot the average results per each class, we see that this fall in \dana's accuracy is due to the poor performance for class 0 and class 1. This shows that in situations where there is a low correlation between sensors, \dana cannot easily offer performance similar to the original model. 

Second, the correlation between two sensors plays an important role, as we see that both the original and \dana models perform better when there are such correlations. However, the correlation is much more important to \dana than the original model. Interestingly, we see that even a moderate correlation can be very helpful to \dana, as we see that \dana can take advantage of correlation in the data very well in the moderated and high correlation settings. Correlation also helps baseline methods, as we see that in the low correlation setting, the copying data of one sensor is worse than  just using the mean values. But in the setting of high correlation, we can get much better results by copying the data of the available sensor than just using the mean value of training data. Interestingly, when the correlation is moderate, these two approaches are almost the same.

Third, we see that the variance in \dana is negligible in Fig.~\ref{fig:synth_data_results}, but baseline approaches show much higher variance, especially for situations of missing a sensor. Notice that on average, \dana works much better than the original model, even for the low correlation setting. In addition, Fig.~\ref{fig:synth_data_perclass} shows that when there is a moderate or high correlation, \dana shows better performance in terms of similarity between different classes. However, the original model shows very different results for each class. We believe this is due to the better generalization capability of \dana, as we also observed in our experiments presented on real-world datasets.  

 {Finally, in this experiment, the concept of correlations between patterns in sensor data is not necessarily restricted to the different sensors of the same device. In general, we can also consider the correlations between the same sensors of different devices. For instance, wearable devices for monitoring people's health conditions might include sensors at different locations of the user's body. Since all  sensors share a common source of pattern generation~(\ie the user's), their data should have meaningful correlations. However, as our results throughout this paper have shown, when DNNs have not encountered sufficient variability in their input dimensions, they do not capture such correlations and consequently show poor performances when they encounter such variability at inference time. Moreover, it is not intuitive and easy to know what are the underlying rules and correlation patterns between different sensors or different locations. Therefore, another motivation of using \dana, and in general adaptive DNNs, is that they can (based on our experimental observations) capture complex rules in the hierarchy of patterns extracted by the stacked layers in the DNN.}

 {As an example, based on the results in Fig.~\ref{fig:synth_data_results}, we observed that if we know how much two sensors are correlated, then we can better choose which imputation technique to apply. Like, if $correlation < C_{low}$ then impute the mean values, and if $correlation > C_{high}$ then impute a copy of available sensors, where $C_{low}$ and $C_{high}$ are two threshold values that can be set based on empirical observations. Readers can immediately notice that how finding and applying such rules make the task more complicated. On the other hand, we observed that an adaptive model can implicitly find and apply such a rule; showing that by shaping and training a model to be adaptive, we can take better advantage of the capacity of DNNs and capture much more complex rules.}

\section{Discussion for Future Studies} \label{sec:discc}

In this section, we discuss challenges and new research directions related to \dana and the problem of adaptive sensor and sampling rate selection.

The focus and evaluation of \dana were on motion sensor data. It will be interesting to apply \dana to other types of temporal data, such as audio streams.  This needs optimizing the code of the proposed DAP layer for other data types. Moreover, to cover the  range of possible data dimensions at inference time, we randomly covered a subset of the possible situations at each round of training \dana. One can study the behavior of this randomized selection alongside the DAT procedure and offer an analysis of the convergence of such training approaches.

An important research direction is also related to measuring the impact of \dana in  power saving when running on a wearable or mobile device  and the interplay between the operating system of the user's device is an important direction. \dana has applications for situations where users want to control applications' permissions over sensors, which also require research in other areas of such human-computer interactions.

Finally, the majority of public datasets of motion-sensor data do not simultaneously satisfy the requirements of abundance and variety of activities, users, devices, and the number of sensors. Collecting, or getting access to, a larger dataset can help to extend the evaluation of \dana to larger experimental scenarios.

\section{Conclusion}

We presented \dana, a solution to make deep neural networks adaptive to changes in the dimensions of the input data to cope with sensor availability and adaptive sampling at inference time. \dana provides a single trained model that retains high classification accuracy across a range of feasible scenarios, thus avoiding the need for a separate classifier for each setting. Moreover, \dana imposes no limitations to existing DNNs for multivariate sensor data, and it makes DNNs flexible to changes without adding or removing trainable parameters.

We showed that \dana outperforms the state of the art over a range of sampling rates and retains accuracy when some sensors are unavailable  at inference time.  For instance, on a dataset of 3 sensors and 13 activities, \dana keeps classification accuracy similar to the original DNN in a range of 6Hz to 50Hz and its accuracy only falls from 95\% to around 90\% and 85\% in case of missing one or two of the three sensors, respectively, while the original DNN cannot handle these changes, or achieve at most 75\% and 55\% accuracy with resampling and imputation prepossessing. Our evaluations on three synthetic datasets have shown that \dana better captures correlated patterns in multi-variate sensor data.

\begin{acks}
The work was supported by the Life Sciences Initiative at Queen Mary University of London. Mohammad Malekzadeh was partially supported by the UK EPSRC (grant no. EP/T023600/1) within the CHIST-ERA program. Andrea Cavallaro wishes to thank The Alan Turing Institute (EP/N510129/1), which is funded by the EPSRC, for its support through the project PRIMULA.  Hamed Haddadi was partially supported by EPSRC Databox: Privacy-Aware Infrastructure for Managing Personal Data (EP/N028260/1) and EPSRC DADA: Defence Against Dark Artefacts (EP/R03351X/1). Authors thank anonymous reviewers for their constructive suggestions provided on the initial version of this paper.

\end{acks}
\clearpage
\bibliographystyle{ACM-Reference-Format}
\bibliography{references}


\begin{thebibliography}{73}


\ifx \showCODEN    \undefined \def \showCODEN     #1{\unskip}     \fi
\ifx \showDOI      \undefined \def \showDOI       #1{#1}\fi
\ifx \showISBNx    \undefined \def \showISBNx     #1{\unskip}     \fi
\ifx \showISBNxiii \undefined \def \showISBNxiii  #1{\unskip}     \fi
\ifx \showISSN     \undefined \def \showISSN      #1{\unskip}     \fi
\ifx \showLCCN     \undefined \def \showLCCN      #1{\unskip}     \fi
\ifx \shownote     \undefined \def \shownote      #1{#1}          \fi
\ifx \showarticletitle \undefined \def \showarticletitle #1{#1}   \fi
\ifx \showURL      \undefined \def \showURL       {\relax}        \fi
\providecommand\bibfield[2]{#2}
\providecommand\bibinfo[2]{#2}
\providecommand\natexlab[1]{#1}
\providecommand\showeprint[2][]{arXiv:#2}

\bibitem[\protect\citeauthoryear{Abadi, Barham, Chen, Chen, Davis, Dean, Devin,
  Ghemawat, Irving, Isard, et~al\mbox{.}}{Abadi et~al\mbox{.}}{2016}]%
        {abadi2016tensorflow}
\bibfield{author}{\bibinfo{person}{Martin Abadi}, \bibinfo{person}{Paul
  Barham}, \bibinfo{person}{Jianmin Chen}, \bibinfo{person}{Zhifeng Chen},
  \bibinfo{person}{Andy Davis}, \bibinfo{person}{Jeffrey Dean},
  \bibinfo{person}{Matthieu Devin}, \bibinfo{person}{Sanjay Ghemawat},
  \bibinfo{person}{Geoffrey Irving}, \bibinfo{person}{Michael Isard},
  {et~al\mbox{.}}} \bibinfo{year}{2016}\natexlab{}.
\newblock \showarticletitle{Tensorflow: A system for large-scale machine
  learning}. In \bibinfo{booktitle}{\emph{12th USENIX Symposium on Operating
  Systems Design and Implementation (OSDI 16)}}. \bibinfo{pages}{265--283}.
\newblock


\bibitem[\protect\citeauthoryear{Abdu-Aguye and Gomaa}{Abdu-Aguye and
  Gomaa}{2019}]%
        {abdu2019versatl}
\bibfield{author}{\bibinfo{person}{Mubarak~G Abdu-Aguye} {and}
  \bibinfo{person}{Walid Gomaa}.} \bibinfo{year}{2019}\natexlab{}.
\newblock \showarticletitle{VersaTL: versatile transfer learning for IMU-based
  activity recognition using convolutional neural networks}. In
  \bibinfo{booktitle}{\emph{The 16th International Conference on Informatics in
  Control, Automation and Robotics (ICINCO)}}.
\newblock


\bibitem[\protect\citeauthoryear{Ae~Lee and Gill}{Ae~Lee and Gill}{2018}]%
        {ae2018missing}
\bibfield{author}{\bibinfo{person}{Jung Ae~Lee} {and} \bibinfo{person}{Jeff
  Gill}.} \bibinfo{year}{2018}\natexlab{}.
\newblock \showarticletitle{Missing value imputation for physical activity data
  measured by accelerometer}.
\newblock \bibinfo{journal}{\emph{Statistical methods in medical research}}
  \bibinfo{volume}{27}, \bibinfo{number}{2} (\bibinfo{year}{2018}),
  \bibinfo{pages}{490--506}.
\newblock


\bibitem[\protect\citeauthoryear{Anguita, Ghio, Oneto, Parra, and
  Reyes-Ortiz}{Anguita et~al\mbox{.}}{2013}]%
        {anguita2013public}
\bibfield{author}{\bibinfo{person}{Davide Anguita}, \bibinfo{person}{Alessandro
  Ghio}, \bibinfo{person}{Luca Oneto}, \bibinfo{person}{Xavier Parra}, {and}
  \bibinfo{person}{Jorge~Luis Reyes-Ortiz}.} \bibinfo{year}{2013}\natexlab{}.
\newblock \showarticletitle{A public domain dataset for human activity
  recognition using smartphones.}. In \bibinfo{booktitle}{\emph{Proceedings of
  the 21st European Symposium on Artificial Neural Networks, Computational
  Intelligence and Machine Learning, ESANN}}. \bibinfo{pages}{437--442}.
\newblock


\bibitem[\protect\citeauthoryear{Belghazi, Baratin, Rajeshwar, Ozair, Bengio,
  Courville, and Hjelm}{Belghazi et~al\mbox{.}}{2018}]%
        {belghazi2018mutual}
\bibfield{author}{\bibinfo{person}{Mohamed~Ishmael Belghazi},
  \bibinfo{person}{Aristide Baratin}, \bibinfo{person}{Sai Rajeshwar},
  \bibinfo{person}{Sherjil Ozair}, \bibinfo{person}{Yoshua Bengio},
  \bibinfo{person}{Aaron Courville}, {and} \bibinfo{person}{Devon Hjelm}.}
  \bibinfo{year}{2018}\natexlab{}.
\newblock \showarticletitle{Mutual information neural estimation}. In
  \bibinfo{booktitle}{\emph{International Conference on Machine Learning}}.
  PMLR, \bibinfo{pages}{531--540}.
\newblock


\bibitem[\protect\citeauthoryear{Bengio, Courville, and Vincent}{Bengio
  et~al\mbox{.}}{2013}]%
        {bengio2013representation}
\bibfield{author}{\bibinfo{person}{Yoshua Bengio}, \bibinfo{person}{Aaron
  Courville}, {and} \bibinfo{person}{Pascal Vincent}.}
  \bibinfo{year}{2013}\natexlab{}.
\newblock \showarticletitle{Representation learning: A review and new
  perspectives}.
\newblock \bibinfo{journal}{\emph{IEEE transactions on pattern analysis and
  machine intelligence}} \bibinfo{volume}{35}, \bibinfo{number}{8}
  (\bibinfo{year}{2013}), \bibinfo{pages}{1798--1828}.
\newblock


\bibitem[\protect\citeauthoryear{Bonawitz, Eichner, Grieskamp, Huba, Ingerman,
  Ivanov, Kiddon, Konecny, Mazzocchi, McMahan, Van~Overveldt, Petrou, Ramage,
  and Roselander}{Bonawitz et~al\mbox{.}}{2019}]%
        {bonawitz2019towards}
\bibfield{author}{\bibinfo{person}{Keith Bonawitz}, \bibinfo{person}{Hubert
  Eichner}, \bibinfo{person}{Wolfgang Grieskamp}, \bibinfo{person}{Dzmitry
  Huba}, \bibinfo{person}{Alex Ingerman}, \bibinfo{person}{Vladimir Ivanov},
  \bibinfo{person}{Chloe Kiddon}, \bibinfo{person}{Jakub Konecny},
  \bibinfo{person}{Stefano Mazzocchi}, \bibinfo{person}{H~Brendan McMahan},
  \bibinfo{person}{Timon Van~Overveldt}, \bibinfo{person}{David Petrou},
  \bibinfo{person}{Daniel Ramage}, {and} \bibinfo{person}{Jason. Roselander}.}
  \bibinfo{year}{2019}\natexlab{}.
\newblock \showarticletitle{Towards federated learning at scale: System
  design}. In \bibinfo{booktitle}{\emph{Proceedings of the 2nd SysML
  Conference, Palo Alto, CA, USA}}.
\newblock


\bibitem[\protect\citeauthoryear{Bulling, Blanke, and Schiele}{Bulling
  et~al\mbox{.}}{2014}]%
        {bulling2014tutorial}
\bibfield{author}{\bibinfo{person}{Andreas Bulling}, \bibinfo{person}{Ulf
  Blanke}, {and} \bibinfo{person}{Bernt Schiele}.}
  \bibinfo{year}{2014}\natexlab{}.
\newblock \showarticletitle{A tutorial on human activity recognition using
  body-worn inertial sensors}.
\newblock \bibinfo{journal}{\emph{ACM Computing Surveys (CSUR)}}
  \bibinfo{volume}{46}, \bibinfo{number}{3} (\bibinfo{year}{2014}),
  \bibinfo{pages}{1--33}.
\newblock


\bibitem[\protect\citeauthoryear{Chen, Lin, Chen, Lane, Cardone, Wang, Li,
  Chen, Choudhury, and Campbell}{Chen et~al\mbox{.}}{2013}]%
        {chen2013unobtrusive}
\bibfield{author}{\bibinfo{person}{Zhenyu Chen}, \bibinfo{person}{Mu Lin},
  \bibinfo{person}{Fanglin Chen}, \bibinfo{person}{Nicholas~D Lane},
  \bibinfo{person}{Giuseppe Cardone}, \bibinfo{person}{Rui Wang},
  \bibinfo{person}{Tianxing Li}, \bibinfo{person}{Yiqiang Chen},
  \bibinfo{person}{Tanzeem Choudhury}, {and} \bibinfo{person}{Andrew~T
  Campbell}.} \bibinfo{year}{2013}\natexlab{}.
\newblock \showarticletitle{Unobtrusive sleep monitoring using smartphones}. In
  \bibinfo{booktitle}{\emph{7th International Conference on Pervasive Computing
  Technologies for Healthcare and Workshops}} (Venice, Italy). IEEE,
  \bibinfo{pages}{145--152}.
\newblock


\bibitem[\protect\citeauthoryear{Cheng, Erfani, Zhang, and Kotagiri}{Cheng
  et~al\mbox{.}}{2018}]%
        {cheng2018learning}
\bibfield{author}{\bibinfo{person}{Weihao Cheng}, \bibinfo{person}{Sarah
  Erfani}, \bibinfo{person}{Rui Zhang}, {and} \bibinfo{person}{Ramamohanarao
  Kotagiri}.} \bibinfo{year}{2018}\natexlab{}.
\newblock \showarticletitle{Learning datum-wise sampling frequency for
  energy-efficient human activity recognition}. In
  \bibinfo{booktitle}{\emph{Thirty-Second AAAI Conference on Artificial
  Intelligence}}.
\newblock


\bibitem[\protect\citeauthoryear{Choi and Lee}{Choi and Lee}{2019}]%
        {choi2019embracenet}
\bibfield{author}{\bibinfo{person}{Jun-Ho Choi} {and}
  \bibinfo{person}{Jong-Seok Lee}.} \bibinfo{year}{2019}\natexlab{}.
\newblock \showarticletitle{EmbraceNet: A robust deep learning architecture for
  multimodal classification}.
\newblock \bibinfo{journal}{\emph{Information Fusion}}  \bibinfo{volume}{51}
  (\bibinfo{year}{2019}), \bibinfo{pages}{259--270}.
\newblock


\bibitem[\protect\citeauthoryear{Choubey, Pateria, Saxena, SB, Jha, and
  PM}{Choubey et~al\mbox{.}}{2015}]%
        {choubey2015power}
\bibfield{author}{\bibinfo{person}{Prafulla~Kumar Choubey},
  \bibinfo{person}{Shubham Pateria}, \bibinfo{person}{Aseem Saxena},
  \bibinfo{person}{Vaisakh Punnekkattu~Chirayil SB},
  \bibinfo{person}{Krishna~Kishor Jha}, {and} \bibinfo{person}{Sharana~Basaiah
  PM}.} \bibinfo{year}{2015}\natexlab{}.
\newblock \showarticletitle{Power efficient, bandwidth optimized and fault
  tolerant sensor management for IOT in Smart Home}. In
  \bibinfo{booktitle}{\emph{2015 IEEE International Advance Computing
  Conference (IACC)}}. IEEE, \bibinfo{pages}{366--370}.
\newblock


\bibitem[\protect\citeauthoryear{Chu, Lane, Lai, Pang, Meng, Guo, Li, and
  Zhao}{Chu et~al\mbox{.}}{2011}]%
        {chu2011balancing}
\bibfield{author}{\bibinfo{person}{David Chu}, \bibinfo{person}{Nicholas~D
  Lane}, \bibinfo{person}{Ted Tsung-Te Lai}, \bibinfo{person}{Cong Pang},
  \bibinfo{person}{Xiangying Meng}, \bibinfo{person}{Qing Guo},
  \bibinfo{person}{Fan Li}, {and} \bibinfo{person}{Feng Zhao}.}
  \bibinfo{year}{2011}\natexlab{}.
\newblock \showarticletitle{Balancing energy, latency and accuracy for mobile
  sensor data classification}. In \bibinfo{booktitle}{\emph{Proceedings of the
  9th ACM Conference on Embedded Networked Sensor Systems}}.
  \bibinfo{pages}{54--67}.
\newblock


\bibitem[\protect\citeauthoryear{Corder and Foreman}{Corder and
  Foreman}{2014}]%
        {corder2014nonparametric}
\bibfield{author}{\bibinfo{person}{Gregory~W Corder} {and}
  \bibinfo{person}{Dale~I Foreman}.} \bibinfo{year}{2014}\natexlab{}.
\newblock \bibinfo{booktitle}{\emph{Nonparametric statistics: A step-by-step
  approach}}.
\newblock \bibinfo{publisher}{John Wiley \& Sons}.
\newblock


\bibitem[\protect\citeauthoryear{Finn, Abbeel, and Levine}{Finn
  et~al\mbox{.}}{2017}]%
        {finn2017model}
\bibfield{author}{\bibinfo{person}{Chelsea Finn}, \bibinfo{person}{Pieter
  Abbeel}, {and} \bibinfo{person}{Sergey Levine}.}
  \bibinfo{year}{2017}\natexlab{}.
\newblock \showarticletitle{Model-agnostic meta-learning for fast adaptation of
  deep networks}. In \bibinfo{booktitle}{\emph{Proceedings of the 34th
  International Conference on Machine Learning-Volume 70}}. JMLR. org,
  \bibinfo{pages}{1126--1135}.
\newblock


\bibitem[\protect\citeauthoryear{French}{French}{1999}]%
        {french1999catastrophic}
\bibfield{author}{\bibinfo{person}{Robert~M French}.}
  \bibinfo{year}{1999}\natexlab{}.
\newblock \showarticletitle{Catastrophic forgetting in connectionist networks}.
\newblock \bibinfo{journal}{\emph{Trends in cognitive sciences}}
  \bibinfo{volume}{3}, \bibinfo{number}{4} (\bibinfo{year}{1999}),
  \bibinfo{pages}{128--135}.
\newblock


\bibitem[\protect\citeauthoryear{Ghasemzadeh, Amini, Saeedi, and
  Sarrafzadeh}{Ghasemzadeh et~al\mbox{.}}{2014}]%
        {ghasemzadeh2014power}
\bibfield{author}{\bibinfo{person}{Hassan Ghasemzadeh}, \bibinfo{person}{Navid
  Amini}, \bibinfo{person}{Ramyar Saeedi}, {and} \bibinfo{person}{Majid
  Sarrafzadeh}.} \bibinfo{year}{2014}\natexlab{}.
\newblock \showarticletitle{Power-aware computing in wearable sensor networks:
  An optimal feature selection}.
\newblock \bibinfo{journal}{\emph{IEEE Transactions on Mobile Computing}}
  \bibinfo{volume}{14}, \bibinfo{number}{4} (\bibinfo{year}{2014}),
  \bibinfo{pages}{800--812}.
\newblock


\bibitem[\protect\citeauthoryear{Goodfellow, Bengio, Courville, and
  Bengio}{Goodfellow et~al\mbox{.}}{2016}]%
        {goodfellow2016deep}
\bibfield{author}{\bibinfo{person}{Ian Goodfellow}, \bibinfo{person}{Yoshua
  Bengio}, \bibinfo{person}{Aaron Courville}, {and} \bibinfo{person}{Yoshua
  Bengio}.} \bibinfo{year}{2016}\natexlab{}.
\newblock \bibinfo{booktitle}{\emph{Deep learning}}. Vol.~\bibinfo{volume}{1}.
\newblock \bibinfo{publisher}{MIT press Cambridge}.
\newblock


\bibitem[\protect\citeauthoryear{Gordon, Czerny, Miyaki, and Beigl}{Gordon
  et~al\mbox{.}}{2012}]%
        {gordon2012energy}
\bibfield{author}{\bibinfo{person}{Dawud Gordon}, \bibinfo{person}{Jurgen
  Czerny}, \bibinfo{person}{Takashi Miyaki}, {and} \bibinfo{person}{Michael
  Beigl}.} \bibinfo{year}{2012}\natexlab{}.
\newblock \showarticletitle{Energy-efficient activity recognition using
  prediction}. In \bibinfo{booktitle}{\emph{2012 16th International Symposium
  on Wearable Computers}}. IEEE, \bibinfo{pages}{29--36}.
\newblock


\bibitem[\protect\citeauthoryear{Hansel, Poguntke, Haddadi, Alomainy, and
  Schmidt}{Hansel et~al\mbox{.}}{2018}]%
        {hansel2018put}
\bibfield{author}{\bibinfo{person}{Katrin Hansel}, \bibinfo{person}{Romina
  Poguntke}, \bibinfo{person}{Hamed Haddadi}, \bibinfo{person}{Akram Alomainy},
  {and} \bibinfo{person}{Albrecht Schmidt}.} \bibinfo{year}{2018}\natexlab{}.
\newblock \showarticletitle{What to put on the user: Sensing technologies for
  studies and physiology aware systems}. In
  \bibinfo{booktitle}{\emph{Proceedings of the 2018 CHI Conference on Human
  Factors in Computing Systems}}. \bibinfo{pages}{1--14}.
\newblock


\bibitem[\protect\citeauthoryear{He, Zhang, Ren, and Sun}{He
  et~al\mbox{.}}{2015}]%
        {he2015spatial}
\bibfield{author}{\bibinfo{person}{Kaiming He}, \bibinfo{person}{Xiangyu
  Zhang}, \bibinfo{person}{Shaoqing Ren}, {and} \bibinfo{person}{Jian Sun}.}
  \bibinfo{year}{2015}\natexlab{}.
\newblock \showarticletitle{Spatial pyramid pooling in deep convolutional
  networks for visual recognition}.
\newblock \bibinfo{journal}{\emph{IEEE transactions on pattern analysis and
  machine intelligence}} \bibinfo{volume}{37}, \bibinfo{number}{9}
  (\bibinfo{year}{2015}), \bibinfo{pages}{1904--1916}.
\newblock


\bibitem[\protect\citeauthoryear{Hochreiter and Schmidhuber}{Hochreiter and
  Schmidhuber}{1997}]%
        {hochreiter1997long}
\bibfield{author}{\bibinfo{person}{Sepp Hochreiter} {and}
  \bibinfo{person}{Jurgen Schmidhuber}.} \bibinfo{year}{1997}\natexlab{}.
\newblock \showarticletitle{Long short-term memory}.
\newblock \bibinfo{journal}{\emph{Neural computation}} \bibinfo{volume}{9},
  \bibinfo{number}{8} (\bibinfo{year}{1997}), \bibinfo{pages}{1735--1780}.
\newblock


\bibitem[\protect\citeauthoryear{Hounslow, Brewster, Lear, Guttridge, Daly,
  Whitney, and Gleiss}{Hounslow et~al\mbox{.}}{2019}]%
        {hounslow2019assessing}
\bibfield{author}{\bibinfo{person}{JL Hounslow}, \bibinfo{person}{LR Brewster},
  \bibinfo{person}{KO Lear}, \bibinfo{person}{TL Guttridge}, \bibinfo{person}{R
  Daly}, \bibinfo{person}{NM Whitney}, {and} \bibinfo{person}{AC Gleiss}.}
  \bibinfo{year}{2019}\natexlab{}.
\newblock \showarticletitle{Assessing the effects of sampling frequency on
  behavioural classification of accelerometer data}.
\newblock \bibinfo{journal}{\emph{Journal of experimental marine biology and
  ecology}}  \bibinfo{volume}{512} (\bibinfo{year}{2019}),
  \bibinfo{pages}{22--30}.
\newblock


\bibitem[\protect\citeauthoryear{Hsu, Chu, Zhou, and Cheng}{Hsu
  et~al\mbox{.}}{2015}]%
        {hsu2015two}
\bibfield{author}{\bibinfo{person}{Hui-Huang Hsu}, \bibinfo{person}{Chin-Ting
  Chu}, \bibinfo{person}{Yinghui Zhou}, {and} \bibinfo{person}{Zixue Cheng}.}
  \bibinfo{year}{2015}\natexlab{}.
\newblock \showarticletitle{Two-phase activity recognition with smartphone
  sensors}. In \bibinfo{booktitle}{\emph{2015 18th International Conference on
  Network-Based Information Systems}}. IEEE, \bibinfo{pages}{611--615}.
\newblock


\bibitem[\protect\citeauthoryear{Ignatov}{Ignatov}{2018}]%
        {ignatov2018real}
\bibfield{author}{\bibinfo{person}{Andrey Ignatov}.}
  \bibinfo{year}{2018}\natexlab{}.
\newblock \showarticletitle{Real-time human activity recognition from
  accelerometer data using Convolutional Neural Networks}.
\newblock \bibinfo{journal}{\emph{Applied Soft Computing}}
  \bibinfo{volume}{62} (\bibinfo{year}{2018}), \bibinfo{pages}{915--922}.
\newblock


\bibitem[\protect\citeauthoryear{Jeyakumar, Lai, Suda, and
  Srivastava}{Jeyakumar et~al\mbox{.}}{2019}]%
        {jeyakumar2019sensehar}
\bibfield{author}{\bibinfo{person}{Jeya~Vikranth Jeyakumar},
  \bibinfo{person}{Liangzhen Lai}, \bibinfo{person}{Naveen Suda}, {and}
  \bibinfo{person}{Mani Srivastava}.} \bibinfo{year}{2019}\natexlab{}.
\newblock \showarticletitle{SenseHAR: a robust virtual activity sensor for
  smartphones and wearables}. In \bibinfo{booktitle}{\emph{Proceedings of the
  17th Conference on Embedded Networked Sensor Systems}}.
  \bibinfo{pages}{15--28}.
\newblock


\bibitem[\protect\citeauthoryear{Karim, Majumdar, Darabi, and Chen}{Karim
  et~al\mbox{.}}{2017}]%
        {karim2017lstm}
\bibfield{author}{\bibinfo{person}{Fazle Karim}, \bibinfo{person}{Somshubra
  Majumdar}, \bibinfo{person}{Houshang Darabi}, {and} \bibinfo{person}{Shun
  Chen}.} \bibinfo{year}{2017}\natexlab{}.
\newblock \showarticletitle{{LSTM} fully convolutional networks for time series
  classification}.
\newblock \bibinfo{journal}{\emph{IEEE access}}  \bibinfo{volume}{6}
  (\bibinfo{year}{2017}), \bibinfo{pages}{1662--1669}.
\newblock


\bibitem[\protect\citeauthoryear{Katevas, Leontiadis, Pielot, and
  Serra}{Katevas et~al\mbox{.}}{2017}]%
        {katevas2017practical}
\bibfield{author}{\bibinfo{person}{Kleomenis Katevas}, \bibinfo{person}{Ilias
  Leontiadis}, \bibinfo{person}{Martin Pielot}, {and} \bibinfo{person}{Joan
  Serra}.} \bibinfo{year}{2017}\natexlab{}.
\newblock \showarticletitle{Practical processing of mobile sensor data for
  continual deep learning predictions}. In
  \bibinfo{booktitle}{\emph{Proceedings of the 1st International Workshop on
  Deep Learning for mobile systems and applications}}. \bibinfo{pages}{19--24}.
\newblock


\bibitem[\protect\citeauthoryear{Kayhan and Gemert}{Kayhan and Gemert}{2020}]%
        {kayhan2020translation}
\bibfield{author}{\bibinfo{person}{Osman~Semih Kayhan} {and}
  \bibinfo{person}{Jan C~van Gemert}.} \bibinfo{year}{2020}\natexlab{}.
\newblock \showarticletitle{On translation invariance in cnns: Convolutional
  layers can exploit absolute spatial location}. In
  \bibinfo{booktitle}{\emph{Proceedings of the IEEE/CVF Conference on Computer
  Vision and Pattern Recognition}}. \bibinfo{pages}{14274--14285}.
\newblock


\bibitem[\protect\citeauthoryear{Khan, Hammerla, Mellor, and Plotz}{Khan
  et~al\mbox{.}}{2016}]%
        {khan2016optimising}
\bibfield{author}{\bibinfo{person}{Aftab Khan}, \bibinfo{person}{Nils
  Hammerla}, \bibinfo{person}{Sebastian Mellor}, {and} \bibinfo{person}{Thomas
  Plotz}.} \bibinfo{year}{2016}\natexlab{}.
\newblock \showarticletitle{Optimising sampling rates for accelerometer-based
  human activity recognition}.
\newblock \bibinfo{journal}{\emph{Pattern Recognition Letters}}
  \bibinfo{volume}{73} (\bibinfo{year}{2016}), \bibinfo{pages}{33--40}.
\newblock


\bibitem[\protect\citeauthoryear{Kingma and Ba}{Kingma and Ba}{2014}]%
        {kingma2014adam}
\bibfield{author}{\bibinfo{person}{Diederik~P Kingma} {and}
  \bibinfo{person}{Jimmy Ba}.} \bibinfo{year}{2014}\natexlab{}.
\newblock \showarticletitle{Adam: A method for stochastic optimization}. In
  \bibinfo{booktitle}{\emph{Proceedings of the 3rd International Conference on
  Learning Representations (ICLR)}}.
\newblock


\bibitem[\protect\citeauthoryear{Koping, Shirahama, and Grzegorzek}{Koping
  et~al\mbox{.}}{2018}]%
        {koping2018general}
\bibfield{author}{\bibinfo{person}{Lukas Koping}, \bibinfo{person}{Kimiaki
  Shirahama}, {and} \bibinfo{person}{Marcin Grzegorzek}.}
  \bibinfo{year}{2018}\natexlab{}.
\newblock \showarticletitle{A general framework for sensor-based human activity
  recognition}.
\newblock \bibinfo{journal}{\emph{Computers in biology and medicine}}
  \bibinfo{volume}{95} (\bibinfo{year}{2018}), \bibinfo{pages}{248--260}.
\newblock


\bibitem[\protect\citeauthoryear{LeCun and Bengio}{LeCun and Bengio}{1995}]%
        {lecun1995convolutional}
\bibfield{author}{\bibinfo{person}{Yann LeCun} {and} \bibinfo{person}{Yoshua
  Bengio}.} \bibinfo{year}{1995}\natexlab{}.
\newblock \showarticletitle{Convolutional networks for images, speech, and time
  series}.
\newblock \bibinfo{journal}{\emph{The handbook of brain theory and neural
  networks}} \bibinfo{volume}{3361}, \bibinfo{number}{10}
  (\bibinfo{year}{1995}), \bibinfo{pages}{1995}.
\newblock


\bibitem[\protect\citeauthoryear{LeCun, Bengio, and Hinton}{LeCun
  et~al\mbox{.}}{2015}]%
        {lecun2015deep}
\bibfield{author}{\bibinfo{person}{Yann LeCun}, \bibinfo{person}{Yoshua
  Bengio}, {and} \bibinfo{person}{Geoffrey Hinton}.}
  \bibinfo{year}{2015}\natexlab{}.
\newblock \showarticletitle{Deep learning}.
\newblock \bibinfo{journal}{\emph{nature}} \bibinfo{volume}{521},
  \bibinfo{number}{7553} (\bibinfo{year}{2015}), \bibinfo{pages}{436--444}.
\newblock


\bibitem[\protect\citeauthoryear{Liang, Zhou, Yu, and Guo}{Liang
  et~al\mbox{.}}{2014}]%
        {liang2014energy}
\bibfield{author}{\bibinfo{person}{Yunji Liang}, \bibinfo{person}{Xingshe
  Zhou}, \bibinfo{person}{Zhiwen Yu}, {and} \bibinfo{person}{Bin Guo}.}
  \bibinfo{year}{2014}\natexlab{}.
\newblock \showarticletitle{Energy-efficient motion related activity
  recognition on mobile devices for pervasive healthcare}.
\newblock \bibinfo{journal}{\emph{Mobile Networks and Applications}}
  \bibinfo{volume}{19}, \bibinfo{number}{3} (\bibinfo{year}{2014}),
  \bibinfo{pages}{303--317}.
\newblock


\bibitem[\protect\citeauthoryear{Lin, Chen, and Yan}{Lin et~al\mbox{.}}{2014}]%
        {lin2013network}
\bibfield{author}{\bibinfo{person}{Min Lin}, \bibinfo{person}{Qiang Chen},
  {and} \bibinfo{person}{Shuicheng Yan}.} \bibinfo{year}{2014}\natexlab{}.
\newblock \showarticletitle{Network In Network}. In
  \bibinfo{booktitle}{\emph{Proceedings of the 2nd International Conference on
  Learning Representations, {ICLR}}}.
\newblock


\bibitem[\protect\citeauthoryear{Long, Shelhamer, and Darrell}{Long
  et~al\mbox{.}}{2015}]%
        {long2015fully}
\bibfield{author}{\bibinfo{person}{Jonathan Long}, \bibinfo{person}{Evan
  Shelhamer}, {and} \bibinfo{person}{Trevor Darrell}.}
  \bibinfo{year}{2015}\natexlab{}.
\newblock \showarticletitle{Fully convolutional networks for semantic
  segmentation}. In \bibinfo{booktitle}{\emph{Proceedings of the IEEE
  conference on computer vision and pattern recognition}}.
  \bibinfo{pages}{3431--3440}.
\newblock


\bibitem[\protect\citeauthoryear{Malekzadeh, Clegg, Cavallaro, and
  Haddadi}{Malekzadeh et~al\mbox{.}}{2018}]%
        {malekzadeh2018protecting}
\bibfield{author}{\bibinfo{person}{Mohammad Malekzadeh},
  \bibinfo{person}{Richard~G. Clegg}, \bibinfo{person}{Andrea Cavallaro}, {and}
  \bibinfo{person}{Hamed Haddadi}.} \bibinfo{year}{2018}\natexlab{}.
\newblock \showarticletitle{Protecting Sensory Data Against Sensitive
  Inferences}. In \bibinfo{booktitle}{\emph{Proceedings of the 1st Workshop on
  Privacy by Design in Distributed Systems}} (Porto, Portugal)
  \emph{(\bibinfo{series}{W-P2DS18})}. \bibinfo{publisher}{ACM}, Article
  \bibinfo{articleno}{2}, \bibinfo{numpages}{6}~pages.
\newblock


\bibitem[\protect\citeauthoryear{Malekzadeh, Clegg, Cavallaro, and
  Haddadi}{Malekzadeh et~al\mbox{.}}{2019}]%
        {malekzadeh2018mobile}
\bibfield{author}{\bibinfo{person}{Mohammad Malekzadeh},
  \bibinfo{person}{Richard~G. Clegg}, \bibinfo{person}{Andrea Cavallaro}, {and}
  \bibinfo{person}{Hamed Haddadi}.} \bibinfo{year}{2019}\natexlab{}.
\newblock \showarticletitle{Mobile Sensor Data Anonymization}. In
  \bibinfo{booktitle}{\emph{Proceedings of the International Conference on
  Internet of Things Design and Implementation (IoTDI)}} (Montreal, Quebec,
  Canada). \bibinfo{publisher}{ACM}, \bibinfo{pages}{49--58}.
\newblock


\bibitem[\protect\citeauthoryear{Malekzadeh, Clegg, Cavallaro, and
  Haddadi}{Malekzadeh et~al\mbox{.}}{2020}]%
        {malekzadeh2020privacy}
\bibfield{author}{\bibinfo{person}{Mohammad Malekzadeh},
  \bibinfo{person}{Richard~G Clegg}, \bibinfo{person}{Andrea Cavallaro}, {and}
  \bibinfo{person}{Hamed Haddadi}.} \bibinfo{year}{2020}\natexlab{}.
\newblock \showarticletitle{Privacy and Utility Preserving Sensor-Data
  Transformations}.
\newblock \bibinfo{journal}{\emph{Pervasive and Mobile Computing}}
  (\bibinfo{year}{2020}).
\newblock


\bibitem[\protect\citeauthoryear{Mathur, Isopoussu, Berthouze, Lane, and
  Kawsar}{Mathur et~al\mbox{.}}{2019}]%
        {mathur2019unsupervised}
\bibfield{author}{\bibinfo{person}{Akhil Mathur}, \bibinfo{person}{Anton
  Isopoussu}, \bibinfo{person}{Nadia Berthouze}, \bibinfo{person}{Nicholas~D
  Lane}, {and} \bibinfo{person}{Fahim Kawsar}.}
  \bibinfo{year}{2019}\natexlab{}.
\newblock \showarticletitle{Unsupervised domain adaptation for robust sensory
  systems}. In \bibinfo{booktitle}{\emph{Adjunct Proceedings of the 2019 ACM
  International Joint Conference on Pervasive and Ubiquitous Computing and
  Proceedings of the 2019 ACM International Symposium on Wearable Computers}}.
  \bibinfo{pages}{505--509}.
\newblock


\bibitem[\protect\citeauthoryear{Mohr, Zhang, and Schueller}{Mohr
  et~al\mbox{.}}{2017}]%
        {mohr2017personal}
\bibfield{author}{\bibinfo{person}{David~C Mohr}, \bibinfo{person}{Mi Zhang},
  {and} \bibinfo{person}{Stephen~M Schueller}.}
  \bibinfo{year}{2017}\natexlab{}.
\newblock \showarticletitle{Personal sensing: understanding mental health using
  ubiquitous sensors and machine learning}.
\newblock \bibinfo{journal}{\emph{Annual review of clinical psychology}}
  \bibinfo{volume}{13} (\bibinfo{year}{2017}), \bibinfo{pages}{23--47}.
\newblock


\bibitem[\protect\citeauthoryear{Nam, Kim, and Lee}{Nam et~al\mbox{.}}{2016}]%
        {nam2016sleep}
\bibfield{author}{\bibinfo{person}{Yunyoung Nam}, \bibinfo{person}{Yeesock
  Kim}, {and} \bibinfo{person}{Jinseok Lee}.} \bibinfo{year}{2016}\natexlab{}.
\newblock \showarticletitle{Sleep monitoring based on a tri-axial accelerometer
  and a pressure sensor}.
\newblock \bibinfo{journal}{\emph{Sensors}} \bibinfo{volume}{16},
  \bibinfo{number}{5} (\bibinfo{year}{2016}), \bibinfo{pages}{750}.
\newblock


\bibitem[\protect\citeauthoryear{Nichol, Achiam, and Schulman}{Nichol
  et~al\mbox{.}}{2018}]%
        {nichol2018first}
\bibfield{author}{\bibinfo{person}{Alex Nichol}, \bibinfo{person}{Joshua
  Achiam}, {and} \bibinfo{person}{John Schulman}.}
  \bibinfo{year}{2018}\natexlab{}.
\newblock \showarticletitle{On first-order meta-learning algorithms}.
\newblock \bibinfo{journal}{\emph{arXiv preprint arXiv:1803.02999}}
  (\bibinfo{year}{2018}).
\newblock


\bibitem[\protect\citeauthoryear{Ordonez and Roggen}{Ordonez and
  Roggen}{2016}]%
        {ordonez2016deep}
\bibfield{author}{\bibinfo{person}{Francisco Ordonez} {and}
  \bibinfo{person}{Daniel Roggen}.} \bibinfo{year}{2016}\natexlab{}.
\newblock \showarticletitle{Deep convolutional and {LSTM} recurrent neural
  networks for multimodal wearable activity recognition}.
\newblock \bibinfo{journal}{\emph{Sensors}} \bibinfo{volume}{16},
  \bibinfo{number}{1} (\bibinfo{year}{2016}), \bibinfo{pages}{115}.
\newblock


\bibitem[\protect\citeauthoryear{Poole, Ozair, Van Den~Oord, Alemi, and
  Tucker}{Poole et~al\mbox{.}}{2019}]%
        {poole2019variational}
\bibfield{author}{\bibinfo{person}{Ben Poole}, \bibinfo{person}{Sherjil Ozair},
  \bibinfo{person}{Aaron Van Den~Oord}, \bibinfo{person}{Alex Alemi}, {and}
  \bibinfo{person}{George Tucker}.} \bibinfo{year}{2019}\natexlab{}.
\newblock \showarticletitle{On variational bounds of mutual information}. In
  \bibinfo{booktitle}{\emph{International Conference on Machine Learning}}.
  PMLR, \bibinfo{pages}{5171--5180}.
\newblock


\bibitem[\protect\citeauthoryear{Qi, Keally, Zhou, Li, and Ren}{Qi
  et~al\mbox{.}}{2013}]%
        {qi2013adasense}
\bibfield{author}{\bibinfo{person}{Xin Qi}, \bibinfo{person}{Matthew Keally},
  \bibinfo{person}{Gang Zhou}, \bibinfo{person}{Yantao Li}, {and}
  \bibinfo{person}{Zhen Ren}.} \bibinfo{year}{2013}\natexlab{}.
\newblock \showarticletitle{AdaSense: Adapting sampling rates for activity
  recognition in body sensor networks}. In \bibinfo{booktitle}{\emph{2013 IEEE
  19th Real-Time and Embedded Technology and Applications Symposium (RTAS)}}.
  IEEE, \bibinfo{pages}{163--172}.
\newblock


\bibitem[\protect\citeauthoryear{Qian}{Qian}{1999}]%
        {qian1999momentum}
\bibfield{author}{\bibinfo{person}{Ning Qian}.}
  \bibinfo{year}{1999}\natexlab{}.
\newblock \showarticletitle{On the momentum term in gradient descent learning
  algorithms}.
\newblock \bibinfo{journal}{\emph{Neural networks}} \bibinfo{volume}{12},
  \bibinfo{number}{1} (\bibinfo{year}{1999}), \bibinfo{pages}{145--151}.
\newblock


\bibitem[\protect\citeauthoryear{Quiring, Klein, Arp, Johns, and Rieck}{Quiring
  et~al\mbox{.}}{2020}]%
        {quiring2020adversarial}
\bibfield{author}{\bibinfo{person}{Erwin Quiring}, \bibinfo{person}{David
  Klein}, \bibinfo{person}{Daniel Arp}, \bibinfo{person}{Martin Johns}, {and}
  \bibinfo{person}{Konrad Rieck}.} \bibinfo{year}{2020}\natexlab{}.
\newblock \showarticletitle{Adversarial Preprocessing: Understanding and
  Preventing Image-Scaling Attacks in Machine Learning}. In
  \bibinfo{booktitle}{\emph{29th {USENIX} Security Symposium ({USENIX} Security
  20)}}. \bibinfo{publisher}{{USENIX} Association},
  \bibinfo{pages}{1363--1380}.
\newblock
\showISBNx{978-1-939133-17-5}


\bibitem[\protect\citeauthoryear{Raij, Ghosh, Kumar, and Srivastava}{Raij
  et~al\mbox{.}}{2011}]%
        {raij2011privacy}
\bibfield{author}{\bibinfo{person}{Andrew Raij}, \bibinfo{person}{Animikh
  Ghosh}, \bibinfo{person}{Santosh Kumar}, {and} \bibinfo{person}{Mani
  Srivastava}.} \bibinfo{year}{2011}\natexlab{}.
\newblock \showarticletitle{Privacy risks emerging from the adoption of
  innocuous wearable sensors in the mobile environment}. In
  \bibinfo{booktitle}{\emph{Proceedings of the SIGCHI Conference on Human
  Factors in Computing Systems}}. \bibinfo{pages}{11--20}.
\newblock


\bibitem[\protect\citeauthoryear{Reinertsen and Clifford}{Reinertsen and
  Clifford}{2018}]%
        {reinertsen2018review}
\bibfield{author}{\bibinfo{person}{Erik Reinertsen} {and}
  \bibinfo{person}{Gari~D Clifford}.} \bibinfo{year}{2018}\natexlab{}.
\newblock \showarticletitle{A review of physiological and behavioral monitoring
  with digital sensors for neuropsychiatric illnesses}.
\newblock \bibinfo{journal}{\emph{Physiological measurement}}
  \bibinfo{volume}{39}, \bibinfo{number}{5} (\bibinfo{year}{2018}),
  \bibinfo{pages}{05TR01}.
\newblock


\bibitem[\protect\citeauthoryear{Richoz, Wang, Birch, and Roggen}{Richoz
  et~al\mbox{.}}{2020}]%
        {richoz2020transportation}
\bibfield{author}{\bibinfo{person}{Sebastien Richoz}, \bibinfo{person}{Lin
  Wang}, \bibinfo{person}{Philip Birch}, {and} \bibinfo{person}{Daniel
  Roggen}.} \bibinfo{year}{2020}\natexlab{}.
\newblock \showarticletitle{Transportation mode recognition fusing wearable
  motion, sound and vision sensors}.
\newblock \bibinfo{journal}{\emph{IEEE Sensors Journal}}
  (\bibinfo{year}{2020}).
\newblock


\bibitem[\protect\citeauthoryear{Ronao and Cho}{Ronao and Cho}{2016}]%
        {ronao2016human}
\bibfield{author}{\bibinfo{person}{Charissa~Ann Ronao} {and}
  \bibinfo{person}{Sung-Bae Cho}.} \bibinfo{year}{2016}\natexlab{}.
\newblock \showarticletitle{Human activity recognition with smartphone sensors
  using deep learning neural networks}.
\newblock \bibinfo{journal}{\emph{Expert systems with applications}}
  \bibinfo{volume}{59} (\bibinfo{year}{2016}), \bibinfo{pages}{235--244}.
\newblock


\bibitem[\protect\citeauthoryear{Saeedi and El-Sheimy}{Saeedi and
  El-Sheimy}{2015}]%
        {saeedi2015activity}
\bibfield{author}{\bibinfo{person}{Sara Saeedi} {and} \bibinfo{person}{Naser
  El-Sheimy}.} \bibinfo{year}{2015}\natexlab{}.
\newblock \showarticletitle{Activity recognition using fusion of low-cost
  sensors on a smartphone for mobile navigation application}.
\newblock \bibinfo{journal}{\emph{Micromachines}} \bibinfo{volume}{6},
  \bibinfo{number}{8} (\bibinfo{year}{2015}), \bibinfo{pages}{1100--1134}.
\newblock


\bibitem[\protect\citeauthoryear{Schmidhuber}{Schmidhuber}{2015}]%
        {schmidhuber2015deep}
\bibfield{author}{\bibinfo{person}{Jurgen Schmidhuber}.}
  \bibinfo{year}{2015}\natexlab{}.
\newblock \showarticletitle{Deep learning in neural networks: An overview}.
\newblock \bibinfo{journal}{\emph{Neural networks}}  \bibinfo{volume}{61}
  (\bibinfo{year}{2015}), \bibinfo{pages}{85--117}.
\newblock


\bibitem[\protect\citeauthoryear{Shepard, Wilson, Halsey, Quintana, Laich,
  Gleiss, Liebsch, Myers, and Norman}{Shepard et~al\mbox{.}}{2008}]%
        {shepard2008derivation}
\bibfield{author}{\bibinfo{person}{Emily~LC Shepard}, \bibinfo{person}{Rory~P
  Wilson}, \bibinfo{person}{Lewis~G Halsey}, \bibinfo{person}{Flavio Quintana},
  \bibinfo{person}{Agustina~Gomez Laich}, \bibinfo{person}{Adrian~C Gleiss},
  \bibinfo{person}{Nikolai Liebsch}, \bibinfo{person}{Andrew~E Myers}, {and}
  \bibinfo{person}{Brad Norman}.} \bibinfo{year}{2008}\natexlab{}.
\newblock \showarticletitle{Derivation of body motion via appropriate smoothing
  of acceleration data}.
\newblock \bibinfo{journal}{\emph{Aquatic Biology}} \bibinfo{volume}{4},
  \bibinfo{number}{3} (\bibinfo{year}{2008}), \bibinfo{pages}{235--241}.
\newblock


\bibitem[\protect\citeauthoryear{Shoaib, Bosch, Incel, Scholten, and
  Havinga}{Shoaib et~al\mbox{.}}{2016}]%
        {shoaib2016complex}
\bibfield{author}{\bibinfo{person}{Muhammad Shoaib}, \bibinfo{person}{Stephan
  Bosch}, \bibinfo{person}{Ozlem~Durmaz Incel}, \bibinfo{person}{Hans
  Scholten}, {and} \bibinfo{person}{Paul~JM Havinga}.}
  \bibinfo{year}{2016}\natexlab{}.
\newblock \showarticletitle{Complex human activity recognition using smartphone
  and wrist-worn motion sensors}.
\newblock \bibinfo{journal}{\emph{Sensors}} \bibinfo{volume}{16},
  \bibinfo{number}{4} (\bibinfo{year}{2016}), \bibinfo{pages}{426}.
\newblock


\bibitem[\protect\citeauthoryear{Shumway, Stoffer, and Stoffer}{Shumway
  et~al\mbox{.}}{2000}]%
        {shumway2000time}
\bibfield{author}{\bibinfo{person}{Robert~H Shumway}, \bibinfo{person}{David~S
  Stoffer}, {and} \bibinfo{person}{David~S Stoffer}.}
  \bibinfo{year}{2000}\natexlab{}.
\newblock \bibinfo{booktitle}{\emph{Time series analysis and its
  applications}}. Vol.~\bibinfo{volume}{3}.
\newblock \bibinfo{publisher}{Springer}.
\newblock


\bibitem[\protect\citeauthoryear{Srivastava, Hinton, Krizhevsky, Sutskever, and
  Salakhutdinov}{Srivastava et~al\mbox{.}}{2014}]%
        {srivastava2014dropout}
\bibfield{author}{\bibinfo{person}{Nitish Srivastava},
  \bibinfo{person}{Geoffrey Hinton}, \bibinfo{person}{Alex Krizhevsky},
  \bibinfo{person}{Ilya Sutskever}, {and} \bibinfo{person}{Ruslan
  Salakhutdinov}.} \bibinfo{year}{2014}\natexlab{}.
\newblock \showarticletitle{Dropout: a simple way to prevent neural networks
  from overfitting}.
\newblock \bibinfo{journal}{\emph{The journal of machine learning research}}
  \bibinfo{volume}{15}, \bibinfo{number}{1} (\bibinfo{year}{2014}),
  \bibinfo{pages}{1929--1958}.
\newblock


\bibitem[\protect\citeauthoryear{Tieleman and Hinton}{Tieleman and
  Hinton}{2012}]%
        {tieleman2012lecture}
\bibfield{author}{\bibinfo{person}{Tijmen Tieleman} {and}
  \bibinfo{person}{Geoffrey Hinton}.} \bibinfo{year}{2012}\natexlab{}.
\newblock \showarticletitle{Lecture 6.5-rmsprop: Divide the gradient by a
  running average of its recent magnitude}.
\newblock \bibinfo{journal}{\emph{COURSERA: Neural networks for machine
  learning}} \bibinfo{volume}{4}, \bibinfo{number}{2} (\bibinfo{year}{2012}),
  \bibinfo{pages}{26--31}.
\newblock


\bibitem[\protect\citeauthoryear{Vavoulas, Chatzaki, Malliotakis, Pediaditis,
  and Tsiknakis}{Vavoulas et~al\mbox{.}}{2016}]%
        {vavoulas2016mobiact}
\bibfield{author}{\bibinfo{person}{George Vavoulas},
  \bibinfo{person}{Charikleia Chatzaki}, \bibinfo{person}{Thodoris
  Malliotakis}, \bibinfo{person}{Matthew Pediaditis}, {and}
  \bibinfo{person}{Manolis Tsiknakis}.} \bibinfo{year}{2016}\natexlab{}.
\newblock \showarticletitle{The MobiAct Dataset: Recognition of Activities of
  Daily Living using Smartphones.}. In
  \bibinfo{booktitle}{\emph{ICT4AgeingWell}}. \bibinfo{pages}{143--151}.
\newblock


\bibitem[\protect\citeauthoryear{Walton, Casey, Mitsch, Vazquez-Diosdado, Yan,
  Dottorini, Ellis, Winterlich, and Kaler}{Walton et~al\mbox{.}}{2018}]%
        {walton2018evaluation}
\bibfield{author}{\bibinfo{person}{Emily Walton}, \bibinfo{person}{Christy
  Casey}, \bibinfo{person}{Jurgen Mitsch}, \bibinfo{person}{Jorge~A
  Vazquez-Diosdado}, \bibinfo{person}{Juan Yan}, \bibinfo{person}{Tania
  Dottorini}, \bibinfo{person}{Keith~A Ellis}, \bibinfo{person}{Anthony
  Winterlich}, {and} \bibinfo{person}{Jasmeet Kaler}.}
  \bibinfo{year}{2018}\natexlab{}.
\newblock \showarticletitle{Evaluation of sampling frequency, window size and
  sensor position for classification of sheep behaviour}.
\newblock \bibinfo{journal}{\emph{Royal Society open science}}
  \bibinfo{volume}{5}, \bibinfo{number}{2} (\bibinfo{year}{2018}),
  \bibinfo{pages}{171442}.
\newblock


\bibitem[\protect\citeauthoryear{Wang, Chen, Hao, Peng, and Hu}{Wang
  et~al\mbox{.}}{2019}]%
        {wang2019deep}
\bibfield{author}{\bibinfo{person}{Jindong Wang}, \bibinfo{person}{Yiqiang
  Chen}, \bibinfo{person}{Shuji Hao}, \bibinfo{person}{Xiaohui Peng}, {and}
  \bibinfo{person}{Lisha Hu}.} \bibinfo{year}{2019}\natexlab{}.
\newblock \showarticletitle{Deep learning for sensor-based activity
  recognition: A survey}.
\newblock \bibinfo{journal}{\emph{Pattern Recognition Letters}}
  \bibinfo{volume}{119} (\bibinfo{year}{2019}), \bibinfo{pages}{3--11}.
\newblock


\bibitem[\protect\citeauthoryear{Wang, Aung, Abdullah, Brian, Campbell,
  Choudhury, Hauser, Kane, Merrill, Scherer, Tseng, and Ben-Zeev}{Wang
  et~al\mbox{.}}{2016}]%
        {wang2016crosscheck}
\bibfield{author}{\bibinfo{person}{Rui Wang}, \bibinfo{person}{Min S.~H. Aung},
  \bibinfo{person}{Saeed Abdullah}, \bibinfo{person}{Rachel Brian},
  \bibinfo{person}{Andrew~T. Campbell}, \bibinfo{person}{Tanzeem Choudhury},
  \bibinfo{person}{Marta Hauser}, \bibinfo{person}{John Kane},
  \bibinfo{person}{Michael Merrill}, \bibinfo{person}{Emily~A. Scherer},
  \bibinfo{person}{Vincent W.~S. Tseng}, {and} \bibinfo{person}{Dror
  Ben-Zeev}.} \bibinfo{year}{2016}\natexlab{}.
\newblock \showarticletitle{CrossCheck: Toward Passive Sensing and Detection of
  Mental Health Changes in People with Schizophrenia}. In
  \bibinfo{booktitle}{\emph{Proceedings of the 2016 ACM International Joint
  Conference on Pervasive and Ubiquitous Computing}} (Heidelberg, Germany).
  \bibinfo{publisher}{ACM}, \bibinfo{pages}{886–897}.
\newblock


\bibitem[\protect\citeauthoryear{Xu, Xiao, Zhang, Yang, and Zhang}{Xu
  et~al\mbox{.}}{2015}]%
        {xu2014scale}
\bibfield{author}{\bibinfo{person}{Yichong Xu}, \bibinfo{person}{Tianjun Xiao},
  \bibinfo{person}{Jiaxing Zhang}, \bibinfo{person}{Kuiyuan Yang}, {and}
  \bibinfo{person}{Zheng Zhang}.} \bibinfo{year}{2015}\natexlab{}.
\newblock \showarticletitle{Scale-invariant convolutional neural networks}. In
  \bibinfo{booktitle}{\emph{Proceedings of the IEEE conference on computer
  vision and pattern recognition}}.
\newblock


\bibitem[\protect\citeauthoryear{Yan, Subbaraju, Chakraborty, Misra, and
  Aberer}{Yan et~al\mbox{.}}{2012}]%
        {yan2012energy}
\bibfield{author}{\bibinfo{person}{Zhixian Yan}, \bibinfo{person}{Vigneshwaran
  Subbaraju}, \bibinfo{person}{Dipanjan Chakraborty}, \bibinfo{person}{Archan
  Misra}, {and} \bibinfo{person}{Karl Aberer}.}
  \bibinfo{year}{2012}\natexlab{}.
\newblock \showarticletitle{Energy-efficient continuous activity recognition on
  mobile phones: An activity-adaptive approach}. In
  \bibinfo{booktitle}{\emph{2012 16th international symposium on wearable
  computers}}. IEEE, \bibinfo{pages}{17--24}.
\newblock


\bibitem[\protect\citeauthoryear{Yang, Nguyen, San, Li, and Krishnaswamy}{Yang
  et~al\mbox{.}}{2015}]%
        {yang2015deep}
\bibfield{author}{\bibinfo{person}{Jianbo Yang}, \bibinfo{person}{Minh~Nhut
  Nguyen}, \bibinfo{person}{Phyo~Phyo San}, \bibinfo{person}{Xiao~Li Li}, {and}
  \bibinfo{person}{Shonali Krishnaswamy}.} \bibinfo{year}{2015}\natexlab{}.
\newblock \showarticletitle{Deep convolutional neural networks on multichannel
  time series for human activity recognition}. In
  \bibinfo{booktitle}{\emph{Twenty-Fourth International Joint Conference on
  Artificial Intelligence}}.
\newblock


\bibitem[\protect\citeauthoryear{Yang, Chen, Yu, Zhang, Lu, and Sun}{Yang
  et~al\mbox{.}}{2020}]%
        {yang2020instance}
\bibfield{author}{\bibinfo{person}{Xiaodong Yang}, \bibinfo{person}{Yiqiang
  Chen}, \bibinfo{person}{Hanchao Yu}, \bibinfo{person}{Yingwei Zhang},
  \bibinfo{person}{Wang Lu}, {and} \bibinfo{person}{Ruizhe Sun}.}
  \bibinfo{year}{2020}\natexlab{}.
\newblock \showarticletitle{Instance-Wise Dynamic Sensor Selection for Human
  Activity Recognition}. In \bibinfo{booktitle}{\emph{Proceedings of the AAAI
  Conference on Artificial Intelligence}}, Vol.~\bibinfo{volume}{34}.
  \bibinfo{pages}{1104--1111}.
\newblock


\bibitem[\protect\citeauthoryear{Yao, Hu, Zhao, Zhang, and Abdelzaher}{Yao
  et~al\mbox{.}}{2017}]%
        {yao2017deepsense}
\bibfield{author}{\bibinfo{person}{Shuochao Yao}, \bibinfo{person}{Shaohan Hu},
  \bibinfo{person}{Yiran Zhao}, \bibinfo{person}{Aston Zhang}, {and}
  \bibinfo{person}{Tarek Abdelzaher}.} \bibinfo{year}{2017}\natexlab{}.
\newblock \showarticletitle{Deepsense: A unified deep learning framework for
  time-series mobile sensing data processing}. In
  \bibinfo{booktitle}{\emph{Proceedings of the 26th International Conference on
  World Wide Web}}. \bibinfo{pages}{351--360}.
\newblock


\bibitem[\protect\citeauthoryear{Zappi, Lombriser, Stiefmeier, Farella, Roggen,
  Benini, and Troster}{Zappi et~al\mbox{.}}{2008}]%
        {zappi2008activity}
\bibfield{author}{\bibinfo{person}{Piero Zappi}, \bibinfo{person}{Clemens
  Lombriser}, \bibinfo{person}{Thomas Stiefmeier}, \bibinfo{person}{Elisabetta
  Farella}, \bibinfo{person}{Daniel Roggen}, \bibinfo{person}{Luca Benini},
  {and} \bibinfo{person}{Gerhard Troster}.} \bibinfo{year}{2008}\natexlab{}.
\newblock \showarticletitle{Activity recognition from on-body sensors:
  accuracy-power trade-off by dynamic sensor selection}. In
  \bibinfo{booktitle}{\emph{European Conference on Wireless Sensor Networks}}.
  Springer, \bibinfo{pages}{17--33}.
\newblock


\bibitem[\protect\citeauthoryear{Zhao, Shi, Qi, Wang, and Jia}{Zhao
  et~al\mbox{.}}{2017}]%
        {zhao2017pyramid}
\bibfield{author}{\bibinfo{person}{Hengshuang Zhao}, \bibinfo{person}{Jianping
  Shi}, \bibinfo{person}{Xiaojuan Qi}, \bibinfo{person}{Xiaogang Wang}, {and}
  \bibinfo{person}{Jiaya Jia}.} \bibinfo{year}{2017}\natexlab{}.
\newblock \showarticletitle{Pyramid scene parsing network}. In
  \bibinfo{booktitle}{\emph{Proceedings of the IEEE conference on computer
  vision and pattern recognition}}. \bibinfo{pages}{2881--2890}.
\newblock


\bibitem[\protect\citeauthoryear{Zhao, Yang, Chevalier, Xu, and Zhang}{Zhao
  et~al\mbox{.}}{2018}]%
        {zhao2018deep}
\bibfield{author}{\bibinfo{person}{Yu Zhao}, \bibinfo{person}{Rennong Yang},
  \bibinfo{person}{Guillaume Chevalier}, \bibinfo{person}{Ximeng Xu}, {and}
  \bibinfo{person}{Zhenxing Zhang}.} \bibinfo{year}{2018}\natexlab{}.
\newblock \showarticletitle{Deep residual bidir-{LSTM} for human activity
  recognition using wearable sensors}.
\newblock \bibinfo{journal}{\emph{Mathematical Problems in Engineering}}
  \bibinfo{volume}{2018} (\bibinfo{year}{2018}).
\newblock


\bibitem[\protect\citeauthoryear{Zhu, Li, and Chen}{Zhu et~al\mbox{.}}{2013}]%
        {zhu2013apt}
\bibfield{author}{\bibinfo{person}{Xiaojun Zhu}, \bibinfo{person}{Qun Li},
  {and} \bibinfo{person}{Guihai Chen}.} \bibinfo{year}{2013}\natexlab{}.
\newblock \showarticletitle{APT: Accurate outdoor pedestrian tracking with
  smartphones}. In \bibinfo{booktitle}{\emph{2013 Proceedings IEEE INFOCOM}}.
  IEEE, \bibinfo{pages}{2508--2516}.
\newblock


\end{thebibliography}
\end{document}